\documentclass[twocolumn,runningheads]{llncs}
\usepackage[T1]{fontenc}
\usepackage{graphicx}
\usepackage[margin=0.75in]{geometry}
\usepackage{amsmath,amssymb}
\usepackage{bm}
\usepackage{booktabs}
\usepackage{multirow}
\usepackage{array}
\usepackage{xcolor}
\usepackage[hidelinks]{hyperref}

\usepackage{balance}
\usepackage{float}
\newcommand{\R}{\mathbb{R}}
\newcommand{\E}{\mathbb{E}}
\renewcommand{\Pr}{\mathbb{P}}
\newcommand{\ind}{\mathbf{1}}
\newcommand{\vol}{v}                       
\newcommand{\Om}{\Omega}                   
\newcommand{\Cref}{C}                      
\newcommand{\FP}{\mathrm{FP}}
\newcommand{\TP}{\mathrm{TP}}
\newcommand{\Dice}{\mathrm{D}}
\newcommand{\val}{\mathcal{V}}             
\newcommand{\test}{\mathcal{T}}            
\newcommand{\tauval}{\hat\tau_b}           
\newcommand{\taustar}{\tau^{\star}_b}      
\newcommand{\cm}{\,\mathrm{cm}^3}
\newcommand{\mm}{\,\mathrm{mm}^3}
\newcommand{\pH}{p_{\mathrm{H}}}
\newcommand{\ci}[2]{\,[#1,\,#2]}
\newcommand{\argmax}{\operatorname*{arg\,max}}

\newcommand{\AUROC}{\mathrm{AUROC}}

\newcommand{\AROC}{\mathrm{AROC}}

\spnewtheorem{identity}{Identity}{\bfseries}{\itshape}
\begin{document}
\title{MIRTO: a registration-gated, multiverse-tested evaluation protocol for unsupervised anomaly segmentation in brain MRI}
\titlerunning{MIRTO}
\author{Negin Kafee Hernashki \inst{1,2}\orcidID{0009-0004-9006-5234} \and
Soumick Chatterjee\inst{3,4}\orcidID{0000-0001-7594-1188}}
\authorrunning{N. K. Hernashki and S. Chatterjee}
\institute{Università degli Studi di Milano, Milan, Italy \and
Politecnico di Milano, Milan, Italy \and
Human Technopole, Milan, Italy \and
Otto von Guericke University Magdeburg, Germany\\
\email{negin.kafeehernashki@studenti.unimi.it ; contact@soumick.com}}
\maketitle
\begin{abstract}
Unsupervised anomaly detection (UAD) methods for brain MRI are ranked by a single score, yet that score rests on choices that are rarely reported: how each anomaly map is aligned with the reference, how and on which data the threshold is set, and which false-positive budget, metric, aggregation and lesion definition are used. We present MIRTO, an evaluation protocol that makes these choices explicit and measures their effect. It gates the geometry of every comparison with a registration check and label-free diagnostics of known power, sets thresholds on validation data alone and reports the false-positive volume actually realised on test, repeats each comparison over 15,552 defensible evaluation pipelines, and attaches paired subject-bootstrap intervals with multiplicity control. Applied to four UAD methods trained on the same healthy data and tested on 312 BraTS 2020 subjects, MIRTO showed that an axis-order mismatch between stored maps and the reference lowered a diffusion model's voxel AUROC from 0.873 to 0.583 whilst barely moving its slice-level AUROC. Within each metric, the method explained at least 0.95 of the variance in voxel AUROC and AUPRC and 0.77 in Dice, but only 0.14 in lesion sensitivity, where the lesion definition and hit criterion dominated. A Dice advantage that was significant at validation thresholds vanished at equal realised false-positive burden, and an exact identity attributes it to threshold transfer. A training-free change to REFLECT's latent aggregation raised Dice at equal burden by 0.052. Nine hypotheses were tested against explicit criteria; because the same cohort served to develop the protocol, all inference is exploratory.

\end{abstract}

\section{Introduction}\label{sec:intro}

Unsupervised anomaly detection (UAD) in brain MRI learns what healthy anatomy looks like and flags whatever departs from it, so that pathology can be localised without lesion annotations~\cite{baur2021autoencoders,lagogiannis2024deepdive}. The field has moved quickly, from variational and context-encoding autoencoders, including our own StRegA~\cite{baur2021autoencoders,chatterjee2022strega}, to diffusion models conditioned on the input~\cite{behrendt2025cddpm,patricio2024ucad,bercea2023autoddpm,bercea2024thor}, rectified flows that correct an image in one step~\cite{beizaee2025reflect}, and frozen foundation-model features~\cite{damm2025anomalydino,guo2025dinomaly}. New methods are usually introduced with a table in which they rank first.

Such a table is the end of a long chain of decisions, most of which a reader never sees. Every anomaly map lives on the grid of its own preprocessing and must be brought into the space of a common reference; a threshold has to be chosen, on some data, to meet some false-positive budget; Dice can be averaged over subjects, taken as a median, or pooled over voxels; a lesion must be defined before it can be counted as found; and the models were trained on particular healthy data with particular budgets. Each decision is defensible on its own. Together they span a large space of analysis paths, and it is well documented outside imaging that a reasonable change of path can change the conclusion~\cite{steegen2016multiverse,silberzahn2018manyanalysts,simonsohn2020specification}. In biomedical image analysis, rankings of challenge entries are known to be fragile to the metric and the aggregation~\cite{maierhein2018rankings,reinke2024pitfalls}, and choices made with sight of the test data bias the estimate of performance~\cite{varma2006bias,cawley2010overfitting}. UAD adds failure modes of its own. Thresholds are often tuned on the test cohort; slice-level detection scores reward any map that tracks how much brain a slice contains; and simply thresholding FLAIR intensity has been reported to give better Dice than several learnt models~\cite{meissen2022challenging}.

\begin{figure*}[t]
\centering
\includegraphics[width=\textwidth]{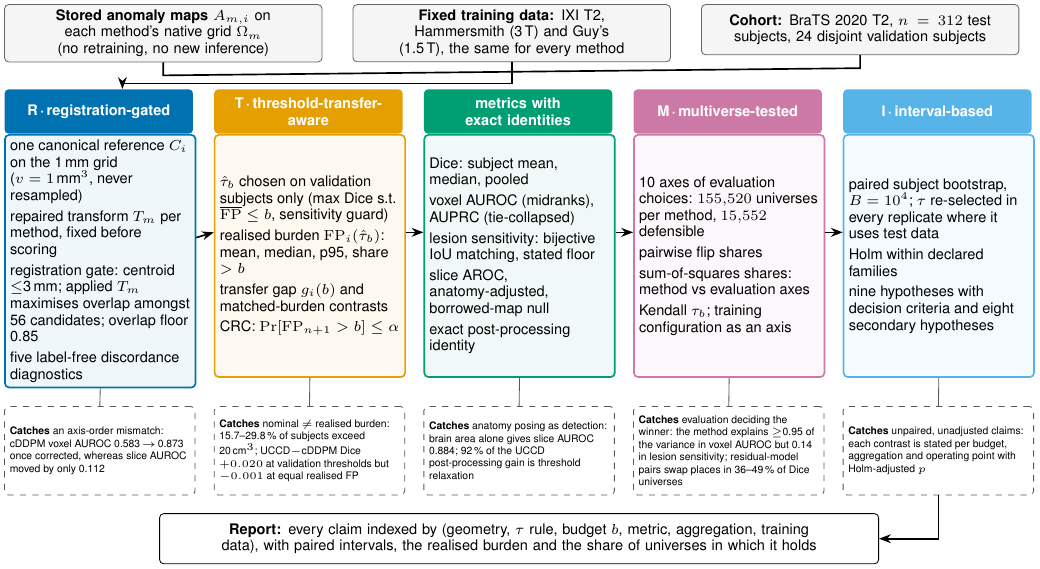}
\caption{\textbf{The MIRTO protocol.} Frozen anomaly maps from every method enter four components. \textbf{R}: one canonical reference and a repaired, gated transform per method. \textbf{T}: thresholds chosen on validation data alone, with the realised (not the nominal) false-positive burden, the threshold-transfer gap and a conformal per-scan guarantee. Exact identities accompany the metrics and explain how they relate. \textbf{M}: a multiverse of defensible evaluation pipelines. \textbf{I}: paired subject-bootstrap intervals with Holm adjustment inside declared families. Dashed boxes give, for each component, a failure it caught in this study (numbers from Section~\ref{sec:results}).}
\label{fig:overview}
\end{figure*}

We asked two questions, and framed each component of the answer as a hypothesis to be tested. How much of a UAD leaderboard is decided by the evaluation rather than by the model? And can a protocol make that dependence visible, catch the errors it hides, and still say something definite about the models? Our answer is MIRTO (\emph{M}ultiverse-tested, \emph{I}nterval-based, \emph{R}egistration-gated, \emph{T}hreshold-transfer-aware evaluati\emph{O}n; Fig.~\ref{fig:overview}). Where possible, each of its four components was checked on cases in which the right answer is known (synthetic volumes, injected corruptions, random calibration splits), and each changed at least one conclusion when applied to four current UAD families trained on the same healthy data and tested on 312 BraTS 2020 subjects.

\subsection{Related work}\label{sec:related}

\paragraph{UAD methods for brain MRI.}
Reconstruction-based methods train a generative model on healthy scans and read anomalies from the residual between an input and its pseudo-healthy reconstruction. Autoencoders and their variants were surveyed by Baur et al.~\cite{baur2021autoencoders}. In our earlier work, StRegA, we combined a compact context-encoding VAE with pre- and post-processing steps chosen for clinical data, and reported a Dice of 0.642 on BraTS T2 tumours against 0.522 for the best baseline~\cite{chatterjee2022strega}. Denoising diffusion probabilistic models~\cite{ho2020ddpm} improved reconstruction fidelity but tend either to copy the anomaly or to hallucinate healthy tissue that differs from the subject. Patched diffusion~\cite{behrendt2023pddpm}, automatic masking and re-sampling (AutoDDPM)~\cite{bercea2023autoddpm}, implicit guidance (THOR)~\cite{bercea2024thor}, masked diffusion (MAD-AD)~\cite{beizaee2025madad}, cold-diffusion restorations (DISYRE)~\cite{naval2024disyre} and aggregated normative diffusion (ANDi)~\cite{frotscher2023andi} each address this dilemma differently. Kascenas et al. showed that the spatial scale and magnitude of the training noise largely determine how well denoising autoencoders and diffusion models detect anomalies, and that a denoising autoencoder trained with coarse noise is competitive with far more complex models~\cite{kascenas2023noise}. Two lines condition the denoiser on the input itself: cDDPM conditions on a learnt encoding of the image so that the reconstruction keeps the subject's intensity characteristics~\cite{behrendt2025cddpm}, and UCAD-CDM (here UCCD) conditions on a contrastive encoder trained to be invariant to the anomaly factor~\cite{patricio2024ucad}. REFLECT replaces iterative sampling by a rectified flow~\cite{liu2023rectified} in the latent space of a VAE, transporting an abnormal latent to the healthy manifold in essentially one Euler step~\cite{beizaee2025reflect}. Feature-based methods compare deep features with those of healthy references: AnomalyDINO uses frozen DINOv2~\cite{oquab2024dinov2} patch features and nearest-neighbour distances to a memory bank~\cite{damm2025anomalydino}; Dinomaly and Dinomaly2 reconstruct foundation-model features~\cite{guo2025dinomaly,guo2025dinomaly2}; SCRD4AD uses scale-aware contrastive reverse distillation~\cite{li2025scrd}. IterMask3D refines masks iteratively in 3D at test time~\cite{liang2025itermask3d}. We study REFLECT, UCCD, cDDPM and AnomalyDINO, which cover one-step transport, two forms of conditioned diffusion and a training-free foundation-model memory bank.

\paragraph{How UAD is evaluated.}
Comparative studies have repeatedly found evaluation to be the weak link. Meissen et al. showed that thresholding FLAIR scans gives better Dice and precision-recall curves than several learnt anomaly detectors on tumour and multiple-sclerosis data~\cite{meissen2022challenging}. Lagogiannis et al. re-ran a diverse set of methods in a common framework on several datasets and found that feature-modelling methods, many from industrial inspection, set the state of the art~\cite{lagogiannis2024deepdive}. Two broad benchmarks unify methods across medical domains: BMAD assembles six datasets from five domains, amongst them brain MRI, and fifteen algorithms~\cite{bao2024bmad}, and MedIAnomaly compares thirty methods on seven datasets of five modalities, for both image-level classification and pixel-level segmentation~\cite{cai2025medianomaly}. Bercea et al. argued that the quality of the normative reconstruction, which the usual metrics ignore, governs downstream behaviour~\cite{bercea2025normative}. A recent multi-centre benchmark selected thresholds on separate validation scans, reported Dice from 0.03 to 0.65 across algorithms and found biases with scanner, lesion size, age and sex~\cite{frotscher2025benchmark}. Myles et al. recast UAD as a Bayesian inverse problem and derived several existing scores as instances of one energy~\cite{myles2026principled}. These works compare methods under a fixed protocol, or improve the scoring rule; none quantifies how much of the resulting ranking the protocol itself determines, nor checks the geometric correctness of the comparison.

\paragraph{Validation methodology.}
Maier-Hein et al. showed that challenge rankings change with the metric and the aggregation~\cite{maierhein2018rankings}, and the Metrics Reloaded consortium catalogued metric pitfalls and recommendations~\cite{maierhein2024metrics,reinke2024pitfalls}, including the dependence of overlap metrics on structure size~\cite{taha2015metrics}. Multiverse analysis~\cite{steegen2016multiverse} and specification curves~\cite{simonsohn2020specification} evaluate a conclusion over all defensible analysis paths; the many-analysts design shows how widely reasonable paths diverge~\cite{silberzahn2018manyanalysts}. Selection on the evaluation data inflates estimates~\cite{varma2006bias,cawley2010overfitting}. Across medical imaging, Varoquaux and Cheplygina traced how biases enter at every step from data to evaluation~\cite{varoquaux2022failures}; Christodoulou et al. found that more than half of the MICCAI 2023 segmentation papers did not assess performance variability and that only one reported confidence intervals~\cite{christodoulou2024ci}; and a Bayesian analysis of published comparisons estimated a probability of a false outperformance claim above 5\% for 86\% of classification and 53\% of segmentation papers~\cite{christodoulou2025falsepromises}. In machine learning generally, data sampling, initialisation and hyperparameter choice add enough variance to change benchmark conclusions~\cite{bouthillier2021variance}. MIRTO brings these ideas to UAD and adds what segmentation by thresholding needs: an explicit operating point.

\paragraph{Distribution-free control of operating points.}
Conformal risk control (CRC) chooses a threshold on calibration data so that the expected value of a bounded, monotone loss on a new exchangeable case is at most $\alpha$~\cite{angelopoulos2024crc}; risk-controlling prediction sets (RCPS)~\cite{bates2021rcps} and Learn-then-Test~\cite{angelopoulos2025ltt} give control with high probability over the calibration set instead; see~\cite{angelopoulos2021gentle} for an introduction. We use the loss ``this scan exceeds its false-positive budget''. For such an indicator loss CRC reduces to split-conformal calibration, whose coverage conditional on the calibration set follows a known Beta law~\cite{vovk2012conditional,angelopoulos2021gentle}; we use that law to state how often a small calibration set misses its target. Covariate-adjusted ROC analysis~\cite{janes2009aroc} provides the tool we use to separate anatomy from detection in slice-level scores.

\subsection{Contributions}\label{sec:contrib}

Our contribution is an evaluation protocol, the exact accounting that makes what it reports interpretable, and one method-level improvement evaluated with it; the experiments validate the protocol and apply it.
\begin{enumerate}
\item \textbf{Registration gate and label-free diagnostics.} A gate that accepts correct registrations and rejects flips, transposes and two-slice offsets, and five discordance diagnostics computable without extra labels, two of which detect injected in-plane errors with power $1.000$ at a false-alarm rate of $0.05$ in 30 subjects (Section~\ref{sec:R}).
\item \textbf{Operating points that account for threshold transfer.} Validation-only thresholds, realised rather than nominal burden, a threshold-transfer gap with an exact decomposition of any contrast into a map term and a transfer term (Identity~\ref{prop:transfer}), and a conformal per-scan burden guarantee, for which we apply the known training-conditional law of split-conformal calibration to state, from the calibration size alone, how often a calibration set misses its target (Lemma~\ref{prop:crc}; Section~\ref{sec:T}).
\item \textbf{Exact accounting of post-processing and metrics.} Elementary identities, applied to UAD, that split a post-processing gain at equal burden into an effect at fixed threshold and a threshold relaxation, state exactly when deleting a component raises Dice (Identities~\ref{prop:purity} and~\ref{prop:m4}), and decompose pooled Dice and pooled AUROC so as to explain when subject-level and pooled rankings disagree (Identities~\ref{prop:pooleddice} and~\ref{prop:pooledauc}). Their value lies in what they let a reader attribute, not in their difficulty.
\item \textbf{An evaluation multiverse.} Ten axes of evaluation choices, flip shares read together with the size of each difference, per-metric variance shares with bootstrap intervals, and the training data as an eleventh axis (Section~\ref{sec:M}).
\item \textbf{Latent aggregation for REFLECT.} A training-free change to how REFLECT combines its latent channels, which raises Dice at equal burden in two independently trained models (Sections~\ref{sec:C1} and~\ref{sec:res-C1}).
\end{enumerate}
We apply the protocol to four UAD methods and two model-free baselines. We do not claim that any method is best, nor that the inference is confirmatory: the 312 test subjects served both to develop the protocol and to test its hypotheses, so every test is exploratory (Section~\ref{sec:stats}).

\section{Method}\label{sec:method}

\subsection{Setting and notation}\label{sec:setup}

Let $\Om\subset\mathbb{Z}^3$ be the canonical voxel grid of a subject, with physical voxel volume $\vol$ (here $\vol=1\mm$). For subject $i$ we observe a brain mask $B_i\subseteq\Om$ and a binary reference $\Cref_i\subseteq B_i$ (whole tumour). A UAD method $m$ is trained on healthy scans only and, at inference, returns an anomaly map $A_{m,i}:\Om_m\to\R$ on its own native grid $\Om_m$. The test cohort $\test$ has $n=|\test|=312$ subjects and the validation set $\val$ has $|\val|=24$; the two are disjoint. MIRTO operates on stored anomaly maps: no model is retrained, fine-tuned or re-run during evaluation.

After mapping to $\Om$ (Section~\ref{sec:R}) we normalise each map inside its evaluation support $S_i\subseteq B_i$ (the brain, or the method's field of view intersected with the brain),
\begin{equation}
s_i(x)=\frac{\tilde A_{m,i}(x)-\min_{S_i}\tilde A_{m,i}}{\max_{S_i}\tilde A_{m,i}-\min_{S_i}\tilde A_{m,i}+\epsilon},\qquad \epsilon=10^{-8},
\label{eq:norm}
\end{equation}
and threshold it on a grid $\mathcal G=\{0,0.005,\dots,1\}$:
\begin{equation}
P_i(\tau)=\Pi\big(\{x\in S_i: s_i(x)\ge\tau\}\big),
\label{eq:pred}
\end{equation}
where $\Pi$ is a post-processing operator (the identity for plain thresholding). With plain thresholding $P_i(\tau)$ is nested, $\tau<\tau'\Rightarrow P_i(\tau')\subseteq P_i(\tau)$. With $\TP_i=|P_i\cap\Cref_i|$, $p_i=|P_i|$ and $G_i=|\Cref_i|$, the false-positive burden and the Dice coefficient are
\begin{equation}
\FP_i(\tau)=\vol\,\big|P_i(\tau)\setminus \Cref_i\big|,\qquad
\Dice_i(\tau)=\frac{2\,\TP_i(\tau)}{p_i(\tau)+G_i}.
\label{eq:fp-dice}
\end{equation}
Nestedness makes $\FP_i(\tau)$ non-increasing in $\tau$, a property used repeatedly below and checked on every curve.

\subsection{Methods under evaluation}\label{sec:baselines}

All four models were trained with the authors' public implementations, adapted only to read our data; architectures and losses were left unchanged. We summarise each briefly to fix the notation of its anomaly map.

\paragraph{REFLECT~\cite{beizaee2025reflect}.} A KL-regularised autoencoder~\cite{rombach2022ldm} with $8\times$ spatial down-sampling and four latent channels maps a slice $x$ to $z_0=\mathcal E(x)$. A velocity field $v_\theta$ is trained by rectified flow~\cite{liu2023rectified} on pairs $(z^{\mathrm a},z^{\mathrm h})$ of synthetically corrupted and healthy latents, minimising
\begin{equation}
\mathcal L_{\mathrm{RF}}(\theta)=\E_{t\sim U[0,1]}\,\big\|\,(z^{\mathrm h}-z^{\mathrm a})-v_\theta\big(t z^{\mathrm h}+(1-t)z^{\mathrm a},\,t\big)\big\|_2^2,
\label{eq:rf}
\end{equation}
followed by a re-flow stage that straightens the trajectories. At inference a single Euler step gives the corrected latent $z_1=z_0+v_\theta(z_0,0)$ and the corrected image $\hat x=\mathcal D(z_1)$. The published map fuses an image residual $|x-\hat x|$ and a latent residual $|z_1-z_0|$ (Section~\ref{sec:C1}).

\paragraph{cDDPM~\cite{behrendt2025cddpm}.} A denoising diffusion model~\cite{ho2020ddpm} with forward process $q(x_t\mid x_0)=\mathcal N\big(\sqrt{\bar\alpha_t}\,x_0,(1-\bar\alpha_t)I\big)$ whose denoiser is conditioned on an encoding $c=E_\phi(x)$ of the input through feature-wise affine modulation~\cite{perez2018film}, $h\mapsto\gamma(c)\odot h+\beta(c)$. Training minimises
\begin{equation}
\mathcal L_{\mathrm{DM}}=\E_{x_0,t,\varepsilon}\big\|\varepsilon-\varepsilon_\theta\big(\sqrt{\bar\alpha_t}x_0+\sqrt{1-\bar\alpha_t}\,\varepsilon,\;t,\;c\big)\big\|_2^2 .
\label{eq:ddpm}
\end{equation}
The input is partially noised and denoised; the map is the residual $|x-\hat x_0|$. Conditioning lets the reconstruction keep the subject's intensity profile without copying the anomaly.

\paragraph{UCCD (UCAD-CDM)~\cite{patricio2024ucad}.} The same residual principle, but the condition comes from a self-supervised contrastive encoder trained on healthy slices to be invariant to the factor that distinguishes a target from its background; the diffusion model is trained with~\eqref{eq:ddpm}.

\paragraph{AnomalyDINO~\cite{damm2025anomalydino}.} Frozen DINOv2 ViT-S/14 patch features $f(\cdot)$~\cite{oquab2024dinov2} of healthy reference slices form a memory bank $\mathcal M$. A test patch $u$ scores
\begin{equation}
a(u)=\min_{g\in\mathcal M}\Big(1-\frac{\langle f(u),g\rangle}{\|f(u)\|\,\|g\|}\Big),
\label{eq:dino}
\end{equation}
the cosine distance to its nearest healthy neighbour. The method is training-free.

\paragraph{Model-free baselines.} The z-scored intensity $s(x)=(I(x)-\mu_{B})/\sigma_{B}$ of T2 or FLAIR inside the brain mask, scored through the same evaluator. The models see T2 only, so FLAIR is a cross-contrast reference for what one intensity channel carries, not a fair competitor.

\subsection{Registration gating}\label{sec:R}

\paragraph{Canonical reference and physical units.} Each subject has exactly one reference $\Cref_i$, the original whole-tumour mask on the canonical grid, never resampled, cropped or masked per method, and identical for every method. The voxel volume is fixed from the dataset specification and the array shape, not from file headers, which had been altered in our copies of the data (Appendix~\ref{app:units}). Every burden in this paper is therefore on the canonical grid.

\begin{figure*}[t]
\centering
\includegraphics[width=\textwidth]{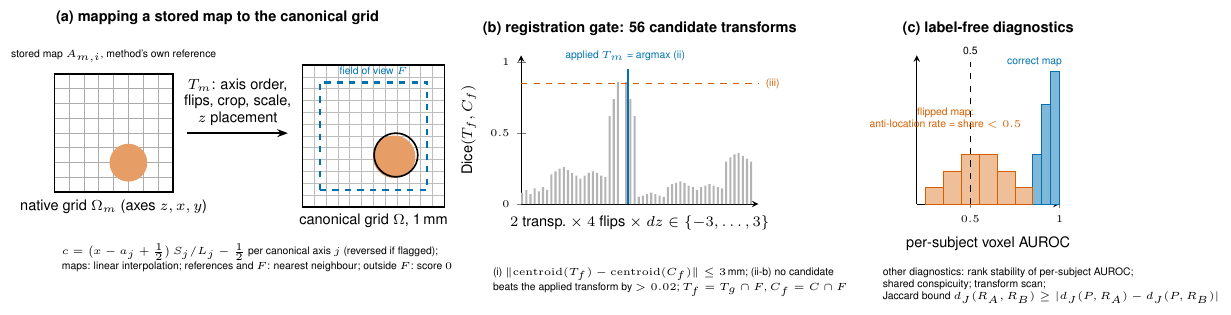}
\caption{\textbf{Registration gating (schematic).} (a) Each stored map is mapped to the canonical grid by a per-method transform $T_m$ (Eq.~\ref{eq:transform}); voxels outside the field of view $F$ score zero. (b) The registration gate compares the applied transform with the other candidate transforms on the overlap between the method's own reference, mapped, and the canonical reference inside $F$ (Eq.~\ref{eq:gate}). (c) Label-free diagnostics: an in-plane error pushes per-subject voxel AUROC below $0.5$ (anti-location rate) and scrambles its ranks (rank stability). Bars are illustrative, not data.}
\label{fig:gate}
\end{figure*}

\paragraph{Mapping.} For method $m$ a transform $T_m$ lists, for each canonical axis $j$, the source axis, a start $a_j$, an extent $L_j$, a source size $S_j$ and a reversal flag. A canonical coordinate $x_j$ maps to the source coordinate
\begin{equation}
c_j=\big(x_j-a_j+\tfrac12\big)\frac{S_j}{L_j}-\tfrac12 ,
\label{eq:transform}
\end{equation}
reflected if flagged and clipped to the source; maps are interpolated linearly, references and the field of view $F$ by nearest neighbour, and $\tilde A_{m,i}=(A_{m,i}\circ T_m)\,\ind_F$. Voxels outside $F$ score zero and are counted as negatives. $T_m$ was set once per method from ten subjects, without reference to any score (Fig.~\ref{fig:gate}a).

\paragraph{Registration gate.} Let $T_g$ be the method's own reference (the label resampled by its preprocessing) mapped by $T_m$, $T_f=T_g\cap F$ and $C_f=\Cref\cap F$. Let $\Theta$ be the 56 candidate transforms obtained from two transpositions, four in-plane flip patterns and through-plane offsets $dz\in\{-3,\dots,3\}$ (the applied transform is one of them). A subject passes if
\begin{align}
&\text{(i)}\;\;\|\mathrm{cen}(T_f)-\mathrm{cen}(C_f)\|_2\le 3\,\mathrm{mm},\nonumber\\
&\text{(ii-b)}\;\max_{\theta\in\Theta}\Dice(T_f^{\theta},C_f)-\Dice(T_f,C_f)\le 0.02,\label{eq:gate}\\
&\text{(iii)}\;\Dice(T_f,C_f)\ge 0.85,\nonumber
\end{align}
and its method passes (ii-a), that is, the applied transform maximises the cohort mean of $\Dice(T_f^{\theta},C_f)$ over $\Theta$ (Fig.~\ref{fig:gate}b). We do not use the more obvious rule of a fixed overlap threshold (centroid within 3\,mm, at least 95\% of the tumour inside the field of view and $\Dice\ge0.95$), because its last two clauses measure label fidelity and field-of-view truncation rather than registration: it would fail between 61 and 242 correctly registered subjects per method (Section~\ref{sec:res-R}). On synthetic volumes the gate passes the identity, a field of view that removes $40\%$ of a sphere, a one-voxel erosion and a $20\%$ dilation, and fails an in-plane flip, a transpose and a two-slice offset.

\paragraph{Label-free diagnostics.} Five quantities flag discordance from maps and references alone. \emph{Rank stability} is the Spearman correlation of per-subject voxel AUROC between two conditions that should describe the same tumour. \emph{Shared conspicuity} is the cross-method correlation of per-subject AUROC, since how conspicuous a tumour is belongs to the subject. The \emph{anti-location rate} is the share of subjects with AUROC $<0.5$. The \emph{transform scan} is the best mean AUROC over $\Theta$ minus that of the applied transform. The \emph{Jaccard bound} uses the fact that the Jaccard distance $d_J(X,Y)=1-|X\cap Y|/|X\cup Y|$ is a metric, so that for any prediction $P$
\begin{equation}
d_J(R_A,R_B)\ \ge\ \big|\,d_J(P,R_A)-d_J(P,R_B)\,\big| ,
\label{eq:d5}
\end{equation}
and a violation certifies that two references $R_A,R_B$ disagree. The power of rank stability and of the anti-location rate was measured by injecting known corruptions into two independently verified pipelines and setting each decision threshold at the 5th or 95th percentile, respectively, of its bootstrap distribution without corruption, which fixes the false-alarm rate at $0.05$ by construction.

\subsection{Operating points and threshold transfer}\label{sec:T}

\paragraph{Validation-only threshold.} For a nominal budget $b\in\{5,10,20,40\}\cm$, let $\overline{\FP}_{\val}(\tau)$, $\bar\Dice_{\val}(\tau)$ and $\mathrm{Sens}_{\val}(\tau)$ be the validation mean burden, mean Dice and above-floor lesion sensitivity (Section~\ref{sec:metrics}). With the feasible set $\mathcal F_b=\{\tau\in\mathcal G:\overline{\FP}_{\val}(\tau)\le b\}$,
\begin{equation}
\begin{aligned}
\tauval&=\max\Big\{\argmax_{\tau\in\mathcal F'_b}\bar\Dice_{\val}(\tau)\Big\},\\
\mathcal F'_b&=\Big\{\tau\in\mathcal F_b:\mathrm{Sens}_{\val}(\tau)\ge\max_{\tau'\in\mathcal F_b}\mathrm{Sens}_{\val}(\tau')-0.03\Big\}.
\end{aligned}
\label{eq:valtau}
\end{equation}
The 3-point guard keeps sensitivity within about one lesion of the feasible best (36 above-floor validation lesions); ties go to the larger $\tau$. Sensitivity is not monotone in $\tau$, so the feasible set is searched rather than assumed to be an interval.

\paragraph{Realised burden.} The nominal $b$ names how $\tauval$ was chosen; it is not an operating point. Every table reports the realised $\FP_i(\tauval)$ on test subjects: mean with interval, median, 95th percentile, maximum and the share of subjects with $\FP_i>b$.

\paragraph{Test-matched threshold and the transfer gap.} As a diagnostic of map quality at equal burden, and never as a result that could be used in practice, we also use the threshold matched on the test cohort,
\begin{equation}
\taustar=\min\big\{\tau\in\mathcal G:\overline{\FP}_{\test}(\tau)\le b\big\},
\label{eq:tm}
\end{equation}
and its interpolated version that sets the test mean burden to exactly $b$. Both are re-selected inside every bootstrap replicate. The per-subject threshold-transfer gap is
\begin{equation}
g_i(b)=\Dice_i(\taustar)-\Dice_i(\tauval).
\label{eq:gap}
\end{equation}
(The quantity $\bar g$ is sometimes called the ``regret'' of test-set tuning. We avoid the term, because the gap is not purely the optimism of tuning on the test set, as the next identity shows.)

Write $\beta(\tau)=\overline{\FP}_{\test}(\tau)$ and $\bar\Dice(\tau)$ for the test means. Because $\beta$ is non-increasing, on any interval where it is strictly decreasing the curve $\Psi=\bar\Dice\circ\beta^{-1}$ (mean Dice as a function of mean realised burden) is well defined.

\begin{figure*}[t]
\centering
\includegraphics[width=0.97\textwidth]{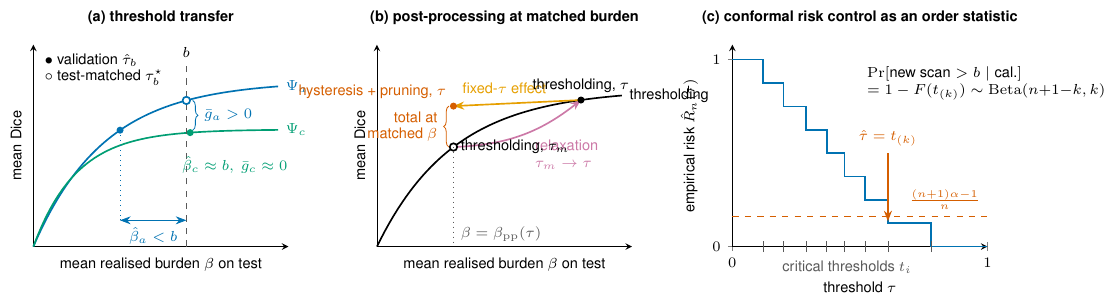}
\caption{\textbf{Operating-point geometry (schematic).} (a) Each method traces mean Dice against mean realised burden on test, $\Psi_m$. The validation threshold lands at a realised burden $\hat\beta_m$ that need not equal the nominal $b$; the transfer gap $\bar g_m$ is the rise of $\Psi_m$ between $\hat\beta_m$ and $b$ (Identity~\ref{prop:transfer}). (b) A post-processed map at threshold $\tau$ realises a smaller burden than plain thresholding at the same $\tau$; its gain over plain thresholding at equal burden splits exactly into an effect at fixed $\tau$ and a relaxation of the plain threshold from $\tau_m$ to $\tau$ (Identity~\ref{prop:m4}). (c) For the loss $\ind[\FP>b]$, CRC selects an order statistic of the per-subject critical thresholds, which yields the known Beta law of Lemma~\ref{prop:crc}.}
\label{fig:oppoints}
\end{figure*}

\begin{identity}[Transfer decomposition]\label{prop:transfer}
For methods $a$ and $c$ at budget $b$, let $\Delta^{\mathrm{val}}=\bar\Dice_a(\hat\tau_{a,b})-\bar\Dice_c(\hat\tau_{c,b})$ and $\Delta^{\star}=\bar\Dice_a(\tau^\star_{a,b})-\bar\Dice_c(\tau^\star_{c,b})$. Then exactly
\begin{equation}
\Delta^{\mathrm{val}}=\Delta^{\star}-\big(\bar g_a(b)-\bar g_c(b)\big),
\label{eq:transfer}
\end{equation}
and, if $\Psi_m$ is differentiable between the realised burden $\hat\beta_m=\beta_m(\hat\tau_{m,b})$ and $\beta_m(\tau^\star_{m,b})\approx b$,
\begin{equation}
\bar g_m(b)=\Psi_m(b)-\Psi_m(\hat\beta_m)=\Psi_m'(\xi_m)\,(b-\hat\beta_m)
\label{eq:gapmvt}
\end{equation}
for some $\xi_m$ between $\hat\beta_m$ and $b$.
\end{identity}
The identity needs only two lines of algebra and the mean-value theorem (Appendix~\ref{app:proofs}); its use is in what it lets a reader attribute. Equation~\eqref{eq:transfer} says that a contrast at validation thresholds is the contrast at equal burden minus a difference of transfer gaps; Eq.~\eqref{eq:gapmvt} says that each gap is a burden mismatch multiplied by a local slope. A method whose validation threshold happens to realise its budget on test has $\bar g\approx0$ whatever its map quality, and a method with a conservative threshold ($\hat\beta<b$) on a steep part of its curve has a large positive gap (Fig.~\ref{fig:oppoints}a). The decomposition is exact for subject means; for medians and pooled values it holds only approximately.

\paragraph{Conformal per-scan burden guarantee.} Nominal budgets constrain a cohort mean; a clinician cares about the scan in front of them. Define the bounded, non-increasing loss $L_i(\tau)=\ind[\FP_i(\tau)>b]$ and the empirical risk $\hat R_n(\tau)=n^{-1}\sum_{i\in\val}L_i(\tau)$ on $n=|\val|$ calibration subjects. CRC~\cite{angelopoulos2024crc} picks
\begin{equation}
\hat\tau=\inf\Big\{\tau:\ \frac{n}{n+1}\hat R_n(\tau)+\frac{1}{n+1}\le\alpha\Big\},
\label{eq:crc}
\end{equation}
which guarantees $\E[L_{n+1}(\hat\tau)]\le\alpha$ under exchangeability: the probability that a new scan exceeds $b$ is at most $\alpha$. The rule is feasible only if $\alpha\ge1/(n+1)$, i.e.\ $\alpha\ge0.04$ for $n=24$. We also evaluate RCPS~\cite{bates2021rcps} with a binomial upper bound at $\delta=0.10$, and a label-free CRC that controls predicted volume rather than false positives, each at $\alpha\in\{0.05,0.10,0.20\}$; RCPS at $\alpha=0.05$ is infeasible with $n=24$, which leaves eight calibrated rules. The CRC guarantee is marginal over calibration sets. Its behaviour conditional on one calibration set follows from a known result, which we restate for this loss.

\begin{lemma}[Exceedance law of CRC for an indicator loss]\label{prop:crc}
Let $t_i=\inf\{\tau:\FP_i(\tau)\le b\}$ be subject $i$'s critical threshold, so that $L_i(\tau)=\ind[t_i>\tau]$, and suppose the $t_i$ are i.i.d.\ with continuous distribution function $F$. Let $j=\lfloor(n+1)\alpha-1\rfloor$ and $k=n-j$. Then, for $j\ge0$ and no ties, $\hat\tau=t_{(k)}$, the $k$-th order statistic of the calibration $t_i$, and the exceedance probability conditional on the calibration set,
\begin{equation}
\begin{aligned}
\rho(\hat\tau)&=\Pr\big[\FP_{n+1}(\hat\tau)>b\,\big|\,t_{1:n}\big]\\
&=1-F\big(t_{(k)}\big)\ \sim\ \mathrm{Beta}(n+1-k,\;k).
\end{aligned}
\label{eq:crcbeta}
\end{equation}
In particular $\E[\rho]=(n+1-k)/(n+1)\le\alpha$.
\end{lemma}
Lemma~\ref{prop:crc} is not new. With an indicator loss, CRC selects an order statistic of the calibration scores exactly as split-conformal prediction does, and $1-\rho$ is then the coverage conditional on the calibration set, whose Beta law is the classical training-conditional result of the conformal literature~\cite{vovk2012conditional,angelopoulos2021gentle}. We state it because, applied to the per-scan burden, it turns a marginal guarantee into a prediction that a user can make before seeing any test data. For $n=24$ and $\alpha=0.10$, $j=1$ and $\rho\sim\mathrm{Beta}(2,23)$, whose mean is $0.080$, whose 97.5th percentile is $0.211$, and which puts only $70.8\%$ of its mass at or below $\alpha$. CRC therefore controls the average over calibration sets whilst leaving roughly three calibration sets in ten above the target; a guarantee that holds with high probability for the calibration set in hand requires RCPS or Learn-then-Test~\cite{bates2021rcps,angelopoulos2025ltt}, at the price of a stricter threshold. We compare this prediction with 1,000 random calibration splits in Section~\ref{sec:res-T}. Figure~\ref{fig:oppoints}c illustrates the construction.

\subsection{Metrics and their identities}\label{sec:metrics}

\paragraph{Dice and its aggregations.} We report the subject mean $\bar\Dice=n^{-1}\sum_i\Dice_i$, the subject median and the pooled value $\Dice^{\mathrm{pool}}=2\sum_i\TP_i/\sum_i(p_i+G_i)$. The reference estimand is the subject mean at $b=20\cm$, validation threshold, plain thresholding, canonical reference and full support.

\begin{identity}[Pooled Dice is a size-weighted mean]\label{prop:pooleddice}
$\Dice^{\mathrm{pool}}=\sum_i w_i\Dice_i$ with $w_i=(p_i+G_i)/\sum_k(p_k+G_k)$.
\end{identity}
The identity follows from $\TP_i=\tfrac12\Dice_i(p_i+G_i)$. Since $\sum_i(w_i-1/n)=0$,
\begin{equation}
\Dice^{\mathrm{pool}}-\bar\Dice=\sum_i\Big(w_i-\frac1n\Big)\big(\Dice_i-\bar\Dice\big)=n\,\widehat{\mathrm{Cov}}(w,\Dice),
\label{eq:poolcov}
\end{equation}
where $\widehat{\mathrm{Cov}}$ is the empirical covariance over subjects with divisor $n$. Pooling therefore helps exactly those methods whose Dice is positively associated with $p_i+G_i$. Because $p_i$ tends to grow with $G_i$, pooling gives large tumours most of the weight, and a method that segments large tumours well but small ones poorly gains from pooling relative to the subject mean. This is the mechanism behind several aggregation reversals in Section~\ref{sec:res-lead}.

\paragraph{Voxel AUROC and AUPRC.} Per subject, inside the support, AUROC is the Mann--Whitney statistic~\cite{mann1947,hanley1982auc} with mid-ranks,
\begin{equation}
\begin{aligned}
\AUROC_i=\frac{1}{n_i^+n_i^-}\sum_{x\in\Cref_i}\;\sum_{y\in S_i\setminus\Cref_i}\Big(&\ind[s(x)>s(y)]\\[-4pt]
&+\tfrac12\ind[s(x)=s(y)]\Big),
\end{aligned}
\label{eq:auroc}
\end{equation}
and AUPRC is average precision with tied scores collapsed to one threshold. Ties matter: out-of-view voxels set to zero form a large tie block, and ordinal ranks would resolve it by array order. Both metrics are invariant to $\tau$ and to post-processing, so they cannot support a claim about delineation. Pooled values are computed from per-subject histograms of the normalised score with 4,096 bins.

\begin{identity}[Pooled AUROC mixes within- and between-subject comparisons]\label{prop:pooledauc}
With $N^\pm=\sum_i n_i^\pm$ and $A_{ij}$ the probability that a positive voxel of subject $i$ outscores a negative voxel of subject $j$ (ties counted one half),
\begin{equation}
\AUROC^{\mathrm{pool}}=\sum_{i}\frac{n_i^+n_i^-}{N^+N^-}\AUROC_i\;+\;\sum_{i\ne j}\frac{n_i^+n_j^-}{N^+N^-}A_{ij}.
\label{eq:pooledauc}
\end{equation}
\end{identity}
This block structure is that of the nonparametric clustered-ROC estimator of Obuchowski~\cite{obuchowski1997clustered}, in which the pooled area averages comparisons within and between clusters; here the clusters are subjects. The second term, which carries weight $1-\sum_i n_i^+n_i^-/(N^+N^-)$, rewards scores that are comparable \emph{across} subjects. A score can therefore rank voxels well inside each subject (high subject-mean AUROC) and still pool poorly if its scale drifts between subjects. We use Eq.~\eqref{eq:pooledauc} to interpret the FLAIR reversal of Section~\ref{sec:res-base}; it is an interpretation, not a test.

\paragraph{Lesions and their matching.} Reference lesions are the 6-connected components of $\Cref_i$ of at least $50\mm$; predicted components use the same connectivity. Components are matched one to one by maximum-weight bipartite matching on the IoU matrix~\cite{kuhn1955hungarian}, and a lesion counts as detected if its matched IoU is at least $0.10$. With $H$ hits amongst $N_L$ reference lesions the pooled sensitivity is $H/N_L$. Because no sub-floor fragment was hit in our data, changing the floor changes $N_L$ but not $H$, and the sensitivity scales by the ratio of denominators (Section~\ref{sec:res-detect}). We do not use ``matched Dice'' computed on the reference region alone: if $P\subseteq\Cref$, then $\Dice=2r/(1+r)$ with coverage $r=|P|/|\Cref|$, so it measures coverage only. Lesions are called small below $17{,}414.67\mm$, the first tertile of above-floor validation lesion volumes, computed on validation data only.

\paragraph{Slice-level AUROC and anatomy.} A slice is scored by its maximum normalised value and labelled positive if it contains tumour. The \emph{marginal} AUROC averages within-subject AUROCs. To separate detection from anatomy we use the covariate-adjusted ROC of Janes and Pepe~\cite{janes2009aroc}: each positive slice $j$ with covariates $z_j$ (brain area, relative position) receives the placement value
\begin{equation}
\begin{aligned}
\pi_j&=\Pr\big[s^-<s_j\,\big|\,z^-\approx z_j\big]+\tfrac12\Pr\big[s^-=s_j\,\big|\,z^-\approx z_j\big],\\
\AROC&=\frac{1}{n^+}\sum_j\pi_j ,
\end{aligned}
\label{eq:aroc}
\end{equation}
estimated amongst negative slices of the same subject by brain-area quintile strata (the primary estimator), nearest-neighbour matching with calipers of $0.2$, $0.1$ and $0.05$ standard deviations, or spline regression of negative scores on the covariates (exploratory). A scorer that sees only the covariates must obtain $\AROC=0.5$ if the adjustment works; we test this with two such scorers (brain area; closeness to mid-brain). A \emph{borrowed-map} null replaces a subject's slice-score profile by those of the 50 subjects with the closest mean brain area, aligned from the first brain slice, keeping the target's labels; the observed minus borrowed AROC isolates subject-specific information.

\subsection{Exact accounting of post-processing}\label{sec:m4}

Two post-processing schemes are compared with plain thresholding: hysteresis with seed threshold $\min(0.999,\tau+0.15)$ and grow threshold $\tau$~\cite{canny1986}, keeping each component of $\{s\ge\tau\}$ that contains a seed voxel (\emph{hysteresis}); and the same restricted to the most persistent branch of the map's superlevel-set filtration~\cite{edelsbrunner2002persistence} (\emph{hysteresis with pruning}). The seed margin is a fixed constant and was not tuned.

\begin{identity}[Purity rule]\label{prop:purity}
Let a prediction have $p$ voxels and $T$ true positives against a reference of $G$ voxels, so $\Dice=2T/(p+G)$. Deleting a predicted component of $p'$ voxels containing $t'$ true positives changes Dice by
\begin{equation}
\begin{aligned}
\Delta\Dice&=\frac{2(T-t')}{p-p'+G}-\frac{2T}{p+G},\\
\operatorname{sign}(\Delta\Dice)&=\operatorname{sign}\Big(\frac{\Dice}{2}-\frac{t'}{p'}\Big).
\end{aligned}
\label{eq:purity}
\end{equation}
Symmetrically, adding a region of purity $q$ raises Dice if and only if $q>\Dice/2$.
\end{identity}
Deleting a component helps exactly when its purity is below half the current Dice. Small true lesions inside larger false-positive clusters therefore fall to the same rule that improves Dice, which explains why pruning lowers lesion sensitivity (Section~\ref{sec:res-pp}). We verified Eq.~\eqref{eq:purity} on 299,358 deleted components (largest absolute error $5\times10^{-7}$).

\begin{identity}[Matched-burden decomposition]\label{prop:m4}
Let $\tau$ be the validation threshold of the post-processed map and $\tau_m$ the plain threshold whose test mean realised burden equals that of the post-processed map at $\tau$. Then
\begin{equation}
\begin{aligned}
\underbrace{\bar\Dice_{\mathrm{pp}}(\tau)-\bar\Dice_{\mathrm{g}}(\tau_m)}_{\text{gain at matched burden}}
&=\underbrace{\bar\Dice_{\mathrm{pp}}(\tau)-\bar\Dice_{\mathrm{g}}(\tau)}_{\text{post-processing at fixed }\tau}\\
&\quad+\underbrace{\bar\Dice_{\mathrm{g}}(\tau)-\bar\Dice_{\mathrm{g}}(\tau_m)}_{\text{threshold relaxation}} .
\end{aligned}
\label{eq:m4}
\end{equation}
\end{identity}
The identity is trivial, adding and subtracting one term; its value lies in the reading. Post-processing frees part of the burden budget, and a comparison at matched burden spends it on a lower plain threshold. The relaxation term measures what that spending buys, and it is not an effect of the post-processing step itself (Fig.~\ref{fig:oppoints}b). Each term receives its own paired bootstrap interval.

\subsection{The evaluation multiverse}\label{sec:M}

\paragraph{Axes.} The multiverse spans ten axes of evaluation choices (Table~\ref{tab:axes}): reference (canonical; the method's own reference mapped to canonical space; the canonical reference after a round trip through the method's grid), support, voxel volume, threshold protocol, burden, post-processing, lesion definition, hit criterion, aggregation and metric. Their full factorial gives $3\cdot2\cdot2\cdot5\cdot4\cdot3\cdot3\cdot4\cdot3\cdot6=155{,}520$ \emph{universes} per method. The \emph{defensible} subset fixes the canonical reference and the true voxel volume and admits only the validation, conformal and test-matched threshold protocols: $15{,}552$ universes, $2{,}592$ per metric. Axes that do not affect a metric replicate its value, which keeps the design balanced and weights metrics equally.

\begin{table}[t]
\caption{Axes of the evaluation multiverse. Levels in bold enter the defensible subset; all levels of the remaining axes do.}
\label{tab:axes}
\centering\scriptsize
\setlength{\tabcolsep}{3pt}
\begin{tabular}{@{}l>{\raggedright\arraybackslash}p{0.72\columnwidth}@{}}
\toprule
Axis & Levels\\
\midrule
reference & \textbf{canonical $\Cref$}; method's own reference, mapped; round trip via method grid\\
support & full brain; field of view $\cap$ brain\\
voxel volume & \textbf{true 1\,mm$^3$}; declared legacy value\\
threshold & \textbf{validation (Eq.~\ref{eq:valtau})}; \textbf{CRC $\alpha=0.10$}; \textbf{test-matched}; fixed $0.5$; per-subject optimum\\
burden $b$ & 5, 10, 20, 40\,cm$^3$\\
post-processing & plain thresholding; hysteresis; hysteresis with pruning\\
lesion definition & 6-conn.\ + 50\,mm$^3$ floor; 26-conn.\ + floor; 6-conn.\ no floor\\
hit criterion & IoU $\ge0.10$; $\ge0.25$; $\ge0.50$; any overlap\\
aggregation & subject mean; subject median; pooled\\
metric & Dice; voxel AUROC; voxel AUPRC; lesion sensitivity; slice AUROC marginal; slice AROC stratified\\
\bottomrule
\end{tabular}
\end{table}

\paragraph{Flip shares.} For methods $a,c$ and a universe set $\mathcal U$,
\begin{equation}
\phi_{ac}=\frac{\min\big(N_{a>c},\,N_{c>a}\big)+\tfrac12N_{a=c}}{|\mathcal U|},
\label{eq:flip}
\end{equation}
where $N_{a>c}=\#\{u\in\mathcal U:y_{u,a}>y_{u,c}\}$, $N_{a=c}$ counts ties ($|y_{u,a}-y_{u,c}|<10^{-9}$), and $y_{u,m}$ is the score of method $m$ in universe $u$. Thus $\phi_{ac}\in[0,\tfrac12]$ is the share of universes in which the pair takes its minority order, and $\phi_{\max}=\max_{a<c}\phi_{ac}$. H1 (ranking instability; Section~\ref{sec:res-M}) is supported if the 2.5th bootstrap percentile of $\phi_{\max}$ is at least $0.20$. A flip share near $\tfrac12$ arises in two different situations: when the order of two methods truly depends on evaluation choices, and when the two methods are too close to separate. We therefore never report a flip share alone but next to the size of the difference, namely the median absolute difference across universes, the median lead in each direction, and the number of paired contrasts at validation thresholds whose Holm-adjusted interval excludes zero in each direction. Only sizeable and significant leads in both directions show that the evaluation decides the order.

\paragraph{Variance decomposition.} On the balanced array $y_{u,m}$, with grand mean $\bar y$, method means $\bar y_{\cdot m}$ and, for axis $k$ with level $\ell$, cell means $\bar y_{k\ell}$ and $\bar y_{k\ell,m}$,
\begin{align}
SS_{\mathrm{method}}&=|\mathcal U|\sum_m(\bar y_{\cdot m}-\bar y)^2,\nonumber\\
SS_k&=M\,\frac{|\mathcal U|}{L_k}\sum_{\ell}(\bar y_{k\ell}-\bar y)^2,\nonumber\\
SS_{k\times m}&=\frac{|\mathcal U|}{L_k}\sum_{\ell,m}\big(\bar y_{k\ell,m}-\bar y_{k\ell}-\bar y_{\cdot m}+\bar y\big)^2,
\label{eq:ss}
\end{align}
for $M$ methods and $L_k$ levels. The axis share is $\sum_k(SS_k+SS_{k\times m})/SS_{\mathrm{tot}}$ and the method share $SS_{\mathrm{method}}/SS_{\mathrm{tot}}$; the remainder holds higher-order interactions. Pooled over metrics, the metric axis mixes quantities on different scales, and its main effect measures those scales rather than any evaluation choice. The informative decomposition is therefore the one within each metric, which we report first. Two cautions apply to it. An axis main effect shifts every method equally and cannot by itself reorder them, so the axis share bounds, but does not measure, the influence of evaluation on rankings; reordering is measured directly by within-metric flip shares. And an axis that does not affect a metric replicates its values, which inflates that metric's method share. H2 (evaluation dominance) compares the pooled shares and is supported if the axis share exceeds the method share with non-overlapping intervals.

\paragraph{Order agreement.} Kendall's $\tau_b$~\cite{kendall1945ties} compares each universe's four-method order with that of the reference universe (Dice, $20\cm$, validation $\tau$, subject mean, canonical reference, full support, plain thresholding, 6-connectivity with floor, IoU $\ge0.10$).

\paragraph{Training data as an axis.} Holding the evaluation fixed, we score every method after training on each of five healthy training sets (Table~\ref{tab:trainsets}) and add the training set as an eleventh axis, with a method $\times$ training-set term in Eq.~\eqref{eq:ss}. We also measure how stably a training set can be chosen per method on the 24 validation subjects: each of six selection rules is recomputed in $10^4$ bootstrap resamples of $\val$, and the \emph{test regret} of a choice is the test Dice of the best configuration minus that of the chosen one.

\subsection{Statistical inference and hypotheses}\label{sec:stats}

The subject is the unit of inference; voxels, slices and lesions are clustered within subjects and are never resampled on their own. For a statistic $\hat\theta=\theta(\mathcal D)$ computed on subjects $\mathcal D$, the paired bootstrap draws $\mathcal D^{*b}$ with replacement, recomputes $\hat\theta^{*b}$ for every method on the same resample, and reports the percentile interval $[\hat\theta^*_{(0.025)},\hat\theta^*_{(0.975)}]$~\cite{efron1979bootstrap}, with $B=10^4$ replicates ($B=10^3$, with the same resamples for every universe, in the multiverse). Thresholds that use test data are re-selected in every replicate; validation and conformal thresholds use no test data and stay fixed, so their intervals omit the uncertainty of the 24-subject fit. Two-sided bootstrap $p$-values use an add-one correction, $p=2\min\{(1+\#\{\hat\theta^{*}\le0\}),(1+\#\{\hat\theta^{*}\ge0\})\}/(B+1)$, whose smallest value is $2/(B+1)\approx2\times10^{-4}$. Holm's step-down adjustment~\cite{holm1979} is applied inside declared families,
\begin{equation}
\pH^{(r)}=\max_{q\le r}\min\big\{1,(K-q+1)\,p^{(q)}\big\},
\label{eq:holm}
\end{equation}
for the $r$-th smallest of $K$ raw $p$-values. The floors are therefore $0.0048$ ($K=24$), $0.0032$ ($K=16$) and $0.0064$ ($K=32$); we print $\pH<0.005$ at the floor. Families are the 24 model-pair Dice contrasts (6 pairs $\times$ 4 budgets) per threshold rule and aggregation, the 32 model-versus-baseline Dice contrasts, and the corresponding voxel-metric families. Every per-subject quantity is reported as a mean and a median, never interchangeably.

Every experiment in Section~\ref{sec:results} tests a hypothesis. Nine primary hypotheses, H1 to H9, each carry an explicit decision criterion (Table~\ref{tab:H}); eight secondary hypotheses predict specific effects that follow from the identities and from the primary analyses (Appendix~\ref{app:prereg}); and where an experiment has no decision criterion, its hypothesis is marked exploratory and judged from intervals alone. A hypothesis is called supported only if its criterion is met, and a failed criterion is reported as such. The same 312 test subjects served both to develop the protocol and to test these hypotheses, so the tests are exploratory, not confirmatory: an explicit criterion prevents a hypothesis from being bent to fit its outcome, but it cannot remove the influence of other analyses of the same cohort. Confirmation requires an independent cohort, evaluated under a pre-registered protocol in which the training set and every threshold rule are fixed. All analyses run on stored maps without a GPU; the cost of training and inference is reported in Appendix~\ref{app:compute}.

\subsection{Latent aggregation for REFLECT}\label{sec:C1}

As a methodological contribution in its own right, we propose a training-free change to how REFLECT turns its two residuals into an anomaly map. REFLECT exposes two residual streams, but its published map averages the four latent channels before fusing them with the image residual, although the channels need not agree on where an anomaly lies. Let $\mathcal S_\sigma$ be in-plane Gaussian smoothing. The image stream is
\begin{equation}
I=\frac{\mathcal S_{3\,\mathrm{px}}\big(2.5\min(|x-\hat x|,0.4)\big)}{\max\big(\mathcal S_{3\,\mathrm{px}}(2.5\min(|x-\hat x|,0.4))\big)},
\label{eq:c1img}
\end{equation}
and the latent stream has four channels $\ell_c=\mathcal S_{1}\big(|z_{1,c}-z_{0,c}|\big)$, $c=1,\dots,4$, at one-eighth resolution. An aggregation $a:\R^4\to\R$ collapses the channels and the score is
\begin{equation}
S=\frac{\tfrac12 I+\tfrac12\,\mathrm{up}_8\big(\max(a(\ell),0)\big)}{\max(\cdot)},
\label{eq:c1score}
\end{equation}
with nearest-neighbour up-sampling $\mathrm{up}_8$. We compare the mean ($a_1=\tfrac14\sum_c\ell_c$), the median absolute deviation across channels ($a_2=\mathrm{med}_c|\ell_c-\mathrm{med}_{c'}\ell_{c'}|$, a measure of channel disagreement), the Euclidean norm ($a_3=\|\ell\|_2$), the 0.99 quantile over channels, and the mean of robustly whitened channels, against the published map and three controls (image stream only; latent stream only; no clipping). Nothing is trained: each rule is a different function of quantities that REFLECT already computes. Each rule is mapped and thresholded exactly as the published map and evaluated (i) at its own validation threshold, which is the result a user would obtain, and (ii) at the burden realised by the published map, matched either over the cohort or per subject by linear interpolation in $\FP$. MAD is retained as the recommended aggregation only if its interval did not overlap that of the best other aggregation.

\subsection{Data}\label{sec:data}

\paragraph{Test and validation data.} BraTS 2020~\cite{menze2015brats,bakas2017tcga,bakas2018brats} provides multi-institutional pre-operative glioma MRI, co-registered to a common template, resampled to $1\mm$ and skull-stripped by the organisers. We evaluate on T2; FLAIR is used only by the FLAIR baseline. The reference is the whole tumour (labels 1, 2 and 4). We use 312 test subjects and 24 disjoint validation subjects, split at random. The canonical arrays are our copy of the data, cropped through-plane to remove empty slices: $240\times240$ in-plane at 1\,mm and 98 to 149 slices (median 139); cropping removes no tumour voxel. Median whole-tumour volume is $91.7\cm$ (IQR $53.2$--$147.8$; range $7.3$--$361.8$). The test reference contains $5{,}473$ 6-connected components, of which 472 reach the $50\mm$ floor (165 small, 307 large).

\paragraph{Healthy training data.} IXI~\cite{ixi} provides T1 and T2 scans from three London sites; our preprocessed T2 pool contains the Hammersmith Hospital (HH, 3\,T) and Guy's Hospital (Guys, 1.5\,T) sites. MOOD~\cite{zimmerer2022mood} provides T1 brain scans. From these we formed five training sets (Table~\ref{tab:trainsets}), named after their sources, and tested every model on BraTS T2. \textbf{All main-text results use IXI-HH+Guys for every method.} The choice rests on two grounds stated without reference to any score: IXI-HH+Guys matches the test contrast, and IXI-HH and IXI-Guys are strict subsets of it. Because all twenty method--training-set combinations were scored on the same test cohort, the choice is protected by its stated rationale, by showing that each evaluation finding holds with every set of training data (Appendix~\ref{app:config}), and by a confirmatory evaluation on a new cohort, to be pre-registered with this training set fixed. Within IXI-HH+Guys each method kept its own training protocol (REFLECT 504 volumes; UCCD 324 training and 81 validation volumes; cDDPM 361 and 92; AnomalyDINO a memory bank from 50 reference volumes), so a common training set means the same source data, not the same optimisation budget (Appendix~\ref{app:config}).

\begin{table}[t]
\caption{Healthy training sets. All models were tested on BraTS 2020 T2. IXI-HH+Guys is used for every method in the main text.}
\label{tab:trainsets}
\centering\scriptsize
\setlength{\tabcolsep}{3pt}
\begin{tabular}{@{}lll@{}}
\toprule
Name & Source & Training contrasts\\
\midrule
IXI-HH & IXI, Hammersmith (3\,T) & T2\\
IXI-Guys & IXI, Guy's (1.5\,T) & T2\\
\textbf{IXI-HH+Guys} & IXI, both sites above & T2\\
MOOD+IXI-HH & MOOD and IXI Hammersmith & T1 and T2\\
MOOD+IXI & MOOD and all of IXI & T1 and T2\\
\bottomrule
\end{tabular}
\end{table}

\section{Results}\label{sec:results}

Unless stated otherwise, all methods were trained on IXI-HH+Guys and evaluated on the 312 test subjects, intervals in brackets are 95\% paired subject-bootstrap intervals, ``Dice'' is the subject mean at the validation threshold with plain thresholding, and $b$ is a nominal budget. Each subsection states the hypothesis that the experiment tests, reports the result and closes with a verdict. Hypotheses H1 to H9 carry explicit decision criteria (Table~\ref{tab:H}); secondary hypotheses predict specific effects (Appendix Table~\ref{tab:P}); hypotheses without a decision criterion are marked exploratory and judged from intervals alone. The main text reports the findings that bear on a conclusion; the full set of analyses, at every budget, aggregation and set of training data, is in Appendix~\ref{app:further}, with one subsection for each subsection below.

\subsection{Registration and geometry}\label{sec:res-R}

\paragraph{Hypothesis.} A mapping error that permutes voxels within slices should push per-subject voxel AUROC below 0.5 and scramble its ranks across subjects, whilst leaving slice-level scores almost unchanged, because a slice that contains tumour still does so after any in-plane permutation. If so, label-free diagnostics can detect such errors without extra labels. \emph{H7 (diagnostic power):} rank stability and the anti-location rate each detect in-plane flips and transposes with power of at least 0.95 at a false-alarm rate of 0.05 in 60 subjects. A secondary hypothesis predicts that UCCD's voxel AUROC falls under antero-posterior and left-right flips whereas AnomalyDINO's slice-level AUROC does not fall under an antero-posterior flip.

\paragraph{A mapping error that standard metrics miss.} Each method stores its maps in its own axis order, crop and resolution, so any comparison of methods with different preprocessing must undo these steps, and an error in doing so is easy to make and hard to see. Such errors arose in practice in assembling this benchmark. A mapping that read the stored cDDPM maps in the wrong axis order and ignored their field-of-view crop gave a voxel AUROC of 0.583; with the correct mapping the same maps gave 0.873, a change of $+0.290\ci{0.277}{0.302}$, with $+0.535\ci{0.517}{0.553}$ in AUPRC and $+0.520\ci{0.504}{0.537}$ in Dice at the validation threshold at $20\cm$ (Appendix~\ref{app:geom}, Table~\ref{tab:repair}). Slice-max AUROC moved by only $+0.112\ci{0.099}{0.125}$ (median $+0.083$). This is the pattern the hypothesis predicts: a slice-level score is close to blind to the error that matters most for segmentation. The error is of the kind the registration gate is designed to reject (it fails in-plane flips and transposes on synthetic volumes; Section~\ref{sec:R}), and the label-free diagnostics flag it directly: under the wrong mapping 24\% of cDDPM test subjects were anti-located (per-subject voxel AUROC $<0.5$). Two further instances of the same risk occurred with REFLECT. A second export of its maps with the same kind of axis error left 26\% of test subjects anti-located and lowered voxel AUROC from 0.936 to 0.616 whilst slice AUROC stayed at 0.915 (with the correct mapping, 2 of 312 subjects were anti-located); and resampling REFLECT maps by simple zooming instead of with the mapped transform gave the published map a Dice of 0.089 instead of 0.548, which invalidated a comparison of score rules made with that resampling. Each error produced a plausible table. That is the case for gating geometry rather than assuming it.

\begin{figure*}[t]
\centering
\includegraphics[width=0.33\textwidth]{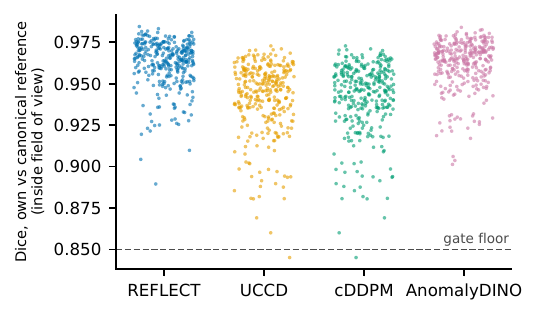}\hfill
\includegraphics[width=0.65\textwidth]{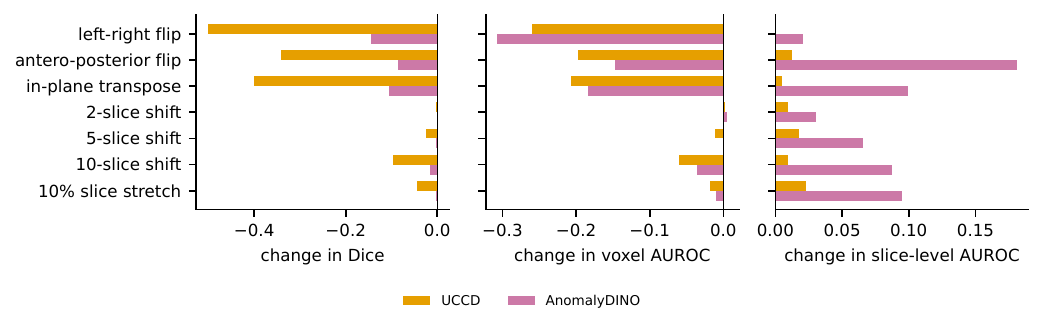}
\caption{\textbf{Registration.} (a) Per-subject overlap of each method's own reference, mapped to the canonical grid, with the canonical reference inside the field of view; the dashed line is the gate's floor (0.85). (b) Change in Dice, voxel AUROC and slice-level AUROC under injected corruptions of two verified pipelines: in-plane errors devastate the voxel metrics and leave slice-level AUROC unchanged or improved.}
\label{fig:R}
\end{figure*}

\paragraph{The registration gate on correct pipelines.} The gate passed 1,342 of 1,344 method--subject pairs (312 test and 24 validation subjects for each of four methods; Fig.~\ref{fig:R}a). Every method's applied transform was the cohort-wide overlap maximiser amongst the 56 candidates (criterion ii-a). The one subject that failed, for UCCD and cDDPM, which share a reference, failed only the overlap floor (iii) with $\Dice=0.845$; its centroid error was within 3\,mm and no alternative did better, so it is a label-fidelity exception, not a misregistration. Excluding it changed the interval-excludes-zero status of 5 of 588 contrasts, each with a bound within 0.0013 of zero, and moved no estimate by more than 0.008. A fixed overlap threshold of 0.95 would have failed 61 (REFLECT), 63 (AnomalyDINO) and 242 (UCCD, cDDPM) correctly registered subjects (Appendix Table~\ref{tab:g1}): their own references had been resampled inside the method's preprocessing (median label fidelity 0.966) or their tumour extended beyond the 100-slice field of view of UCCD and cDDPM (45 of 336 subjects). Across all twenty method--training-set combinations the same subject was the only failure.

\paragraph{Diagnostics where the truth is known.} We injected controlled corruptions into the two pipelines whose geometry had been verified independently (Fig.~\ref{fig:R}b; Appendix Table~\ref{tab:m1}). In-plane flips and the transpose cost UCCD 0.34 to 0.50 Dice and 0.20 to 0.26 voxel AUROC, yet changed its slice-max AUROC by at most $+0.012$. A corrupted map can even \emph{improve} a standard metric: an antero-posterior flip raised AnomalyDINO's slice-max AUROC by $+0.181$ and the transpose by $+0.099$. Rank stability and the anti-location rate detected every in-plane flip and transpose with power 1.000 at a false-alarm rate of 0.05 already at $n=30$, which supports H7 (Table~\ref{tab:m1power}); the secondary hypothesis on flips held as well. They behave differently elsewhere: the anti-location rate stays near zero under through-plane shifts and reference dilation, so it is specific to localisation errors, whereas rank stability reacts to almost any perturbation of the per-subject ranking (power 1.000 for a two-slice shift that moved AUROC by only $+0.003$) and is better read as a change detector. Permuting 5\% of subject identifiers was not detectable at $n\le312$, and the Jaccard bound was never violated. Further robustness checks (excluding the failing subject, restricting to tumours inside every field of view, a through-plane offset check, spatial randomisation and a geometry factorial) are reported in Appendix~\ref{app:geom}.

\subsection{Sensitivity of rankings to evaluation choices}\label{sec:res-M}

\paragraph{Hypotheses.} If evaluation choices matter, the order of two methods should depend on them. \emph{H1 (ranking instability):} some pair of methods takes its minority order in at least 20\% of defensible universes (lower 95\% bound of $\phi_{\max}$ at least 0.20). \emph{H2 (evaluation dominance):} the evaluation axes, with their interactions with method, explain more of the variance of scores than the method does, with non-overlapping intervals. Because H2 pools metrics that live on different scales, we add an exploratory hypothesis that sharpens it: a metric measures a property of the method only if, within that metric, the method dominates the variance and the order of methods does not flip.

\begin{figure*}[t]
\centering
\includegraphics[width=\textwidth]{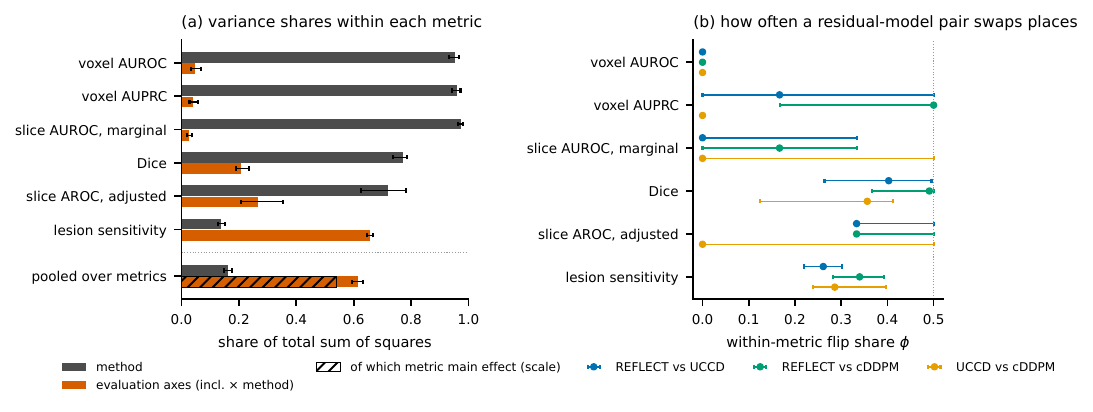}
\caption{\textbf{What decides a ranking, metric by metric.} (a) Share of the variance across defensible universes explained by the method and by the evaluation axes (main effects plus interactions with method), within each metric and pooled over metrics; the hatched bar is the main effect of the metric axis, which reflects scale. (b) Within-metric flip shares of the three residual-model pairs with 95\% bootstrap intervals; $\phi=0.5$ means that the pair is ordered each way equally often. Read (b) together with the size of each difference (Section~\ref{sec:res-M}).}
\label{fig:Mshares}
\end{figure*}

\paragraph{Within each metric.} The decomposition within a metric separates rankings that belong to the method from rankings that the evaluation decides (Table~\ref{tab:H2}, Fig.~\ref{fig:Mshares}). The method explained $0.954\ci{0.933}{0.966}$ of the variance in voxel AUROC and $0.960\ci{0.941}{0.972}$ in voxel AUPRC. Voxel AUROC gave a single order in every universe (REFLECT $>$ cDDPM $>$ UCCD $>$ AnomalyDINO); in voxel AUPRC only the aggregation changed the order, REFLECT and cDDPM exchanging places between subject-level and pooled values (Section~\ref{sec:res-lead}). The method explained $0.770\ci{0.738}{0.787}$ of the variance in Dice, but much of this is AnomalyDINO's deficit. Amongst the three residual models every pair swapped places in a substantial share of Dice universes (within-metric flip shares 0.36 to 0.49), Dice gave six different orders, and each of REFLECT, UCCD and cDDPM ranked first in between 26\% and 39\% of them (Appendix~\ref{app:ext}, Table~\ref{tab:univ}). The budget and the threshold protocol were the largest evaluation axes for Dice (0.108 and 0.056 of its variance, main effect plus interaction with method; point estimates). Lesion sensitivity was governed by the evaluation: the method explained $0.139\ci{0.127}{0.152}$ of its variance and the axes $0.657\ci{0.648}{0.666}$, of which the lesion definition carried 0.338 and the hit criterion 0.269 (point estimates). Part of the lesion-definition share is a change of denominator common to all methods (Section~\ref{sec:res-detect}), which cannot reorder them; the within-metric flip shares of 0.26 to 0.34 between residual models show that the evaluation choices reorder them as well.

\paragraph{Flip shares and the size of differences.} Over all defensible universes the largest flip share was $\phi_{\max}=0.472\ci{0.414}{0.499}$, for REFLECT versus UCCD, which meets the decision criterion of H1; REFLECT versus cDDPM had the same point value and UCCD versus cDDPM $0.393\ci{0.250}{0.493}$. AnomalyDINO almost never changed places with anyone ($\phi\le0.011$). Because a flip share near $\tfrac12$ is also what two indistinguishable methods produce, we read each one against the size of the difference (Section~\ref{sec:M}). Within Dice, UCCD led REFLECT in 59.7\% of universes by a median of 0.054 and REFLECT led in 40.3\% by a median of 0.030; the median absolute difference was 0.043, and the two were within 0.02 of each other in only 26.9\% of Dice universes. At validation thresholds, seven of the twelve REFLECT $-$ UCCD Dice contrasts (four budgets, three aggregations) favoured UCCD and one favoured REFLECT with Holm-adjusted $p<0.05$; for REFLECT $-$ cDDPM (median absolute difference 0.055) four favoured REFLECT and five cDDPM. The flips of these pairs therefore reflect the evaluation rather than closeness: each method leads by several Dice points in a large share of defensible evaluations, significantly so at some operating points, and which one leads is decided by the evaluation. UCCD versus cDDPM is different. Their median absolute Dice difference was 0.029, and although ten of the twelve contrasts favoured UCCD significantly at validation thresholds, none did at equal realised burden (Section~\ref{sec:res-T}); this pair is better described as close, with an apparent advantage that comes from threshold transfer. Only $24.3\%\ci{9.9}{32.4}$ of universes reproduced the four-method order of the reference universe (mean Kendall $\tau_b=0.486$).

\paragraph{Pooled over metrics.} Pooling all six metrics, the evaluation axes and their interactions with method explained $0.614\ci{0.595}{0.634}$ of the total sum of squares and the method $0.162\ci{0.147}{0.177}$, which meets the decision criterion of H2. We do not read this as evidence that evaluation outweighs the method: $0.539\ci{0.516}{0.561}$ is the main effect of the metric axis, which reflects the different scales of the metrics. Lesion definition ($0.014$) and hit criterion ($0.011$) followed, and every other main effect was at most $0.004$. The pattern held with every set of training data we examined (axis share 0.60 to 0.61, method share 0.16 to 0.18). All flip shares, the full decomposition and the distinct orders per metric are given in Appendix~\ref{app:multiverse}.

\paragraph{Verdict.} H1 is supported, and the size of the differences shows that the flips of REFLECT against the two conditioned diffusion models reflect the evaluation, not closeness. H2 is supported by its criterion, but the support comes from the scales of the metrics. The exploratory hypothesis holds for voxel AUROC, holds for voxel AUPRC except for the aggregation, and fails for Dice amongst the residual models and for lesion sensitivity.

\begin{table}[t]
\caption{Sum-of-squares shares across defensible universes within each metric and, last, pooled over metrics, where the metric main effect mostly reflects the scales of the metrics. Within-metric shares test the exploratory hypothesis and carry no decision criterion. $^{\dagger}$Axis main effects plus method $\times$ axis interactions.}
\label{tab:H2}
\centering\scriptsize
\setlength{\tabcolsep}{3.5pt}
\resizebox{\columnwidth}{!}{%
\begin{tabular}{@{}lcc@{}}
\toprule
 & Method share & Axis share$^{\dagger}$\\
\midrule
voxel AUROC & 0.954 [0.933, 0.966] & 0.046 [0.033, 0.067]\\
voxel AUPRC & 0.960 [0.941, 0.972] & 0.040 [0.028, 0.059]\\
slice AUROC, marginal & 0.974 [0.964, 0.980] & 0.025 [0.019, 0.036]\\
Dice & 0.770 [0.738, 0.787] & 0.207 [0.190, 0.234]\\
slice AROC, adjusted & 0.718 [0.627, 0.783] & 0.267 [0.207, 0.355]\\
lesion sensitivity & 0.139 [0.127, 0.152] & 0.657 [0.648, 0.666]\\
\midrule
pooled over metrics & 0.162 [0.147, 0.177] & 0.614 [0.595, 0.634]\\
\quad of which metric main effect & -- & 0.539 [0.516, 0.561]\\
\bottomrule
\end{tabular}}
\end{table}

\begin{figure*}[t]
\centering
\includegraphics[width=0.78\textwidth]{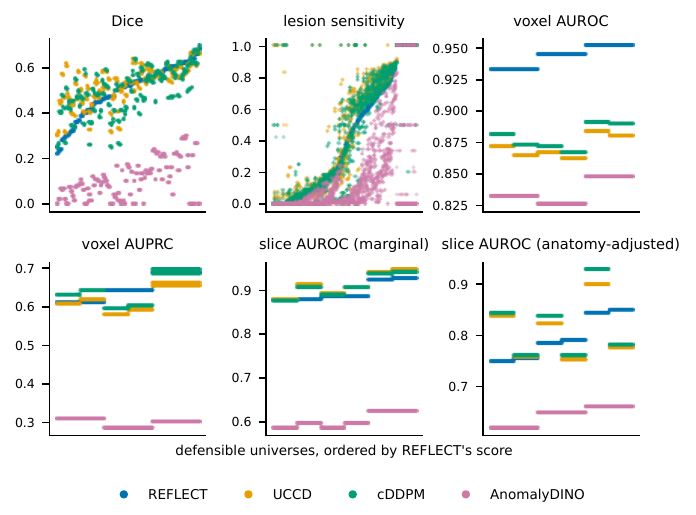}
\caption{\textbf{Multiverse.} Specification curves, one panel per metric; each column is a defensible universe ($2{,}592$ per metric), ordered by REFLECT's score. Dice and lesion sensitivity reorder the three residual models from universe to universe; voxel AUROC does not.}
\label{fig:M}
\end{figure*}

\subsection{Validation thresholds and realised false-positive burden}\label{sec:res-T}

\paragraph{Hypotheses.} A common view holds that tuning a threshold on the test cohort simply inflates Dice. \emph{H6 (test-tuning optimism):} the test-matched threshold overstates Dice by at least 0.02 at $20\cm$ (lower bound of the transfer gap at least 0.02). Identity~\ref{prop:transfer} suggests a competing account, in which the gap is a burden mismatch multiplied by a local slope, so that its sign follows that of $b-\hat\beta$ and differs between methods; a secondary hypothesis predicted a positive gap for UCCD at $5\cm$. For the per-scan burden, \emph{H8 (conformal control):} CRC keeps the test exceedance within $\alpha+0.02$ in all 128 calibrated combinations, and Lemma~\ref{prop:crc} predicts how the exceedance is distributed over calibration sets.

\begin{figure}[t]
\centering
\includegraphics[width=0.49\columnwidth]{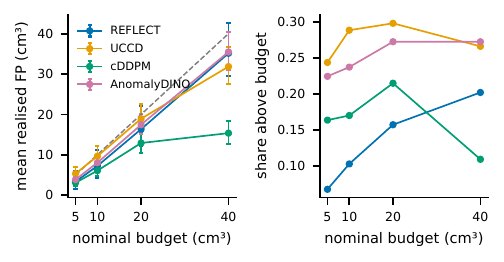}\hfill
\includegraphics[width=0.49\columnwidth]{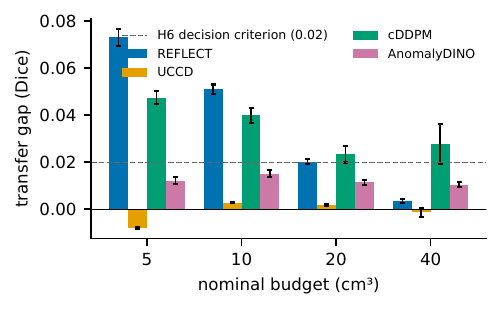}
\caption{\textbf{Threshold transfer.} (a) Mean realised burden on test against the nominal budget (left) and share of subjects above the budget (right) at the validation threshold. (b) Threshold-transfer gap $\bar g(b)$ with 95\% intervals; the dashed line is the decision criterion of H6.}
\label{fig:T}
\end{figure}

\paragraph{Nominal budgets are not realised.} At a nominal $20\cm$ the mean realised burden on test was 16.3 (REFLECT), 18.9 (UCCD), 12.9 (cDDPM) and 17.5\,cm$^3$ (AnomalyDINO). The median was 3.0 to 10.1\,cm$^3$, the 95th percentile 37 to 64\,cm$^3$, the maximum 210 to 604\,cm$^3$, and between 15.7\% and 29.8\% of subjects exceeded $20\cm$ (Table~\ref{tab:op}, Fig.~\ref{fig:T}a). The mean describes no typical subject; it is pulled up by a minority with very large false-positive volumes.

\begin{table*}[t]
\caption{Operating points at the validation threshold (plain thresholding). Dice: subject mean with interval, median and pooled value. Realised FP burden on test (cm$^3$): mean with interval, median, 95th percentile, maximum and share of subjects above $b$. Last column: threshold-transfer gap $\bar g(b)$ (Eq.~\ref{eq:gap}).}
\label{tab:op}
\centering\scriptsize
\setlength{\tabcolsep}{3.2pt}
\begin{tabular}{@{}llc ccc ccccc c@{}}
\toprule
Method & $b$ & $\hat\tau_b$ & Dice mean & median & pooled & FP mean & median & p95 & max & $>b$ & gap $\bar g(b)$\\
\midrule
REFLECT & 5 & 0.930 & 0.241 [0.223, 0.260] & 0.235 & 0.311 & 3.3 [1.5, 6.0] & 0.4 & 6.0 & 293 & 6.7\% & $+0.073$ [0.070, 0.077]\\
 & 10 & 0.870 & 0.382 [0.359, 0.405] & 0.409 & 0.458 & 7.2 [4.4, 11.2] & 2.0 & 17.1 & 431 & 10.3\% & $+0.051$ [0.049, 0.053]\\
 & 20 & 0.785 & 0.497 [0.473, 0.520] & 0.549 & 0.568 & 16.3 [12.2, 22.0] & 8.6 & 37.2 & 604 & 15.7\% & $+0.020$ [0.019, 0.022]\\
 & 40 & 0.685 & 0.554 [0.531, 0.577] & 0.604 & 0.615 & 35.1 [29.4, 42.6] & 24.9 & 76.6 & 801 & 20.2\% & $+0.003$ [0.003, 0.004]\\
\midrule
UCCD & 5 & 0.500 & 0.441 [0.426, 0.455] & 0.440 & 0.448 & 5.3 [4.0, 7.0] & 1.0 & 21.2 & 166 & 24.4\% & $-0.008$ [$-0.008$, $-0.007$]\\
 & 10 & 0.445 & 0.497 [0.482, 0.511] & 0.506 & 0.504 & 9.6 [7.6, 12.1] & 2.8 & 34.2 & 226 & 28.8\% & $+0.003$ [0.002, 0.003]\\
 & 20 & 0.385 & 0.539 [0.524, 0.555] & 0.557 & 0.548 & 18.9 [15.7, 22.6] & 8.9 & 64.0 & 307 & 29.8\% & $+0.002$ [0.002, 0.002]\\
 & 40 & 0.340 & 0.551 [0.535, 0.568] & 0.580 & 0.559 & 31.8 [27.5, 36.8] & 19.4 & 108.0 & 377 & 26.6\% & $-0.001$ [$-0.003$, 0.001]\\
\midrule
cDDPM & 5 & 0.530 & 0.396 [0.381, 0.412] & 0.399 & 0.408 & 3.1 [2.3, 4.0] & 0.2 & 16.9 & 92 & 16.3\% & $+0.048$ [0.045, 0.050]\\
 & 10 & 0.470 & 0.462 [0.446, 0.477] & 0.474 & 0.477 & 6.1 [4.7, 7.6] & 0.8 & 28.0 & 140 & 17.0\% & $+0.040$ [0.037, 0.043]\\
 & 20 & 0.405 & 0.519 [0.503, 0.535] & 0.533 & 0.536 & 12.9 [10.5, 15.7] & 3.0 & 57.1 & 221 & 21.5\% & $+0.023$ [0.020, 0.027]\\
 & 40 & 0.390 & 0.530 [0.513, 0.546] & 0.550 & 0.546 & 15.4 [12.6, 18.5] & 4.2 & 67.2 & 247 & 10.9\% & $+0.028$ [0.019, 0.036]\\
\midrule
AnomalyDINO & 5 & 0.810 & 0.064 [0.054, 0.074] & 0.024 & 0.075 & 3.9 [3.4, 4.4] & 2.4 & 11.7 & 30 & 22.4\% & $+0.012$ [0.011, 0.014]\\
 & 10 & 0.770 & 0.105 [0.092, 0.119] & 0.065 & 0.129 & 8.0 [7.0, 9.2] & 4.7 & 26.9 & 83 & 23.7\% & $+0.015$ [0.014, 0.017]\\
 & 20 & 0.725 & 0.166 [0.149, 0.183] & 0.133 & 0.208 & 17.5 [15.1, 20.0] & 10.1 & 60.1 & 210 & 27.2\% & $+0.012$ [0.010, 0.013]\\
 & 40 & 0.680 & 0.230 [0.210, 0.249] & 0.220 & 0.287 & 35.5 [31.0, 40.4] & 22.3 & 117.9 & 386 & 27.2\% & $+0.011$ [0.009, 0.012]\\
\bottomrule
\end{tabular}
\end{table*}

\paragraph{The transfer gap is a burden mismatch.} The gap was $0.02$ to $0.07$ Dice for REFLECT and cDDPM at 5 to $20\cm$, because their validation thresholds were conservative on test (REFLECT realised $3.3\cm$ at a nominal 5; cDDPM $12.9\cm$ at 20). For UCCD it was close to zero or negative ($-0.008\ci{-0.008}{-0.007}$ at $5\cm$): its validation threshold almost equalled the test-matched one (0.385 against 0.381 at $20\cm$) and was slightly liberal ($5.3\cm$ realised at 5). This is exactly the pattern Eq.~\eqref{eq:gapmvt} predicts: the sign of $\bar g$ follows the sign of $b-\hat\beta$, scaled by the local slope of $\Psi$. Where $\Psi$ is flat the gap vanishes even for a large mismatch (UCCD at $40\cm$: $\hat\beta=31.8$, $\bar g=-0.001$). The gap therefore measures how well a validation threshold transfers, which differs between methods; it is not a uniform test-set optimism. This is why the secondary hypothesis of a positive UCCD gap at $5\cm$ failed, and why H6, which required a lower bound of at least 0.02 at $20\cm$, narrowly failed (lower bounds 0.0191 for REFLECT and 0.0199 for cDDPM).

\begin{figure*}[t]
\centering
\includegraphics[width=0.9\textwidth]{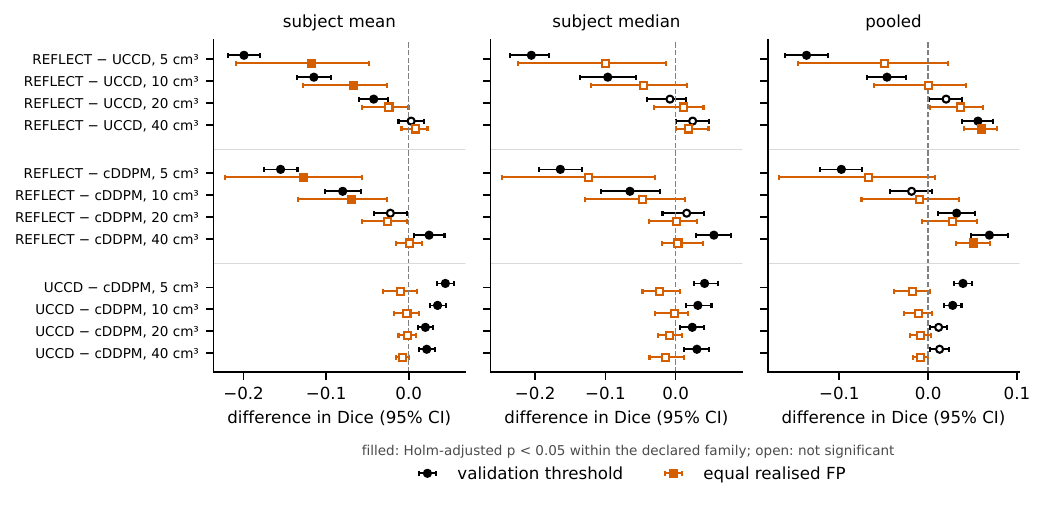}
\caption{\textbf{Paired Dice contrasts between residual models} at the validation threshold (black) and at equal realised burden (orange), by budget and aggregation. Filled markers: Holm-adjusted $p<0.05$ in the declared 24-contrast family. The UCCD $-$ cDDPM advantage at validation thresholds disappears at equal burden; REFLECT $-$ UCCD changes sign with budget and aggregation.}
\label{fig:pairs}
\end{figure*}

\paragraph{Consequence for a ranking.} At the validation threshold UCCD beat cDDPM at every budget, $+0.020\ci{0.011}{0.029}$ at $20\cm$ with $\pH<0.005$ (Fig.~\ref{fig:pairs}). At equal realised burden the difference was $-0.001\ci{-0.012}{0.009}$ ($\pH=1.00$), and every interval at every budget included zero. Identity~\ref{prop:transfer} accounts for the change: $\Delta^{\mathrm{val}}-\Delta^{\star}=-(\bar g_{\mathrm{UCCD}}-\bar g_{\mathrm{cDDPM}})=-(0.002-0.023)=+0.021$, against an observed $+0.021$. Across all 24 model pairs and budgets, the two sides of Eq.~\eqref{eq:transfer}, computed from separately produced tables, agreed within 0.0045 Dice with the grid-matched threshold and within 0.0017 with the interpolated one; the residual is the difference between the threshold grids of the two tables. The UCCD advantage at validation thresholds is a property of how its threshold transfers, not of its map.

\paragraph{A risk-controlled alternative and its price.} With CRC at $\alpha=0.10$ on the 24 validation subjects, the test exceedance $\Pr[\FP_i>20\cm]$ was $0.032$ to $0.096$ across methods, against $0.157$ to $0.298$ with the validation threshold, at a Dice cost of $0.053$ (REFLECT) to $0.116$ (cDDPM) (Appendix~\ref{app:thr}, Table~\ref{tab:crc}). All 128 calibrated combinations (4 methods $\times$ 4 budgets $\times$ 8 calibrated rules) stayed within $\alpha+0.02$, as H8 required; in fact none exceeded $\alpha$, the largest excess being $-0.004$. RCPS at $\alpha=0.05$ is infeasible with $n=24$. Over 1,000 random calibration splits the realised exceedance followed the law of Lemma~\ref{prop:crc} closely (Table~\ref{tab:crcbeta}): with 24 calibration subjects the mean exceedance was $0.078$--$0.082$ against a predicted $0.080$, the share of splits at or below $\alpha$ was $69.5$--$71.2\%$ against $70.8\%$, and the 97.5th percentile $0.205$--$0.222$ against $0.211$. The guarantee holds on average; at $n=24$ it fails for roughly three calibration sets in ten, and Eq.~\eqref{eq:crcbeta} gives the spread from $n$ and $\alpha$ alone.

\paragraph{Verdict.} H6 is not supported: the transfer gap behaves as Identity~\ref{prop:transfer} predicts, not as a uniform optimism, and the secondary hypothesis on UCCD failed for the same reason. H8 is supported, and the spread of the exceedance over calibration sets matches Lemma~\ref{prop:crc}.

\begin{table}[t]
\caption{CRC ($\alpha=0.10$, $b=20\cm$) over 1,000 random calibration splits of the test cohort: observed range over the four methods against the exact law $\mathrm{Beta}(n+1-k,k)$ of Lemma~\ref{prop:crc}.}
\label{tab:crcbeta}
\centering\scriptsize
\setlength{\tabcolsep}{2.6pt}
\resizebox{\columnwidth}{!}{%
\begin{tabular}{@{}c c cc cc cc@{}}
\toprule
 & & \multicolumn{2}{c}{mean exceedance} & \multicolumn{2}{c}{share $\le\alpha$} & \multicolumn{2}{c}{97.5th pct.}\\
\cmidrule(lr){3-4}\cmidrule(lr){5-6}\cmidrule(lr){7-8}
$n$ & law & observed & exact & observed & exact & observed & exact\\
\midrule
24 & Beta(2, 23) & 0.078--0.082 & 0.080 & 69.5--71.2\% & 70.8\% & 0.205--0.222 & 0.211\\
50 & Beta(5, 46) & 0.097--0.099 & 0.098 & 55.6--59.3\% & 56.9\% & 0.191--0.206 & 0.192\\
100 & Beta(10, 91) & 0.096--0.100 & 0.099 & 53.7--57.7\% & 54.9\% & 0.170--0.179 & 0.164\\
\bottomrule
\end{tabular}}
\end{table}

\subsection{Comparison of methods}\label{sec:res-lead}

\paragraph{Hypothesis (exploratory).} With geometry gated and operating points explicit, the residual models should still differ in ways that hold across operating points for some metrics but not others. Where pooled and subject-mean Dice disagree, Identity~\ref{prop:pooleddice} predicts that the method gaining more from pooling is the one whose Dice is concentrated in subjects with large $p_i+G_i$. The contrasts are tested in declared families with Holm adjustment.

With the evaluation made explicit the residual models can be compared (Fig.~\ref{fig:pairs}; Appendix~\ref{app:lead}, Table~\ref{tab:pairs}). At small budgets UCCD and cDDPM achieve higher Dice than REFLECT, also at equal burden: REFLECT $-$ UCCD was $-0.200\ci{-0.219}{-0.180}$ at $5\cm$ and $-0.118\ci{-0.209}{-0.048}$ at equal burden (both $\pH<0.005$). At $20\cm$ the sign depended on the aggregation (mean $-0.042$, median $-0.008$, pooled $+0.020$). At $40\cm$ pooled, REFLECT led by $+0.056\ci{0.038}{0.073}$ at the validation threshold and $+0.060$ at equal burden (both $\pH<0.005$). Identity~\ref{prop:pooleddice} gives the mechanism: pooling raises a method's Dice above its subject mean in proportion to the covariance between $\Dice_i$ and the weight $p_i+G_i$ (Eq.~\ref{eq:poolcov}). At $20\cm$ pooling lifted REFLECT by $0.071$ ($0.497\to0.568$) but UCCD by only $0.009$ ($0.539\to0.548$), so REFLECT's Dice is concentrated in the subjects that carry the most weight. For a typical subject, as measured by the median, REFLECT and UCCD were not distinguishable at 20 or $40\cm$ after adjustment ($\pH$ from 0.14 to 1.00), whereas REFLECT $-$ cDDPM at $40\cm$ was $+0.055\ci{0.029}{0.079}$ in the median as well.

Threshold-free metrics gave a stable order (Table~\ref{tab:thrfree}). REFLECT had the highest voxel AUROC in every universe and every configuration set: $0.932$ against $0.865$ (UCCD), $0.873$ (cDDPM) and $0.832$ (AnomalyDINO), with paired contrasts of $+0.067\ci{0.060}{0.074}$ and $+0.059\ci{0.051}{0.066}$ over the two diffusion models. Medians exceeded means for every residual model, reflecting left-skewed per-subject distributions with a tail of hard subjects. Pooling lowered AUPRC for UCCD and cDDPM but raised it for REFLECT, so the AUPRC order depends on the aggregation. AnomalyDINO was last in every universe of every metric. The comparison is consistent with the hypothesis: the voxel-AUROC order is stable, the Dice order depends on the budget and the aggregation, and the direction of the aggregation effect is the one Identity~\ref{prop:pooleddice} predicts.

\begin{table}[t]
\caption{Threshold-free metrics per method: subject mean [95\% CI], subject median and pooled value.}
\label{tab:thrfree}
\centering\scriptsize
\setlength{\tabcolsep}{2.2pt}
\resizebox{\columnwidth}{!}{%
\begin{tabular}{@{}l ccc ccc@{}}
\toprule
 & \multicolumn{3}{c}{voxel AUROC} & \multicolumn{3}{c}{voxel AUPRC}\\
\cmidrule(lr){2-4}\cmidrule(lr){5-7}
 & mean & med. & pool & mean & med. & pool\\
\midrule
REFLECT & 0.932 [0.925, 0.939] & 0.951 & 0.945 & 0.611 [0.582, 0.639] & 0.689 & 0.642\\
UCCD & 0.865 [0.857, 0.872] & 0.880 & 0.862 & 0.608 [0.589, 0.626] & 0.655 & 0.580\\
cDDPM & 0.873 [0.865, 0.881] & 0.889 & 0.867 & 0.630 [0.611, 0.649] & 0.686 & 0.593\\
AnomalyDINO & 0.832 [0.822, 0.841] & 0.848 & 0.826 & 0.311 [0.290, 0.331] & 0.303 & 0.286\\
\bottomrule
\end{tabular}}
\end{table}

\subsection{Post-processing}\label{sec:res-pp}

\paragraph{Hypotheses.} Identity~\ref{prop:m4} implies that a post-processing gain measured at equal burden may come largely from relaxing the plain threshold rather than from the post-processing itself, and Identity~\ref{prop:purity} that deleting low-purity components raises Dice whilst removing small true lesions. Secondary hypotheses made this concrete: for UCCD at $20\cm$ the relaxation term is positive, excludes zero and exceeds the fixed-threshold term; for AnomalyDINO the total gain is at most zero; and above-floor sensitivity falls with post-processing for UCCD and AnomalyDINO. \emph{H5 (pruning trade-off):} pruning helps lesion detection at $5\cm$ and hurts it at $40\cm$.

\begin{figure*}[t]
\centering
\includegraphics[width=0.72\textwidth]{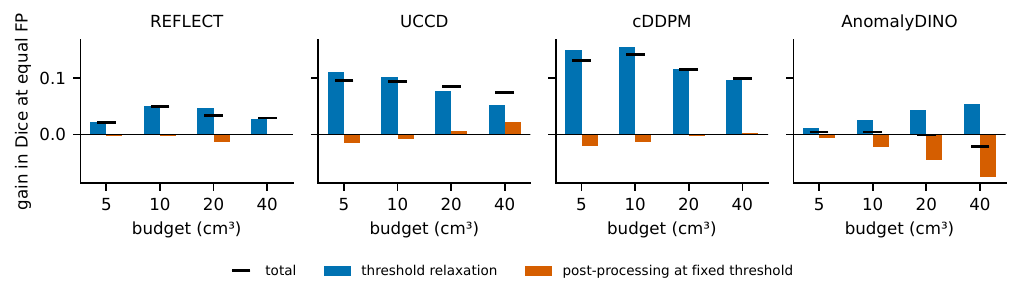}\hfill
\includegraphics[width=0.26\textwidth]{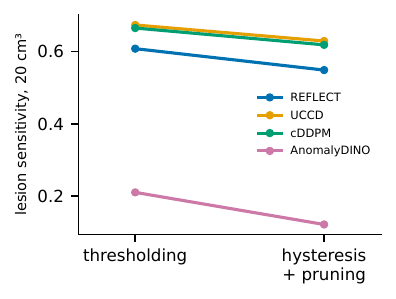}
\caption{\textbf{Post-processing at equal burden.} (a) Gain of hysteresis with pruning over plain thresholding, split into the fixed-threshold effect and the threshold relaxation (Identity~\ref{prop:m4}). (b) Above-floor lesion sensitivity at $20\cm$ falls with post-processing for every method.}
\label{fig:pp}
\end{figure*}

Hysteresis with persistence pruning raised Dice at equal realised burden for the three residual models, but almost all of the gain came from the burden it freed (Table~\ref{tab:m4}, Fig.~\ref{fig:pp}). For UCCD at $20\cm$ the total of $+0.085$ split into $+0.006\ci{-0.004}{0.016}$ from post-processing at fixed threshold and $+0.078\ci{0.073}{0.083}$ from relaxing the plain threshold, as predicted: 92\% of the gain is relaxation. For REFLECT the step itself \emph{lowered} Dice ($-0.013\ci{-0.022}{-0.005}$) and relaxation supplied $+0.046$. For AnomalyDINO the total was null, $-0.001\ci{-0.011}{0.010}$, again as predicted. The step also removed lesions: above-floor sensitivity fell from 0.674 to 0.629 (UCCD), 0.665 to 0.619 (cDDPM), 0.608 to 0.549 (REFLECT) and 0.210 to 0.121 (AnomalyDINO), and small-lesion sensitivity fell to 0.000 for REFLECT. The purity rule explains why: between 73.5\% (AnomalyDINO) and 82.5\% (UCCD) of predicted components had purity $t'/p'<\Dice/2$, so deleting them raises Dice, and components that contain small lesions are amongst them. H5 was not supported: the detection effect was negative at both budgets for every method with Dice $\ge0.3$ (point estimates only; no interval was computed for this measure). The three secondary hypotheses held. Results at every budget are in Appendix~\ref{app:pp}.

\begin{table}[t]
\caption{Exact decomposition (Eq.~\ref{eq:m4}) of the gain from hysteresis with pruning at equal realised burden, $b=20\cm$, and above-floor lesion sensitivity with plain thresholding and with post-processing.}
\label{tab:m4}
\centering\scriptsize
\setlength{\tabcolsep}{2.2pt}
\resizebox{\columnwidth}{!}{%
\begin{tabular}{@{}l ccc c@{}}
\toprule
 & total & fixed-$\tau$ effect & relaxation & sens.\ global$\to$pp\\
\midrule
REFLECT & $+0.034$ & $-0.013$ [$-0.022$, $-0.005$] & $+0.046$ [0.042, 0.050] & 0.608$\to$0.549\\
UCCD & $+0.085$ & $+0.006$ [$-0.004$, 0.016] & $+0.078$ [0.073, 0.083] & 0.674$\to$0.629\\
cDDPM & $+0.116$ & $-0.002$ [$-0.012$, 0.008] & $+0.117$ [0.110, 0.125] & 0.665$\to$0.619\\
AnomalyDINO & $-0.001$ & $-0.044$ [$-0.055$, $-0.035$] & $+0.044$ [0.036, 0.051] & 0.210$\to$0.121\\
\bottomrule
\end{tabular}}
\end{table}

\subsection{Slice-level detection}\label{sec:res-slice}

\paragraph{Hypotheses.} Tumours sit in large mid-brain slices, so a slice-level score may reward a map for tracking how much brain a slice contains. \emph{H9 (anatomy adjustment):} if covariate adjustment removes this anatomy, a scorer that sees only brain area obtains a matched AROC within $0.50\pm0.02$. A secondary hypothesis predicts that UCCD and AnomalyDINO retain subject-specific information, with observed minus borrowed-map AROC above zero.

\begin{figure}[t]
\centering
\includegraphics[width=\columnwidth]{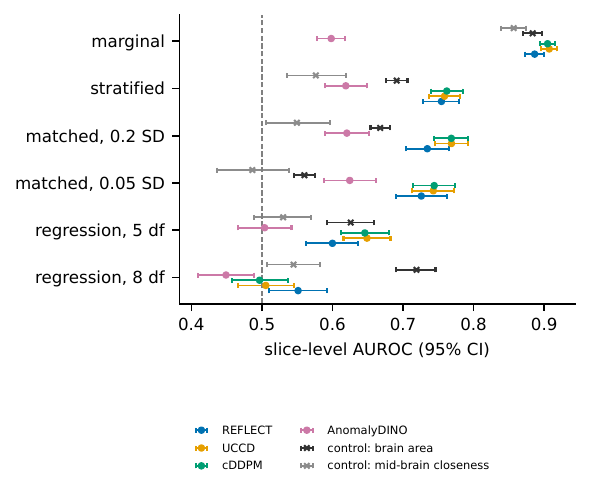}
\caption{\textbf{Slice-level AUROC under six estimators.} Covariate-only controls (crosses) should sit at 0.5 if an estimator removes anatomy; none does.}
\label{fig:slice}
\end{figure}

A scorer that knows only how much brain a slice contains reached a marginal slice AUROC of $0.884\ci{0.869}{0.897}$, within 0.025 of every residual model ($0.887$ to $0.907$; Table~\ref{tab:slice}, Fig.~\ref{fig:slice}). Covariate adjustment lowered the residual models to 0.73--0.77, but it did not remove anatomy entirely: at the primary caliper of 0.2 standard deviations the brain-area control still scored $0.668\ci{0.654}{0.682}$, so H9 was not supported, and 0.560 at 0.05. The spline-regression estimator also failed its own control (0.626 with 5 degrees of freedom, 0.719 with 8). The obstacle is positivity: tumour slices sit in large mid-brain slices, so within a subject there are few negative slices of comparable area, and matching kept only 59\% of positives at 0.2 SD and 46\% at 0.05 SD. The borrowed-map contrast avoids the problem by changing the scores rather than the comparison set: every model retained subject-specific information, observed minus borrowed AROC being $+0.229$ (REFLECT), $+0.266$ (UCCD), $+0.271$ (cDDPM) and $+0.124$ (AnomalyDINO), all intervals excluding zero, so the secondary hypothesis held, for the other two models as well. The marginal order (UCCD $>$ cDDPM $>$ REFLECT) also differs from the voxel-AUROC order. Intervals for every estimator and the dependence on the caliper are given in Appendix~\ref{app:slice}.

\begin{table}[t]
\caption{Slice-level AUROC: marginal, covariate-adjusted (stratified; matched with two calipers; spline regression, 5 df), and observed minus borrowed-map AROC.}
\label{tab:slice}
\centering\scriptsize
\setlength{\tabcolsep}{2.4pt}
\resizebox{\columnwidth}{!}{%
\begin{tabular}{@{}l ccccc c@{}}
\toprule
Scorer & marg. & strat. & m.\,0.2 & m.\,0.05 & reg.\,5 & obs.$-$borrowed\\
\midrule
REFLECT & 0.887 & 0.754 & 0.734 & 0.726 & 0.600 & $+0.229$ [0.201, 0.257]\\
UCCD & 0.907 & 0.759 & 0.769 & 0.743 & 0.649 & $+0.266$ [0.244, 0.288]\\
cDDPM & 0.905 & 0.762 & 0.768 & 0.744 & 0.646 & $+0.271$ [0.249, 0.293]\\
AnomalyDINO & 0.598 & 0.619 & 0.620 & 0.624 & 0.504 & $+0.124$ [0.100, 0.148]\\
\midrule
brain area & 0.884 & 0.691 & 0.668 & 0.560 & 0.626 & --\\
mid-brain closeness & 0.857 & 0.576 & 0.550 & 0.486 & 0.530 & --\\
\bottomrule
\end{tabular}}
\end{table}

\subsection{Lesion-level detection}\label{sec:res-detect}

\paragraph{Hypotheses.} If no sub-floor fragment is ever hit, the lesion floor changes pooled sensitivity only through its denominator, by the ratio $|\mathcal L_f|/|\mathcal L_0|$ (Appendix~\ref{app:proofs}). \emph{H4 (contrast dependence):} detection rises with T2 contrast and with oedema fraction in every residual model. Secondary hypotheses predict that UCCD finds fewer than 10\% of small lesions and that the three residual models miss largely the same lesions (Jaccard index of the missed sets of UCCD with REFLECT and with cDDPM at least 0.70).

The lesion definition rescaled pooled sensitivity about elevenfold through the denominator alone. REFLECT hit 287 lesions under both definitions; $287/472=0.608$ with the $50\mm$ floor and $287/5{,}473=0.052$ without, a ratio of $5{,}473/472=11.6$ that is identical for every method because the 5,001 sub-floor fragments were never hit. Subject-mean sensitivity moved less (0.765 against 0.173). A sensitivity computed over all components is not a sensitivity in any clinical sense, and a lesion-level metric is uninterpretable without its connectivity and floor.

\begin{figure*}[t]
\centering
\includegraphics[width=0.36\textwidth]{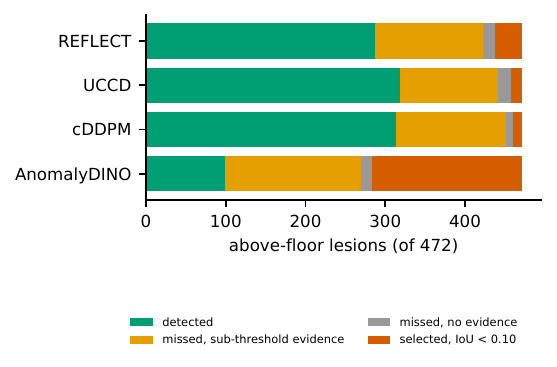}\hfill
\includegraphics[width=0.62\textwidth]{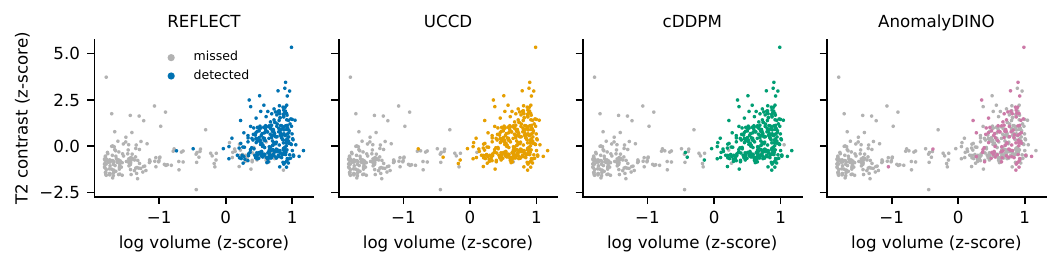}
\caption{\textbf{Detection at $20\cm$, validation threshold.} (a) Error modes of the 472 above-floor lesions. (b) Detected (coloured) and missed (grey) lesions by standardised log volume and T2 contrast: volume separates them almost completely.}
\label{fig:detect}
\end{figure*}

At the validation threshold the residual models found nearly every large lesion and almost no small one: large-lesion sensitivity was 0.915 to 0.997 and small-lesion sensitivity 0.036 to 0.073, at most 12 of 165 small lesions, as the secondary hypothesis predicted. The three residual models missed largely the same lesions (Jaccard index of the missed sets 0.805 to 0.938), which supports the second secondary hypothesis. Most misses carried evidence that the threshold did not select: of the 472 above-floor lesions, 123 to 137 were missed with in-lesion scores above the subject's healthy-brain median, 9 to 17 had no such evidence, and 12 to 35 were selected but delineated with IoU $<0.10$ (Fig.~\ref{fig:detect}a). Lesion volume dominated detection in every model (Firth-penalised log-odds 4.8 to 6.0 per standard deviation of log volume for the residual models; Fig.~\ref{fig:detect}b), whereas in the Firth fit T2 contrast added an effect whose interval excluded zero only for REFLECT ($+1.98\ci{1.22}{2.82}$) and AnomalyDINO ($+0.29\ci{0.03}{0.55}$); for cDDPM only the GEE interval excluded zero (Appendix Table~\ref{tab:s3}). H4 was therefore not supported; it could be tested only in part, because oedema fractions are not available in the binary labels. Label-informed upper bounds that respect the per-subject budget (Appendix~\ref{app:detect}, Table~\ref{tab:oracle}) place the remaining headroom mainly in delineation: perfect boundaries for detected lesions would add 0.40 to 0.46 Dice at $20\cm$, an optimal threshold per lesion 0.12 to 0.16, and an optimal threshold per subject 0.03 to 0.07.

\subsection{Intensity baselines}\label{sec:res-base}

\paragraph{Hypothesis.} Thresholding raw intensity has been reported to rival learnt detectors~\cite{meissen2022challenging}. \emph{H3 (intensity baseline):} z-scored T2 comes within 0.05 Dice of at least one model at equal realised burden at $20\cm$. Identity~\ref{prop:pooledauc} adds an exploratory prediction: a score normalised per subject, such as z-scored FLAIR, can rank voxels well within subjects yet pool poorly.

\begin{figure}[t]
\centering
\includegraphics[width=\columnwidth]{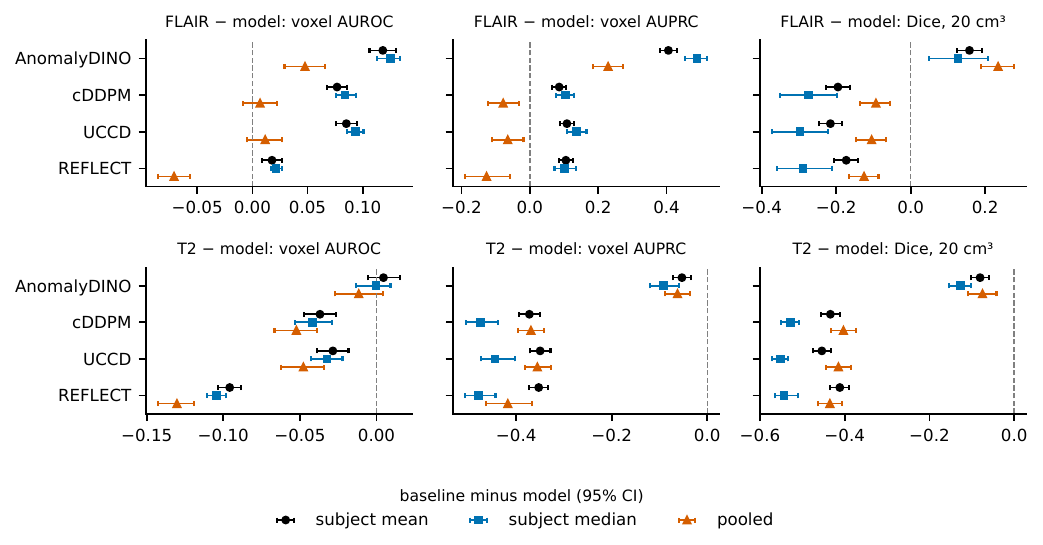}
\caption{\textbf{Baselines minus models} for voxel AUROC, voxel AUPRC and Dice ($20\cm$, validation threshold) under three aggregations, with Holm-adjusted families. FLAIR beats the residual models only in subject-averaged threshold-free metrics.}
\label{fig:base}
\end{figure}

Every model, AnomalyDINO included, beat z-scored T2 intensity in Dice at matched burden at every budget: the residual models by 0.28 to 0.46 across budgets (0.44 to 0.46 at $20\cm$) and AnomalyDINO by $0.098\ci{0.076}{0.119}$ at $20\cm$ so H3 was not supported. FLAIR, which the models never see, exceeded each of the three residual models in subject-mean voxel AUROC and AUPRC, but not when voxels were pooled, and never in Dice; it exceeded AnomalyDINO in every metric and aggregation (Fig.~\ref{fig:base}). For REFLECT the voxel-AUROC difference reversed sign with the aggregation, $+0.018\ci{0.009}{0.027}$ in the subject mean and $-0.071\ci{-0.085}{-0.057}$ pooled, both $\pH<0.005$. Identity~\ref{prop:pooledauc} suggests a reading: per-subject z-scoring makes FLAIR rank voxels well within each subject, but the between-subject term of Eq.~\eqref{eq:pooledauc} penalises its uncalibrated scale across subjects. AnomalyDINO was indistinguishable from raw T2 in voxel AUROC ($+0.005\ci{-0.006}{0.016}$, T2 minus model). All baseline contrasts, at every budget and aggregation, are listed in Appendix~\ref{app:base}.

\subsection{Training data}\label{sec:res-config}

\paragraph{Hypotheses (exploratory).} If the healthy training data were a neutral background to a comparison of methods, giving every method the same alternative training set should leave the contrasts between methods unchanged. And if a training set could be chosen per method from validation data, the choice should be stable under resampling of the 24 validation subjects.

\begin{figure*}[t]
\centering
\includegraphics[width=0.53\textwidth]{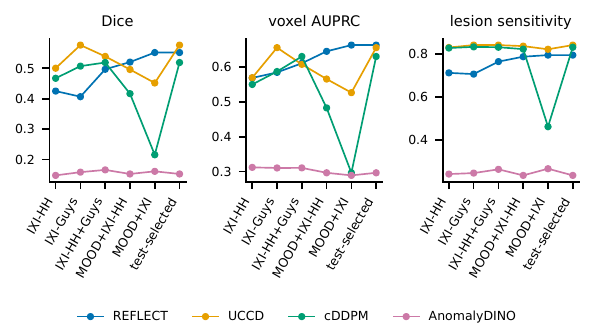}\hfill
\includegraphics[width=0.45\textwidth]{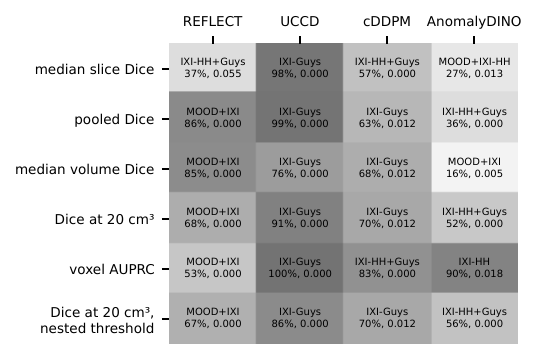}
\caption{\textbf{Training data.} (a) Dice, voxel AUPRC and lesion sensitivity in the reference universe when all four methods share one training set (Table~\ref{tab:trainsets}), and in the test-selected set, in which each method used the training set with its best test Dice. (b) Training set chosen on the 24 validation subjects by six selection rules; each cell gives the choice, the share of $10^4$ bootstrap resamples that reproduce it (also shown by the shading) and the test Dice lost relative to the best training set.}
\label{fig:config}
\end{figure*}

Holding the evaluation fixed and giving all four methods the same alternative training set reversed a significant contrast (Fig.~\ref{fig:config}a). REFLECT $-$ UCCD Dice was $-0.042\ci{-0.060}{-0.025}$ with all methods trained on IXI-HH+Guys and $+0.100\ci{0.079}{0.121}$ with all trained on MOOD+IXI, both $\pH<0.005$. The method $\times$ training-set interaction carried 0.115 of Dice variance across the five training sets, and the four-method order differed from the IXI-HH+Guys order in 50\% to 92\% of defensible universes, depending on the comparison set. The MOOD+IXI comparison needs a caveat: it adds T1 data for every method, and cDDPM's MOOD+IXI training set was 72\% T1, so it compares the same sources with multi-contrast training. The comparison on IXI-HH+Guys is the clean one.

Choosing a training set per method on 24 validation subjects was itself unstable (Fig.~\ref{fig:config}b). The primary selection rule (median validation slice Dice) chose IXI-HH+Guys for REFLECT in only 36.8\% of $10^4$ bootstrap resamples (MOOD+IXI 37.9\%, MOOD+IXI-HH 22.8\%), at a test regret of 0.055 Dice. Across resamples each rule produced between 27 and 137 distinct joint choices for the four methods, and the most frequent joint choice never exceeded 37.5\%. Both hypotheses are therefore rejected, which is the empirical case for fixing a common training set by design, as we did. H1 and H2 held with every common training set (Appendix~\ref{app:config}). All twenty method--training-set combinations and the verdicts of the hypotheses with each set of training data are in Appendix~\ref{app:config}.

\subsection{Latent aggregation for REFLECT}\label{sec:res-C1}

\paragraph{Hypothesis.} REFLECT's four latent channels need not agree on where an anomaly lies, and its published map averages them. If their disagreement carries signal, aggregating the channels by it should sharpen the map and raise Dice at equal burden. The median absolute deviation across channels (MAD) is retained as the recommended aggregation only if its interval does not overlap that of the best other aggregation.

\begin{figure*}[t]
\centering
\includegraphics[width=0.8\textwidth]{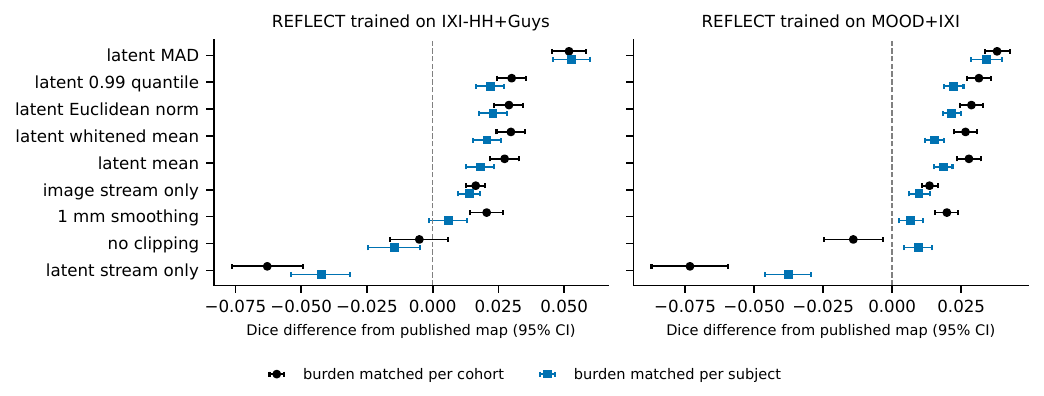}
\caption{\textbf{Latent aggregation for REFLECT.} Dice of each score rule minus the published map at $20\cm$, matched to the published map's realised burden over the cohort (black) or per subject (blue), for REFLECT trained on two different training sets.}
\label{fig:c1}
\end{figure*}

Replacing REFLECT's published map by the fusion of Eq.~\eqref{eq:c1score} raised Dice at equal burden in two independently trained models (Fig.~\ref{fig:c1}). At $20\cm$ with cohort matching, the MAD aggregation improved on the published map by $+0.052\ci{0.045}{0.058}$ for REFLECT trained on IXI-HH+Guys and by $+0.038\ci{0.033}{0.043}$ for REFLECT trained on MOOD+IXI; every latent aggregation improved it in both models; scoring with the latent stream alone hurt ($-0.063$ and $-0.073$); and the image stream alone beat the published fusion ($+0.016$, $+0.014$). At its own validation threshold, which is what a user would obtain, MAD reached $0.544\ci{0.521}{0.566}$ (median 0.597) at $15.3\cm$ realised burden against $0.492\ci{0.468}{0.516}$ for the published map (median 0.542) at $15.7\cm$ (on the 0.01 threshold grid used for this study, hence the small difference from Table~\ref{tab:op}), with above-floor sensitivity 0.631 against 0.606. The retention criterion kept MAD ahead of the next-best aggregation in 10 of 16 settings for the IXI-HH+Guys model, including the primary setting, but in only 5 of 16 for the MOOD+IXI model, where it lost the primary setting. We therefore claim the family of results (latent aggregation helps, latent-only scoring hurts, the image stream beats the published fusion), not MAD in particular. The hypothesis is thus supported for latent aggregation as a family but not for MAD as a unique choice. A family of six rules specified for this experiment could not be computed from the stored latents; the family reported here replaces it and is exploratory (Appendix~\ref{app:c1}).

\subsection{Summary of hypotheses}\label{sec:res-H}

Table~\ref{tab:H} collects the verdicts of H1 to H9 under their decision criteria. Four were supported, four were not, and one could be tested only in part. Seven of the eight secondary hypotheses held; the one that failed, a positive transfer gap for UCCD at $5\cm$, failed for the reason given in Section~\ref{sec:res-T} (Appendix~\ref{app:prereg}, Table~\ref{tab:P}).

\begin{table}[t]
\caption{Hypotheses H1 to H9, their decision criteria and verdicts (all methods trained on IXI-HH+Guys).}
\label{tab:H}
\centering\scriptsize
\setlength{\tabcolsep}{2.5pt}
\begin{tabular}{@{}>{\raggedright\arraybackslash}p{0.17\columnwidth}>{\raggedright\arraybackslash}p{0.43\columnwidth}>{\raggedright\arraybackslash}p{0.34\columnwidth}@{}}
\toprule
 Hypothesis & Decision criterion & Verdict; estimate\\
\midrule
H1 ranking instability & a pair flips in $\ge20\%$ of defensible universes (lower bound) & supported; $\phi_{\max}=0.472$ [0.414, 0.499]\\
H2 evaluation dominance & evaluation choices explain more variance than method, pooled over metrics & supported, driven by the metric axis (scale); 0.614 vs 0.162\\
H3 intensity baseline & T2 within 0.05 Dice of $\ge1$ model at equal FP & not supported; closest $+0.098$ [0.076, 0.119]\\
H4 contrast dependence & detection rises with T2 contrast and oedema fraction in every residual model & partly testable; not supported\\
H5 pruning trade-off & pruning helps detection at 5, hurts at 40\,cm$^3$ & not supported (point estimates)\\
H6 test-tuning optimism & test-matched $\tau$ overstates Dice by $\ge0.02$ at $20\cm$ (lower bound) & not supported; cDDPM $+0.0234$ [0.0199, 0.0269], REFLECT $+0.0203$ [0.0191, 0.0216]\\
H7 diagnostic power & rank stability and anti-location rate detect in-plane errors, power $\ge0.95$ at $n=60$ & supported; power 1.000\\
H8 conformal control & CRC exceedance $\le\alpha+0.02$ in all 128 combinations & supported\\
H9 anatomy adjustment & covariate-only matched AROC within $0.50\pm0.02$ & not supported; 0.668 [0.654, 0.682]\\
\bottomrule
\end{tabular}
\end{table}

\section{Discussion}\label{sec:discussion}

The same four models produced different winners in four distinct ways. The metric decides whether REFLECT or a conditioned diffusion model leads. The aggregation decides it again at a fixed budget, and Identity~\ref{prop:pooleddice} explains the direction. The operating point turns an advantage of UCCD over cDDPM that is significant after Holm adjustment into nothing once both are held to the same realised burden, and Identity~\ref{prop:transfer} accounts for the whole difference as a difference in threshold transfer. And the training data, changed in the same way for every method, reverses a significant contrast. None of these is exotic. Each corresponds to a decision that a typical UAD paper takes once, silently, and reports as though it were not a decision at all.

\paragraph{What each component changes.} Registration gating is the least visible component and, in this study, the one with the largest single effect: an axis-order mismatch cut a diffusion model's voxel AUROC from 0.873 to 0.583, whilst slice-level AUROC, a widely reported detection metric, barely changed. Such mismatches are not peculiar to one pipeline; three arose in assembling this benchmark, each yielding a plausible table. Any evaluation that compares methods with different preprocessing must map their outputs into a common space, and slice-level scores are invariant to precisely the in-plane permutations that wreck localisation. The registration gate and the label-free diagnostics catch such errors with measured power and leave correct pipelines alone; the obvious alternative, a Dice threshold on the method's own reference, would have rejected up to 242 correctly registered subjects per method.

The threshold-transfer component turns a familiar complaint, that thresholds are tuned on the test set, into something measurable. Identity~\ref{prop:transfer} shows that the gap between test-tuned and validation-tuned Dice is not a fixed optimism to be subtracted but a product of a burden mismatch and a local slope, which differs from method to method. A comparison at validation thresholds reflects what a user would obtain but confounds map quality with threshold transfer; a comparison at equal burden isolates map quality but cannot be reproduced in practice. MIRTO reports both, and the identity says exactly how they relate. Conformal risk control adds a per-scan guarantee, which is closer to what matters for an individual patient than a cohort mean. The known training-conditional law of Lemma~\ref{prop:crc} adds the caveat that a 24-subject calibration set honours the guarantee only on average, with a spread that follows from the calibration size alone, and the empirical splits followed that law closely (shares within 2.8 percentage points, 97.5th percentiles within 0.015). A user who needs the guarantee for the calibration set in hand should use RCPS or Learn-then-Test~\cite{bates2021rcps,angelopoulos2025ltt} and accept a stricter threshold.

The multiverse quantifies what the previous components illustrate, and it is most informative metric by metric. Pooled over metrics, the evaluation axes carry most of the variance, but mainly because metrics have different scales, and we do not lean on that share. Within a metric the picture is sharper. Rankings by voxel AUROC are a property of the method. Dice is dominated by the method in variance, chiefly because one method is far behind, yet the order of the three residual models is decided by the evaluation, with leads of several Dice points in each direction and significant contrasts on both sides. Lesion sensitivity is mostly a property of the evaluation. Reading flip shares together with the size of each difference is what separates these cases from mere closeness: UCCD and cDDPM swap places often because they are close, and their significant difference at validation thresholds is a matter of threshold transfer. Lesion-level claims deserve the least confidence unless the lesion definition, the hit criterion and the floor are stated, and our denominator identity shows that the floor alone can change sensitivity elevenfold without a single hit changing.

\paragraph{Recommendations.} For authors of UAD methods we suggest a short list, which the checklist released with the code expands. Score every method against one reference on one grid, and report a registration check. Choose thresholds on data that are not the test set, and report the realised false-positive burden as a distribution (median and 95th percentile, not only the mean). Report subject mean, median and pooled Dice side by side, and when two methods swap places across evaluation choices, report by how much each leads. State the lesion definition in full. Do not report marginal slice AUROC without an anatomy control. Report paired intervals with a declared multiplicity correction. And train competing methods on the same healthy data, or say plainly that they were not.

\paragraph{Relation to other benchmarks.} Recent benchmarks have improved UAD evaluation by enlarging and diversifying the test data~\cite{frotscher2025benchmark}, by examining the normative reconstruction~\cite{bercea2025normative}, or by unifying many methods and datasets within one framework~\cite{lagogiannis2024deepdive,bao2024bmad,cai2025medianomaly}. MIRTO is complementary. It does not propose a new dataset; it asks how much of any benchmark's conclusion survives a change of defensible evaluation choices, and it provides the identities needed to say why a conclusion changes when it does. The same machinery could be wrapped around any of those benchmarks. It also answers, for UAD, the concerns raised for medical imaging AI in general: that performance variability and confidence intervals are rarely reported~\cite{christodoulou2024ci}, that many outperformance claims are probably false~\cite{christodoulou2025falsepromises}, and that biases enter at every step of evaluation~\cite{varoquaux2022failures}. MIRTO adds a source of uncertainty that intervals over subjects do not capture, the choice of evaluation pipeline, and shows how large it is.

\paragraph{Limitations.} The evidence comes from a single cohort and a single contrast. BraTS 2020 whole-tumour labels are large, conspicuous lesions, and evaluation on T2 alone limits what can be said about small or subtle pathology; the small-lesion results here are a floor, not a characterisation. The 312 test subjects served both to develop the protocol and the choice of training set and to test the hypotheses, so all inference is exploratory; explicit decision criteria keep a hypothesis from being bent to fit its outcome, but they cannot remove the influence of other analyses of the same cohort. One trained model exists per method and training set, so training-seed variance, which can be large enough to change benchmark conclusions~\cite{bouthillier2021variance} and is the most likely competing explanation for small between-method differences, is unknown. A common training set means the same source data, not the same optimisation budget, and IXI-HH+Guys is not the best training set for every method: on test it cost REFLECT 0.055 and UCCD 0.037 Dice relative to their best training sets (Appendix Table~\ref{tab:allcfg}), so the comparison of methods should not be read as each method's best attainable result. The injection study validated the diagnostics on two pipelines; their power on other architectures is assumed rather than shown. Validation-threshold intervals do not include the uncertainty of fitting a threshold on 24 subjects, which a nested resampling would add. The covariate-adjusted slice estimators failed their own control, so the borrowed-map contrast, not the adjusted AUROC, is the defensible slice-level measure. Finally, the training-free improvement to REFLECT is shown for one method and two trained models, and the specific winning aggregation did not replicate.

\paragraph{Future work.} The next step is a pre-registered confirmatory evaluation on an untouched cohort, with the training set and every threshold rule fixed by its protocol. Nested resampling of the validation threshold, several training seeds per method and a second pathology with smaller lesions would close the main gaps listed above.

\section{Conclusion}\label{sec:conclusion}

A UAD leaderboard is a statement about models and evaluation jointly. In our study the method decided rankings by voxel AUROC, whereas the evaluation decided the order of the three strongest models in Dice and in lesion sensitivity. MIRTO separates the two. It gates the geometry of the comparison with tests of known power, holds models to operating points chosen without the test set and reports what those operating points actually deliver, evaluates every conclusion over a multiverse of defensible pipelines, reading how often an order flips together with how large each difference is, and attaches paired, multiplicity-adjusted intervals to each claim. Elementary identities explain why a contrast changes between validation thresholds and matched burden, why post-processing appears to help and why pooled and subject-level rankings disagree, and a known conformal result states how often calibration on a small set will miss its target. Applied to four current UAD families, the protocol exposed a geometric error that standard metrics missed, showed which apparent differences were artefacts of threshold transfer, and still supported some definite statements for this cohort and training set: REFLECT had the highest voxel AUROC, the two conditioned diffusion models gave higher Dice at small budgets, no method found more than 12 of the 165 small lesions, and a training-free change to REFLECT's latent aggregation raised its Dice at matched burden in two checkpoints. Reporting evaluation choices with the same care as architectures would make UAD comparisons much easier to interpret.

\begin{credits}
\subsubsection{\ackname} The authors thank Prof.\ Francesca Ieva of Politecnico di Milano and Human Technopole for creating the opportunity for this research and for her support. S.C.\ thanks Dr Craig Glastonbury of Human Technopole for his continuous support and encouragement. 

\subsubsection{Author contributions (CRediT).} \textbf{N.K:} Data curation, Formal analysis, Investigation, Methodology, Software, Validation, Visualisation, Writing -- original draft. \textbf{Soumick Chatterjee:} Conceptualisation, Formal analysis, Methodology, Project administration, Resources, Supervision, Writing -- original draft.

\subsubsection{Data availability and ethics.} This study is a secondary analysis of publicly released datasets; no new data were acquired and no participant was contacted. BraTS 2020~\cite{menze2015brats,bakas2017tcga,bakas2018brats} was used under the organisers' terms, which permit research use provided that the designated publications are cited. The IXI dataset~\cite{ixi} is distributed under the Creative Commons CC BY-SA 3.0 licence. The MOOD 2020 brain data~\cite{zimmerer2022mood} are publicly available on Hugging Face (\url{https://huggingface.co/datasets/dzimmerdkfz/mood_brain})
Informed consent and ethical approval for data acquisition were obtained by the original studies, as described in the cited publications. The data must be obtained from their providers. Per-subject results underlying every table and figure are released with the code.

\subsubsection{Code availability.} The MIRTO evaluator, the analysis code, the hypotheses with their decision criteria and the reporting checklist are available at \url{https://github.com/soumickmj/MIRTO}. 
The four UAD methods were trained (or, for AnomalyDINO, built) and run from forks of their authors' public implementations, which adapt them to read the datasets used here and leave architectures and losses unchanged: REFLECT, \url{https://github.com/negin-kafee/REFLECT}; UCCD, \url{https://github.com/negin-kafee/unsupervised-contrastive-cond-diff}; cDDPM, \url{https://github.com/negin-kafee/Conditioned-Diffusion-Models-UAD}; AnomalyDINO, \url{https://github.com/negin-kafee/AnomalyDINO}.

\subsubsection{\discintname}
The authors have no competing interests to declare that are relevant to the content of this article.
\end{credits}
\bibliographystyle{splncs04}
\bibliography{references}
\clearpage
\appendix
\onecolumn
\section{Derivations of the identities}\label{app:proofs}

\paragraph{Identity~\ref{prop:transfer}.}
By definition $\bar g_m(b)=\bar\Dice_m(\tau^\star_{m,b})-\bar\Dice_m(\hat\tau_{m,b})$, since the mean of differences is the difference of means. Hence
\[
\Delta^{\star}-\Delta^{\mathrm{val}}=\big[\bar\Dice_a(\tau^\star_{a,b})-\bar\Dice_a(\hat\tau_{a,b})\big]-\big[\bar\Dice_c(\tau^\star_{c,b})-\bar\Dice_c(\hat\tau_{c,b})\big]=\bar g_a(b)-\bar g_c(b),
\]
which is Eq.~\eqref{eq:transfer}. For Eq.~\eqref{eq:gapmvt}, on an interval where $\beta_m$ is strictly decreasing, $\bar\Dice_m(\tau)=\Psi_m(\beta_m(\tau))$, so $\bar g_m=\Psi_m(\beta_m(\tau^\star))-\Psi_m(\hat\beta_m)$, and $\beta_m(\tau^\star)=b$ for the interpolated threshold (and $\le b$, within one grid step, for Eq.~\eqref{eq:tm}). The mean-value theorem gives $\xi_m$. Neither statement holds for medians or pooled Dice, which are not linear in the subjects' Dice. \qed

\paragraph{Lemma~\ref{prop:crc}.} The argument is the standard one for the training-conditional coverage of split-conformal prediction~\cite{vovk2012conditional,angelopoulos2021gentle}; we give it for the burden loss. Because $\FP_i(\tau)$ is non-increasing, $\{\tau:\FP_i(\tau)>b\}=\{\tau<t_i\}$ up to the convention at the boundary, so $L_i(\tau)=\ind[t_i>\tau]$ and $n\hat R_n(\tau)=\#\{i:t_i>\tau\}$. The condition in Eq.~\eqref{eq:crc} reads $\#\{i:t_i>\tau\}\le(n+1)\alpha-1$, i.e.\ at most $j=\lfloor(n+1)\alpha-1\rfloor$ calibration subjects exceed. The count is a non-increasing step function of $\tau$ that drops by one at each $t_{(r)}$, so the infimum is attained at $\tau=t_{(n-j)}=t_{(k)}$. For a new subject independent of the calibration set, $\Pr[t_{n+1}>t_{(k)}\mid t_{1:n}]=1-F(t_{(k)})$. By the probability integral transform $U_i=F(t_i)$ are i.i.d.\ uniform, $F(t_{(k)})=U_{(k)}\sim\mathrm{Beta}(k,n+1-k)$, and therefore $1-U_{(k)}\sim\mathrm{Beta}(n+1-k,k)$, with mean $(n+1-k)/(n+1)=(j+1)/(n+1)\le\alpha$. \qed

\emph{Remarks.} (i) With $n=24$ and $\alpha=0.10$, $(n+1)\alpha-1=1.5$, $j=1$, $k=23$, and $\rho\sim\mathrm{Beta}(2,23)$: $\E\rho=0.080$, $\Pr[\rho\le0.10]=0.708$, $\rho_{0.975}=0.211$. For $n=50$: $j=4$, $\mathrm{Beta}(5,46)$, mean 0.098, $\Pr[\rho\le\alpha]=0.569$; for $n=100$: $j=9$, $\mathrm{Beta}(10,91)$, mean 0.099, $0.549$. (ii) The share of calibration sets that achieve exceedance $\le\alpha$ tends to $\tfrac12$ as $n$ grows, because the Beta law concentrates at its mean, which approaches $\alpha$ from below; a per-set guarantee requires RCPS~\cite{bates2021rcps} or Learn-then-Test~\cite{angelopoulos2025ltt}, at the price of a stricter threshold. (iii) Our random splits draw calibration and evaluation sets without replacement from 312 subjects and use a threshold grid, so they approximate the i.i.d.\ continuous setting; Table~\ref{tab:crcbeta} shows the approximation is close.

\paragraph{Identity~\ref{prop:pooleddice}.}
$\Dice^{\mathrm{pool}}=2\sum_i\TP_i/\sum_k(p_k+G_k)$ and $2\TP_i=\Dice_i(p_i+G_i)$. \qed

\paragraph{Identity~\ref{prop:pooledauc}.}
The partition below is that of Obuchowski's clustered-ROC estimator~\cite{obuchowski1997clustered}. The pooled AUROC is the Mann--Whitney statistic over all $N^+N^-$ pairs of a positive and a negative voxel from the pooled set. Partition the pairs by the subjects $(i,j)$ of the positive and the negative voxel; there are $n_i^+n_j^-$ pairs in block $(i,j)$, whose average indicator is $A_{ij}$, and $A_{ii}=\AUROC_i$. \qed

\emph{Remark.} If scores are only ordinal within subjects (an arbitrary strictly increasing transform $h_i$ per subject leaves every $\AUROC_i$ unchanged), the diagonal term of Eq.~\eqref{eq:pooledauc} is invariant but the off-diagonal term is not. The pooled AUROC thus measures within-subject ranking \emph{and} between-subject calibration, whereas the subject mean measures only the former. Per-subject min--max normalisation (Eq.~\ref{eq:norm}) is one such $h_i$.

\paragraph{Identity~\ref{prop:purity}.}
$\Delta\Dice>0\iff(T-t')(p+G)>T(p-p'+G)\iff Tp'>t'(p+G)\iff t'/p'<T/(p+G)=\Dice/2$. Adding a region of $q'$ voxels containing $t''$ true positives: $\Delta\Dice>0\iff(T+t'')(p+G)>T(p+q'+G)\iff t''/q'>\Dice/2$. \qed

\paragraph{Identity~\ref{prop:m4}.} Add and subtract $\bar\Dice_{\mathrm g}(\tau)$. \qed

\paragraph{The Jaccard bound and the triangle inequality.} The Jaccard distance is a metric on finite sets (it is the Steinhaus transform of the symmetric-difference metric). For any $P$, $d_J(P,R_A)\le d_J(P,R_B)+d_J(R_B,R_A)$ and symmetrically, which gives Eq.~\eqref{eq:d5}.

\paragraph{Coverage-only ``matched Dice''.} If $P\subseteq\Cref$ then $\TP=|P|$ and $\Dice=2|P|/(|P|+|\Cref|)=2r/(1+r)$ with $r=|P|/|\Cref|$, a strictly increasing function of coverage alone that is blind to false positives.

\paragraph{Lesion-definition invariance of the hit count.} Let $\mathcal L_0$ be all reference components and $\mathcal L_f\subseteq\mathcal L_0$ those above the floor. If no component of $\mathcal L_0\setminus\mathcal L_f$ is hit, the pooled sensitivities satisfy $S_0/S_f=|\mathcal L_f|/|\mathcal L_0|$ for every method, which is what we observe ($472/5{,}473$).

\paragraph{Bootstrap $p$-value floors.} With the add-one two-sided $p$-value the smallest attainable value is $2/(B+1)$. Holm multiplies the smallest of $K$ raw $p$-values by $K$, so the floor of $\pH$ is $2K/(B+1)$: 0.0048 for $K=24$, 0.0032 for $K=16$, 0.0064 for $K=32$ at $B=10^4$.

\section{Physical units and voxel volumes}\label{app:units}

The method pipelines' own files declared voxel volumes of $0.8786\mm$ (one shared grid), $1.166\mm$ (UCCD) and $0.604\mm$ (cDDPM) for grids whose true voxel was $1\mm$ or $0.8789\mm$ ($256\times256$ in-plane). The same masks therefore produced burdens that differed by a factor of $1.93$ between UCCD and cDDPM, and a nominal $20\cm$ meant $17.2\cm$ for one and $33.1\cm$ for the other. MIRTO recovers the true grid from voxel counts alone: the ratio of the tumour volume in those files, divided by the declared voxel volume, to the canonical voxel count is $1.0004$ for the 240-pixel grid and $1.1380\approx(256/240)^2=1.1378$ for the 256-pixel grid. A round trip of the canonical reference through the UCCD/cDDPM native grid (2\,mm in-plane, through-plane decimation) loses 57\% of the 6-connected components, mostly sub-floor fragments, and 1.8\% of tumour volume, whilst the REFLECT and AnomalyDINO grids lose 0.01--0.02\% (Table~\ref{tab:roundtrip}). No MIRTO metric is computed on a method's own resampled reference. Table~\ref{tab:cohort} describes the cohort, and Tables~\ref{tab:voxcount} and~\ref{tab:declvox} give the voxel-count check and its consequence for the budget.

\begin{table}[H]
\caption{Cohort and canonical arrays.}
\label{tab:cohort}
\centering\scriptsize
\setlength{\tabcolsep}{3pt}
\resizebox{\textwidth}{!}{%
\begin{tabular}{@{}lllll@{}}
\toprule
Split & Canonical array (x $\times$ y $\times$ z voxels, 1 mm) & Whole-tumour volume, median [IQR] (cm$^3$) & Range (cm$^3$) & Distinct reference hashes\\
\midrule
test (312) & 240 $\times$ 240 $\times$ [98, 149], median 139 & 91.7 [53.2, 147.8] & 7.3 to 361.8 & 312\\
validation (24) &  & 89.6 [43.2, 137.8] & 12.5 to 180.7 & 24\\
\bottomrule
\end{tabular}}
\end{table}
\begin{table}[H]
\caption{Voxel-volume check from voxel counts: (tumour volume in the method's files / declared voxel volume) / canonical voxel count.}
\label{tab:voxcount}
\centering\scriptsize
\setlength{\tabcolsep}{3pt}
\begin{tabular}{@{}llllll@{}}
\toprule
Legacy bundle (method & grid) & Median ratio legacy/canonical voxel count & Range & $\approx$ 1.0004 (same 240-px grid) & $\approx$ 1.1380 = (256/240)$^2$\\
\midrule
AnomalyDINO|fourway & 1.0004 & 0.974 to 1.063 & True & False\\
AnomalyDINO|pairA & 1.0004 & 0.974 to 1.063 & True & False\\
REFLECT|fourway & 1.0188 & 0.974 to 1.146 & False & False\\
REFLECT|pairA & 1.0188 & 0.974 to 1.146 & False & False\\
UCCD|fourway & 1.0514 & 0.992 to 1.167 & False & False\\
UCCD|pairB & 1.1380 & 1.122 to 1.196 & False & True\\
cDDPM|fourway & 1.0514 & 0.992 to 1.167 & False & False\\
cDDPM|pairB & 1.1380 & 1.122 to 1.196 & False & True\\
\bottomrule
\end{tabular}
\end{table}
\begin{table}[H]
\caption{Illustration with the uncorrected cDDPM mapping: the declared voxel volume changes the physical budget behind a nominal $20\cm$.}
\label{tab:declvox}
\centering\scriptsize
\setlength{\tabcolsep}{3pt}
\resizebox{\textwidth}{!}{%
\begin{tabular}{@{}llllllll@{}}
\toprule
Method (test-selected pairB grid) & Voxel volume used & Declared vox (mm$^3$) & Effective budget for 'nominal 20 cm$^3$' & $\tau$ & Dice & Sensitivity (all comps.) & Mean FP (cm$^3$)\\
\midrule
UCCD & as declared & 1.166 & 17.2 & 0.600 & 0.360 & 0.049 & 15.3\\
UCCD & under B3 vox=1.0 & 1.166 & 20.0 & 0.575 & 0.382 & 0.051 & 18.4\\
cDDPM & as declared & 0.604 & 33.1 & 0.600 & 0.029 & 0.003 & 30.5\\
cDDPM & under B3 vox=1.0 & 0.604 & 20.0 & 0.700 & 0.018 & 0.001 & 17.9\\
\bottomrule
\end{tabular}}
\end{table}

\section{Extended results}\label{app:ext}

\begin{table}[h]
\caption{Each method's score across the 2,592 defensible universes of a metric: median [IQR] and share of universes in which it ranks first (descriptive; the spread is over evaluation choices, not sampling).}
\label{tab:univ}
\centering\scriptsize
\begin{tabular}{@{}lllll@{}}
\toprule
Metric & REFLECT & UCCD & cDDPM & AnomalyDINO\\
\midrule
Dice & 0.515 [0.428, 0.584]; 35.2\% & 0.543 [0.472, 0.577]; 38.9\% & 0.496 [0.420, 0.568]; 25.9\% & 0.097 [0.030, 0.157]; 0\%\\
voxel AUROC & 0.945 [0.932, 0.951]; 100\% & 0.869 [0.865, 0.880]; 0\% & 0.877 [0.871, 0.889]; 0\% & 0.832 [0.826, 0.848]; 0\%\\
voxel AUPRC & 0.642 [0.611, 0.689]; 50.0\% & 0.613 [0.590, 0.655]; 0\% & 0.636 [0.604, 0.686]; 50.0\% & 0.303 [0.286, 0.311]; 0\%\\
lesion sens. & 0.341 [0.064, 0.699]; 35.3\% & 0.343 [0.077, 0.817]; 53.5\% & 0.333 [0.060, 0.793]; 11.3\% & 0.011 [0.000, 0.134]; 0\%\\
slice AUROC marg. & 0.886 [0.878, 0.921]; 0\% & 0.909 [0.891, 0.940]; 100\% & 0.905 [0.890, 0.938]; 0\% & 0.598 [0.587, 0.625]; 0\%\\
slice AROC adj. & 0.787 [0.754, 0.843]; 33.3\% & 0.800 [0.759, 0.837]; 0\% & 0.808 [0.762, 0.842]; 66.7\% & 0.650 [0.619, 0.661]; 0\%\\
\bottomrule
\end{tabular}
\end{table}

\paragraph{Determinants of detection.} For each above-floor lesion we model $\mathrm{logit}\Pr[\mathrm{hit}_{\ell}]=\beta_0+\beta_1 z^{\mathrm{vol}}_\ell+\beta_2 z^{\mathrm{con}}_\ell+\beta_3 z^{\mathrm{pos}}_\ell+u_{i(\ell)}$, with standardised log volume, T2 contrast (median T2 in the lesion over median T2 in normal-appearing brain on the same slices) and relative through-plane position, and a subject random intercept $u_i\sim\mathcal N(0,\sigma_u^2)$. Because 472 lesions sit in 312 subjects (87 with more than one) and volume nearly separates hits from misses, the random intercept is weakly identified; we therefore also fit Firth-penalised logistic regression~\cite{firth1993}, GEE with exchangeable working correlation~\cite{liang1986gee} and a variational-Bayes GLMM, and report the Laplace fit of lme4~\cite{bates2015lme4}. Firth, GEE and variational Bayes agree in sign and magnitude (log volume $+4.8$ to $+6.8$ per SD for the residual models); the Laplace fit did not converge for REFLECT.

\section{Further results}\label{app:further}

This appendix reports the analyses that support the main text in full. Unless a caption says otherwise, all methods were trained on IXI-HH+Guys, the cohort is the 312 test subjects, and intervals are 95\% paired subject-bootstrap intervals. Analyses computed on the uncorrected geometry are not reported, except in Table~\ref{tab:repair}, where the comparison between the two is the point.

\subsection{Registration, geometry and reference}\label{app:geom}

Table~\ref{tab:repair} gives the full effect of correcting the cDDPM mapping, including Dice, which the main text summarises. Table~\ref{tab:g1} gives the registration-gate statistics per method; the overlap of each method's own reference with the canonical reference inside the field of view never fell below 0.845, and the centroid error never exceeded 1.07\,mm.
\begin{table}[H]
\caption{cDDPM with the incorrect and the correct mapping (same maps; correct minus incorrect, 312 subjects). Dice rows compare each pipeline at its own validation threshold.}
\label{tab:repair}
\centering\scriptsize
\setlength{\tabcolsep}{3pt}
\begin{tabular}{@{}p{0.62\textwidth}ll@{}}
\toprule
cDDPM (IXI-HH+Guys) quantity & Statistic & Repaired $-$ pre-repair [95 \% CI]\\
\midrule
voxel AUROC & mean & +0.290 [0.277, 0.302]\\
voxel AUROC & median & +0.293 [0.277, 0.306]\\
voxel AUPRC & mean & +0.535 [0.517, 0.553]\\
voxel AUPRC & median & +0.577 [0.549, 0.599]\\
slice-max AUROC & mean & +0.112 [0.099, 0.125]\\
slice-max AUROC & median & +0.083 [0.072, 0.095]\\
Dice, thresholding, 5 cm$^3$ (correct ($\tau$ 0.5) $-$ incorrect ($\tau$ 0.875)) & mean & +0.426 [0.410, 0.441]\\
Dice, thresholding, 5 cm$^3$ (correct ($\tau$ 0.5) $-$ incorrect ($\tau$ 0.875)) & median & +0.431 [0.416, 0.455]\\
Dice, thresholding, 20 cm$^3$ (correct ($\tau$ 0.375) $-$ incorrect ($\tau$ 0.7)) & mean & +0.520 [0.504, 0.537]\\
Dice, thresholding, 20 cm$^3$ (correct ($\tau$ 0.375) $-$ incorrect ($\tau$ 0.7)) & median & +0.547 [0.526, 0.564]\\
\bottomrule
\end{tabular}
\end{table}
\begin{table}[H]
\caption{Registration gate per method on 312 test and 24 validation subjects. ``Dice in FOV'' is the overlap of the method's own reference, mapped to the canonical grid, with the canonical reference inside the field of view; the column for the fixed 0.95 overlap rule counts the subjects that would pass it.}
\label{tab:g1}
\centering\scriptsize
\setlength{\tabcolsep}{3pt}
\resizebox{\textwidth}{!}{%
\begin{tabular}{@{}llllllllll@{}}
\toprule
Method & gate pass (test + validation) & Dice in FOV, median [IQR] & min & centroid error median (mm) & max (mm) & FOV tumour fraction, median & subjects with FOV fraction $<$ 0.95 & pass at overlap $\ge$ 0.95 & gate failures (clause)\\
\midrule
REFLECT & 336/336 & 0.966 [0.957, 0.972] & 0.890 & 0.58 & 1.02 & 1.000 & 0 & 275 & none\\
UCCD & 335/336 & 0.947 [0.935, 0.957] & 0.845 & 0.85 & 1.07 & 1.000 & 45 & 93 & BraTS20\_Training\_0044 (iii)\\
cDDPM & 335/336 & 0.947 [0.935, 0.957] & 0.845 & 0.85 & 1.07 & 1.000 & 45 & 93 & BraTS20\_Training\_0044 (iii)\\
AnomalyDINO & 336/336 & 0.966 [0.957, 0.972] & 0.901 & 0.59 & 1.01 & 1.000 & 0 & 273 & none\\
\bottomrule
\end{tabular}}
\end{table}
Table~\ref{tab:m1} gives the bias of every headline metric under every injected corruption for both verified pipelines, and Table~\ref{tab:m1power} gives the power of the rank-stability and anti-location diagnostics. Rank stability detects every geometric corruption at $n=30$, including through-plane shifts that barely change the metrics, which is why we read it as a change detector. The anti-location rate detects the in-plane errors at $n=30$ and otherwise reaches a power of 0.95 only at $n=312$, for two AnomalyDINO conditions (50\% identifier permutation and doubled reference volume). A 5\% permutation of subject identifiers is not detectable by either diagnostic at $n\le312$, and a 20\% permutation only by rank stability at $n=312$. Removing sub-floor reference fragments changed no metric, as expected, since no method ever hit one.
\begin{table}[H]
\caption{Bias of the headline metrics (corrupted minus unaltered, cohort mean) and the rank-stability (Spearman) and anti-location diagnostics under every injected corruption. The Jaccard bound was never violated.}
\label{tab:m1}
\centering\scriptsize
\setlength{\tabcolsep}{3pt}
\resizebox{\textwidth}{!}{%
\begin{tabular}{@{}llllllllll@{}}
\toprule
Injected corruption & Pipeline & n & $\Delta$ Dice (20 cm$^3$) & $\Delta$ voxel AUROC & $\Delta$ voxel AUPRC & $\Delta$ slice-max AUROC & $\Delta$ lesion sensitivity & rank stability & anti-location rate\\
\midrule
LR flip & UCCD & 312 & $-$0.502 & $-$0.260 & $-$0.510 & +0.000 & $-$0.797 & 0.253 & 0.122\\
LR flip & AnomalyDINO & 312 & $-$0.145 & $-$0.307 & $-$0.231 & +0.021 & $-$0.252 & 0.297 & 0.449\\
AP flip & UCCD & 312 & $-$0.341 & $-$0.198 & $-$0.386 & +0.012 & $-$0.467 & 0.140 & 0.170\\
AP flip & AnomalyDINO & 312 & $-$0.086 & $-$0.147 & $-$0.118 & +0.181 & $-$0.158 & 0.360 & 0.151\\
transpose $xy$ & UCCD & 312 & $-$0.400 & $-$0.207 & $-$0.430 & +0.005 & $-$0.560 & 0.073 & 0.215\\
transpose $xy$ & AnomalyDINO & 312 & $-$0.106 & $-$0.184 & $-$0.163 & +0.099 & $-$0.192 & 0.243 & 0.208\\
$z$ shift 2 & UCCD & 312 & $-$0.002 & +0.003 & $-$0.003 & +0.009 & +0.000 & 0.959 & 0.000\\
$z$ shift 2 & AnomalyDINO & 312 & +0.001 & +0.006 & +0.011 & +0.030 & $-$0.000 & 0.987 & 0.000\\
$z$ shift 5 & UCCD & 312 & $-$0.025 & $-$0.011 & $-$0.041 & +0.017 & +0.000 & 0.751 & 0.000\\
$z$ shift 5 & AnomalyDINO & 312 & $-$0.002 & $-$0.000 & +0.012 & +0.066 & $-$0.005 & 0.927 & 0.000\\
$z$ shift 10 & UCCD & 312 & $-$0.097 & $-$0.060 & $-$0.155 & +0.010 & $-$0.007 & 0.363 & 0.006\\
$z$ shift 10 & AnomalyDINO & 312 & $-$0.016 & $-$0.036 & $-$0.022 & +0.087 & $-$0.033 & 0.703 & 0.000\\
$z$ scale 1.1 & UCCD & 312 & $-$0.043 & $-$0.018 & $-$0.082 & +0.023 & $-$0.005 & 0.713 & 0.000\\
$z$ scale 1.1 & AnomalyDINO & 312 & $-$0.002 & $-$0.011 & +0.003 & +0.095 & +0.000 & 0.854 & 0.000\\
ID permutation 5\% & UCCD & 297 & $-$0.002 & $-$0.001 & $-$0.002 & $-$0.000 & $-$0.003 & 0.998 & 0.003\\
ID permutation 5\% & AnomalyDINO & 297 & $-$0.002 & $-$0.002 & $-$0.002 & +0.000 & $-$0.003 & 0.990 & 0.003\\
ID permutation 20\% & UCCD & 256 & $-$0.008 & $-$0.005 & $-$0.008 & $-$0.002 & $-$0.016 & 0.993 & 0.008\\
ID permutation 20\% & AnomalyDINO & 256 & $-$0.005 & $-$0.006 & $-$0.006 & $-$0.001 & $-$0.012 & 0.952 & 0.008\\
ID permutation 50\% & UCCD & 166 & $-$0.022 & $-$0.013 & $-$0.022 & $-$0.008 & $-$0.045 & 0.930 & 0.006\\
ID permutation 50\% & AnomalyDINO & 166 & $-$0.013 & $-$0.016 & $-$0.015 & $-$0.012 & $-$0.033 & 0.868 & 0.018\\
reference dilation 1.5$\times$ & UCCD & 312 & $-$0.123 & $-$0.077 & $-$0.089 & $-$0.018 & +0.004 & 0.923 & 0.003\\
reference dilation 1.5$\times$ & AnomalyDINO & 312 & $-$0.029 & $-$0.046 & +0.030 & $-$0.016 & $-$0.075 & 0.967 & 0.006\\
reference dilation 2$\times$ & UCCD & 312 & $-$0.192 & $-$0.118 & $-$0.124 & $-$0.027 & +0.040 & 0.829 & 0.006\\
reference dilation 2$\times$ & AnomalyDINO & 312 & $-$0.049 & $-$0.077 & +0.042 & $-$0.028 & $-$0.088 & 0.923 & 0.010\\
sub-floor fragments removed & UCCD & 312 & +0.000 & +0.000 & +0.000 & +0.000 & +0.000 & 1.000 & 0.000\\
sub-floor fragments removed & AnomalyDINO & 312 & +0.000 & +0.000 & +0.000 & $-$0.000 & +0.000 & 1.000 & 0.000\\
\bottomrule
\end{tabular}}
\end{table}
\begin{table}[H]
\caption{Power of the rank-stability and anti-location diagnostics at a false-alarm rate fixed at 0.05 by construction; cohort sizes 30, 60 and 312.}
\label{tab:m1power}
\centering\scriptsize
\setlength{\tabcolsep}{3pt}
\resizebox{\textwidth}{!}{%
\begin{tabular}{@{}llllll@{}}
\toprule
Injected corruption & Pipeline & rank-stability power, n = 60 & anti-location power, n = 60 & smallest n with rank-stability power $\ge$ 0.95 & smallest n with anti-location power $\ge$ 0.95\\
\midrule
LR flip & UCCD & 1.000 & 1.000 & 30 & 30\\
LR flip & AnomalyDINO & 1.000 & 1.000 & 30 & 30\\
AP flip & UCCD & 1.000 & 1.000 & 30 & 30\\
AP flip & AnomalyDINO & 1.000 & 1.000 & 30 & 30\\
transpose $xy$ & UCCD & 1.000 & 1.000 & 30 & 30\\
transpose $xy$ & AnomalyDINO & 1.000 & 1.000 & 30 & 30\\
$z$ shift 2 & UCCD & 1.000 & 0.000 & 30 & not reached (n $\le$ 312)\\
$z$ shift 2 & AnomalyDINO & 1.000 & 0.000 & 30 & not reached (n $\le$ 312)\\
$z$ shift 5 & UCCD & 1.000 & 0.000 & 30 & not reached (n $\le$ 312)\\
$z$ shift 5 & AnomalyDINO & 1.000 & 0.000 & 30 & not reached (n $\le$ 312)\\
$z$ shift 10 & UCCD & 1.000 & 0.298 & 30 & not reached (n $\le$ 312)\\
$z$ shift 10 & AnomalyDINO & 1.000 & 0.000 & 30 & not reached (n $\le$ 312)\\
$z$ scale 1.1 & UCCD & 1.000 & 0.000 & 30 & not reached (n $\le$ 312)\\
$z$ scale 1.1 & AnomalyDINO & 1.000 & 0.000 & 30 & not reached (n $\le$ 312)\\
ID permutation 5\% & UCCD & 0.181 & 0.198 & not reached (n $\le$ 312) & not reached (n $\le$ 312)\\
ID permutation 5\% & AnomalyDINO & 0.182 & 0.195 & not reached (n $\le$ 312) & not reached (n $\le$ 312)\\
ID permutation 20\% & UCCD & 0.738 & 0.391 & 312 & not reached (n $\le$ 312)\\
ID permutation 20\% & AnomalyDINO & 0.730 & 0.375 & 312 & not reached (n $\le$ 312)\\
ID permutation 50\% & UCCD & 0.957 & 0.300 & 60 & not reached (n $\le$ 312)\\
ID permutation 50\% & AnomalyDINO & 0.965 & 0.644 & 60 & 312\\
reference dilation 1.5$\times$ & UCCD & 1.000 & 0.201 & 30 & not reached (n $\le$ 312)\\
reference dilation 1.5$\times$ & AnomalyDINO & 1.000 & 0.325 & 30 & not reached (n $\le$ 312)\\
reference dilation 2$\times$ & UCCD & 1.000 & 0.320 & 30 & not reached (n $\le$ 312)\\
reference dilation 2$\times$ & AnomalyDINO & 1.000 & 0.415 & 30 & 312\\
sub-floor fragments removed & UCCD & 0.754 & 0.000 & 312 & not reached (n $\le$ 312)\\
sub-floor fragments removed & AnomalyDINO & 0.500 & 0.000 & 312 & not reached (n $\le$ 312)\\
\bottomrule
\end{tabular}}
\end{table}
The single subject that failed the registration gate and the 42 subjects whose tumour extends beyond the 100-slice field of view of UCCD and cDDPM do not drive any result. Without subject 0044 (Table~\ref{tab:n311}) five of 588 contrasts change interval-excludes-zero status, all by less than 0.0013 at the bound. In the 270 subjects with at least 95\% of the tumour inside every field of view (Table~\ref{tab:fov}) four estimates change sign and none changes status. A pre-specified through-plane offset check (Table~\ref{tab:s2}) was positive only for AnomalyDINO, whose voxel AUROC peaks three slices away from the canonical position on both grids and in both subject subsets; its export was verified to be identical to the canonical data, and its numbers are reported without a shift.
\begin{table}[H]
\caption{Contrasts whose interval-excludes-zero status changes when subject BraTS20\_Training\_0044 is removed ($n=311$); 5 of 588 contrasts, largest absolute change of any point estimate 0.0078.}
\label{tab:n311}
\centering\scriptsize
\setlength{\tabcolsep}{3pt}
\begin{tabular}{@{}lllllll@{}}
\toprule
Metric & Variant & b & Aggregation & Contrast & n = 312 & n = 311 (without BraTS20\_Training\_0044)\\
\midrule
Dice & interpolated & 40 & pooled & UCCD $-$ cDDPM & $-$0.0087 [$-$0.0174, $-$0.0001] & $-$0.0081 [$-$0.0167, 0.0008]\\
Dice & testmatched & 40 & median & REFLECT $-$ UCCD & +0.0184 [0.0003, 0.0466] & +0.0184 [$-$0.0000, 0.0460]\\
Dice & validation & 5 & pooled & T2 $-$ AnomalyDINO & $-$0.0202 [$-$0.0400, 0.0001] & $-$0.0202 [$-$0.0400, $-$0.0002]\\
Dice & validation & 20 & pooled & REFLECT $-$ UCCD & +0.0202 [0.0013, 0.0386] & +0.0190 [$-$0.0001, 0.0370]\\
Dice & validation & 40 & median & REFLECT $-$ UCCD & +0.0242 [0.0006, 0.0480] & +0.0220 [$-$0.0001, 0.0480]\\
\bottomrule
\end{tabular}
\end{table}
\begin{table}[H]
\caption{Estimates whose sign differs between the full cohort and the FOV-complete subset (270 subjects); no interval-excludes-zero status changes.}
\label{tab:fov}
\centering\scriptsize
\setlength{\tabcolsep}{3pt}
\resizebox{\textwidth}{!}{%
\begin{tabular}{@{}lp{0.36\textwidth}lll@{}}
\toprule
Method / contrast & Quantity & Full 312 & FOV-complete subset & Interval-excludes-0 status changed\\
\midrule
cDDPM & FP (cm$^3$) / correct ($\tau$ 0.5) $-$ incorrect ($\tau$ 0.875), mean & +0.195 [$-$1.141, 1.613] & $-$0.010 [$-$1.453, 1.565] & False\\
cDDPM & FP (cm$^3$) / correct ($\tau$ 0.375) $-$ incorrect ($\tau$ 0.7), mean & +0.385 [$-$3.418, 4.281] & $-$0.124 [$-$4.048, 4.090] & False\\
pairwise & REFLECT - UCCD, auprc\_vox, mean & +0.003 [$-$0.015, 0.022] & $-$0.009 [$-$0.029, 0.011] & False\\
pairwise & REFLECT - cDDPM, auprc\_vox, median & +0.006 [$-$0.019, 0.032] & $-$0.007 [$-$0.023, 0.022] & False\\
\bottomrule
\end{tabular}}
\end{table}
\begin{table}[H]
\caption{Through-plane offset check: offset $dz^\ast$ chosen on 20 subjects, replicated on the next 20, and its effect on voxel AUROC in the remaining 292. ``Native'' refers to the method's own grid.}
\label{tab:s2}
\centering\scriptsize
\setlength{\tabcolsep}{3pt}
\resizebox{\textwidth}{!}{%
\begin{tabular}{@{}lllllll@{}}
\toprule
Method & Grid & dz* (selection 20) & dz* (replication 20) & $\Delta$ voxel AUROC at dz* vs 0 (292 subjects) & AUROC at dz = 0 & Offset real by declared rule\\
\midrule
AnomalyDINO & native & 3 & 3 & +0.0082 [0.0063, 0.0101] & 0.826 & True\\
REFLECT & native & 1 & 1 & $-$0.0005 [$-$0.0009, $-$0.0001] & 0.933 & False\\
UCCD & native & 0 & 0 & +0.0000 [0.0000, 0.0000] & 0.896 & False\\
cDDPM & native & 2 & 2 & $-$0.0005 [$-$0.0011, 0.0002] & 0.899 & False\\
AnomalyDINO & canonical & 3 & 3 & +0.0053 [0.0034, 0.0071] & 0.833 & True\\
REFLECT & canonical & 1 & 1 & $-$0.0004 [$-$0.0008, 0.0000] & 0.933 & False\\
UCCD & canonical & 2 & 3 & +0.0025 [0.0006, 0.0044] & 0.864 & False\\
cDDPM & canonical & 2 & 3 & +0.0023 [0.0005, 0.0043] & 0.873 & False\\
\bottomrule
\end{tabular}}
\end{table}
\begin{table}[H]
\caption{Round trip of the canonical reference through each method's native grid (nearest neighbour). The UCCD/cDDPM grid fragments the reference and loses 1.8\% of tumour volume, which is why no MIRTO metric uses a method's own reference.}
\label{tab:roundtrip}
\centering\scriptsize
\setlength{\tabcolsep}{3pt}
\resizebox{\textwidth}{!}{%
\begin{tabular}{@{}llllll@{}}
\toprule
Method grid & Subjects & 6-conn. components, canonical $\to$ round trip & Above-floor lesions & Tumour volume ratio (round trip / canonical) & Round trip equals method's own reference\\
\midrule
AnomalyDINO & 312 & 5,473 $\to$ 5,050 & 472 $\to$ 474 & 0.9999 & 100.0 \%\\
REFLECT & 312 & 5,473 $\to$ 5,065 & 472 $\to$ 472 & 0.9998 & 100.0 \%\\
UCCD & 312 & 5,473 $\to$ 2,355 & 472 $\to$ 543 & 0.9823 & 100.0 \%\\
cDDPM & 312 & 5,473 $\to$ 2,355 & 472 $\to$ 543 & 0.9823 & 100.0 \%\\
\bottomrule
\end{tabular}}
\end{table}
Ties matter for threshold-free metrics. Many brain voxels carry tied scores, a median of 65\% per subject for UCCD and 26\% for cDDPM (Table~\ref{tab:ties}). Ordinal ranks, which break ties by array order, changed a subject's voxel AUROC by up to 0.046 relative to mid-ranks; MIRTO uses mid-ranks and tie-collapsed average precision throughout.
\begin{table}[H]
\caption{Tie audit over 312 subjects per method: ordinal minus mid-rank voxel AUROC ($\Delta$AUROC) and ordinal minus tie-collapsed AUPRC.}
\label{tab:ties}
\centering\scriptsize
\begin{tabular}{@{}lccccc@{}}
\toprule
Method & brain voxels tied (median share) & outside FOV (median share) & mean $\Delta$AUROC & max $|\Delta$AUROC$|$ & max $|\Delta$AUPRC$|$\\
\midrule
REFLECT & 0.038 & 0.001 & -0.0000 & 0.0000 & 0.0000\\
UCCD & 0.649 & 0.091 & -0.0020 & 0.0464 & 0.0020\\
cDDPM & 0.256 & 0.091 & -0.0016 & 0.0391 & 0.0020\\
AnomalyDINO & 0.165 & 0.000 & -0.0000 & 0.0076 & 0.0037\\
\bottomrule
\end{tabular}
\end{table}
Spatial randomisation controls (Table~\ref{tab:p2}) apply known displacements to the stored maps: through-plane shifts by $\pm2$, $\pm5$ and $\pm10$ slices, a permutation of slices within the subject, and the exchange of maps across subjects matched on brain area. Shifting a map by two slices costs at most 0.009 Dice, by ten slices 0.08 to 0.12 for the residual models; permuting slices within a subject or exchanging maps across subjects removes most of the Dice, leaving 0.06 to 0.07 when maps come from another subject. This confirms that Dice at the validation threshold depends on subject-specific localisation, as it should.
\begin{table}[H]
\caption{Spatial randomisation controls: subject-mean Dice at $20\cm$, validation threshold, for the unshifted map, through-plane shifts of the map by the given number of slices, and slice permutation within or map exchange across subjects.}
\label{tab:p2}
\centering\scriptsize
\begin{tabular}{@{}lccccccccc@{}}
\toprule
Method & unshifted & $-2$ & $+2$ & $-5$ & $+5$ & $-10$ & $+10$ & perm.\ within & perm.\ across\\
\midrule
REFLECT & 0.497 & 0.493 & 0.494 & 0.472 & 0.476 & 0.406 & 0.413 & 0.183 & 0.066\\
UCCD & 0.539 & 0.530 & 0.537 & 0.497 & 0.514 & 0.418 & 0.442 & 0.192 & 0.068\\
cDDPM & 0.519 & 0.511 & 0.517 & 0.481 & 0.495 & 0.404 & 0.425 & 0.183 & 0.063\\
AnomalyDINO & 0.166 & 0.166 & 0.167 & 0.161 & 0.164 & 0.141 & 0.149 & 0.069 & 0.026\\
\bottomrule
\end{tabular}
\end{table}
A $2^4$ geometry factorial (Table~\ref{tab:p5}) switched four binary factors one at a time: the export transform (legacy or canonical), the reference (the method's own reference or the canonical one), the evaluation support (the method's own slices or the full canonical support) and the threshold (legacy or refitted on validation data). Apart from the threshold rule, every main effect on Dice was below 0.011 and every effect on voxel AUROC below 0.027 in absolute value; the threshold rule has no effect on AUROC by construction, which is a built-in check. The largest interactions were support $\times$ threshold ($+0.042$) and grid $\times$ threshold ($+0.022$), both for AnomalyDINO's Dice.
\begin{table}[H]
\caption{Geometry factorial: main effects with 95\% intervals on subject-mean Dice and voxel AUROC (second level minus first).}
\label{tab:p5}
\centering\scriptsize\setlength{\tabcolsep}{3pt}
\resizebox{\textwidth}{!}{\begin{tabular}{@{}llllll@{}}
\toprule
Method & quantity & F1 export grid & F2 reference source & F3 support & F4 threshold\\
\midrule
REFLECT & Dice & +0.0008 [0.0007, 0.0010] & +0.0002 [0.0000, 0.0003] & +0.0002 [0.0001, 0.0002] & $-$0.0605 [$-$0.0662, $-$0.0547]\\
REFLECT & voxel AUROC & +0.0002 [0.0001, 0.0004] & +0.0003 [0.0002, 0.0004] & $-$0.0000 [$-$0.0001, 0.0000] & +0.0000 [0.0000, 0.0000]\\
UCCD & Dice & $-$0.0017 [$-$0.0023, $-$0.0012] & $-$0.0016 [$-$0.0022, $-$0.0011] & +0.0001 [0.0001, 0.0001] & +0.2358 [0.2187, 0.2527]\\
UCCD & voxel AUROC & +0.0029 [0.0016, 0.0040] & $-$0.0038 [$-$0.0050, $-$0.0027] & $-$0.0262 [$-$0.0311, $-$0.0216] & +0.0000 [0.0000, 0.0000]\\
cDDPM & Dice & $-$0.0017 [$-$0.0023, $-$0.0011] & $-$0.0017 [$-$0.0023, $-$0.0012] & +0.0000 [0.0000, 0.0000] & +0.1926 [0.1731, 0.2123]\\
cDDPM & voxel AUROC & +0.0037 [0.0023, 0.0049] & $-$0.0041 [$-$0.0054, $-$0.0029] & $-$0.0210 [$-$0.0251, $-$0.0171] & +0.0000 [0.0000, 0.0000]\\
AnomalyDINO & Dice & +0.0011 [$-$0.0003, 0.0025] & $-$0.0001 [$-$0.0001, $-$0.0000] & +0.0109 [0.0081, 0.0138] & $-$0.0137 [$-$0.0296, 0.0026]\\
AnomalyDINO & voxel AUROC & +0.0021 [0.0019, 0.0023] & $-$0.0005 [$-$0.0006, $-$0.0004] & +0.0049 [0.0035, 0.0063] & +0.0000 [0.0000, 0.0000]\\
\bottomrule
\end{tabular}}
\end{table}
\subsection{Evaluation multiverse}\label{app:multiverse}

Tables~\ref{tab:H1full} and~\ref{tab:H2full} give every pairwise flip share and the full variance decomposition, for the defensible set and for all 155,520 universes. Including the non-defensible levels (the method's own reference, the round-trip reference, the declared voxel volume, the fixed threshold and the per-subject optimal threshold) changes neither verdict: $\phi_{\max}=0.450\ci{0.398}{0.498}$ and the axis share is 0.595 against a method share of 0.169. REFLECT versus UCCD and REFLECT versus cDDPM have the same point flip share (7,337 of 15,552 universes each) by coincidence of counts. Within metrics (Table~\ref{tab:flipwithin}) voxel AUROC never changes order, whereas Dice and lesion sensitivity change it for every pair. Agreement with the reference universe was low: mean Kendall $\tau_b$ 0.486 $[0.409, 0.549]$, median 0.667, identical four-method order in 24.3\% $[9.9, 32.4]$ of universes. Table~\ref{tab:orders} lists the distinct orders per metric.
\begin{table}[H]
\caption{Pairwise flip shares (share of universes in which a pair takes its minority order) with 1,000-replicate bootstrap intervals; defensible set (15,552 universes) and all universes (155,520).}
\label{tab:H1full}
\centering\scriptsize
\setlength{\tabcolsep}{3pt}
\begin{tabular}{@{}lll@{}}
\toprule
Quantity & Universe set & Flip share [95 \% CI]\\
\midrule
$\phi_{\max}$ & defensible & 0.472 [0.414, 0.499]\\
REFLECT vs UCCD & defensible & 0.472 [0.389, 0.499]\\
REFLECT vs cDDPM & defensible & 0.472 [0.386, 0.498]\\
REFLECT vs AnomalyDINO & defensible & 0.010 [0.009, 0.014]\\
UCCD vs cDDPM & defensible & 0.393 [0.250, 0.493]\\
UCCD vs AnomalyDINO & defensible & 0.009 [0.008, 0.011]\\
cDDPM vs AnomalyDINO & defensible & 0.011 [0.011, 0.013]\\
$\phi_{\max}$ & all & 0.450 [0.398, 0.498]\\
REFLECT vs UCCD & all & 0.450 [0.388, 0.497]\\
REFLECT vs cDDPM & all & 0.401 [0.350, 0.493]\\
REFLECT vs AnomalyDINO & all & 0.016 [0.014, 0.018]\\
UCCD vs cDDPM & all & 0.340 [0.180, 0.436]\\
UCCD vs AnomalyDINO & all & 0.016 [0.015, 0.020]\\
cDDPM vs AnomalyDINO & all & 0.016 [0.015, 0.021]\\
\bottomrule
\end{tabular}
\end{table}
\begin{table}[H]
\caption{Sum-of-squares shares, defensible and all universes.}
\label{tab:H2full}
\centering\scriptsize
\setlength{\tabcolsep}{3pt}
\begin{tabular}{@{}lll@{}}
\toprule
Sum-of-squares share & Defensible universes & All universes\\
\midrule
method share & 0.162 [0.147, 0.177] & 0.169 [0.153, 0.184]\\
axis share & 0.614 [0.595, 0.634] & 0.595 [0.575, 0.614]\\
remainder & 0.223 [0.212, 0.235] & 0.237 [0.226, 0.249]\\
main effect: metric & 0.539 [0.516, 0.561] & 0.519 [0.497, 0.541]\\
method $\times$ metric & 0.033 [0.030, 0.037] & 0.030 [0.028, 0.034]\\
main effect: lesion definition & 0.014 [0.012, 0.015] & 0.014 [0.013, 0.016]\\
method $\times$ lesion definition & 0.002 [0.002, 0.002] & 0.002 [0.001, 0.002]\\
main effect: hit & 0.011 [0.010, 0.012] & 0.012 [0.011, 0.013]\\
method $\times$ hit & 0.002 [0.001, 0.002] & 0.001 [0.001, 0.001]\\
main effect: burden & 0.004 [0.003, 0.005] & 0.002 [0.002, 0.002]\\
method $\times$ burden & 0.000 [0.000, 0.001] & 0.000 [0.000, 0.000]\\
main effect: aggregation & 0.004 [0.003, 0.005] & 0.005 [0.003, 0.006]\\
method $\times$ aggregation & 0.002 [0.001, 0.003] & 0.002 [0.002, 0.003]\\
main effect: threshold & 0.002 [0.001, 0.002] & 0.003 [0.002, 0.003]\\
method $\times$ threshold & 0.000 [0.000, 0.001] & 0.003 [0.003, 0.003]\\
main effect: post-processing & 0.000 [0.000, 0.000] & 0.000 [0.000, 0.000]\\
method $\times$ post-processing & 0.001 [0.001, 0.001] & 0.001 [0.001, 0.001]\\
main effect: support & 0.000 [0.000, 0.000] & 0.000 [0.000, 0.000]\\
method $\times$ support & 0.000 [0.000, 0.001] & 0.000 [0.000, 0.000]\\
\bottomrule
\end{tabular}
\end{table}
\begin{table}[H]
\caption{Within-metric flip shares for the three residual-model pairs (defensible universes; post hoc).}
\label{tab:flipwithin}
\centering\scriptsize
\setlength{\tabcolsep}{3pt}
\begin{tabular}{@{}llll@{}}
\toprule
Metric & REFLECT vs UCCD & REFLECT vs cDDPM & UCCD vs cDDPM\\
\midrule
voxel AUROC & 0.000 [0.000, 0.000] & 0.000 [0.000, 0.000] & 0.000 [0.000, 0.000]\\
voxel AUPRC & 0.167 [0.000, 0.500] & 0.500 [0.167, 0.500] & 0.000 [0.000, 0.000]\\
Dice & 0.403 [0.264, 0.495] & 0.491 [0.366, 0.500] & 0.356 [0.125, 0.412]\\
lesion sensitivity & 0.261 [0.219, 0.302] & 0.340 [0.282, 0.392] & 0.286 [0.238, 0.398]\\
slice AUROC, marginal & 0.000 [0.000, 0.333] & 0.167 [0.000, 0.333] & 0.000 [0.000, 0.500]\\
slice AROC, adjusted & 0.333 [0.333, 0.500] & 0.333 [0.333, 0.500] & 0.000 [0.000, 0.500]\\
\bottomrule
\end{tabular}
\end{table}
\begin{table}[H]
\caption{Distinct four-method orders per metric across the 2,592 defensible universes of each metric, with the three most frequent.}
\label{tab:orders}
\centering\scriptsize
\setlength{\tabcolsep}{3pt}
\begin{tabular}{@{}lcp{0.6\textwidth}@{}}
\toprule
Metric & Distinct four-method orders & Three most frequent orders (share of universes)\\
\midrule
Dice & 6 & REFLECT $>$ UCCD $>$ cDDPM $>$ AnomalyDINO (25.5 \%); UCCD $>$ cDDPM $>$ REFLECT $>$ AnomalyDINO (25.0 \%); cDDPM $>$ UCCD $>$ REFLECT $>$ AnomalyDINO (20.8 \%)\\
voxel AUROC & 1 & REFLECT $>$ cDDPM $>$ UCCD $>$ AnomalyDINO (100.0 \%)\\
voxel AUPRC & 3 & REFLECT $>$ cDDPM $>$ UCCD $>$ AnomalyDINO (50.0 \%); cDDPM $>$ REFLECT $>$ UCCD $>$ AnomalyDINO (33.3 \%); cDDPM $>$ UCCD $>$ REFLECT $>$ AnomalyDINO (16.7 \%)\\
lesion sensitivity & 8 & UCCD $>$ cDDPM $>$ REFLECT $>$ AnomalyDINO (43.7 \%); REFLECT $>$ UCCD $>$ cDDPM $>$ AnomalyDINO (31.2 \%); UCCD $>$ REFLECT $>$ cDDPM $>$ AnomalyDINO (9.7 \%)\\
slice AUROC, marginal & 2 & UCCD $>$ cDDPM $>$ REFLECT $>$ AnomalyDINO (83.3 \%); UCCD $>$ REFLECT $>$ cDDPM $>$ AnomalyDINO (16.7 \%)\\
slice AROC, adjusted & 2 & cDDPM $>$ UCCD $>$ REFLECT $>$ AnomalyDINO (66.7 \%); REFLECT $>$ cDDPM $>$ UCCD $>$ AnomalyDINO (33.3 \%)\\
\bottomrule
\end{tabular}
\end{table}
\subsection{Operating points and risk control}\label{app:thr}

Table~\ref{tab:gapfull} gives the threshold-transfer gap with the thresholds involved and with the median of the per-subject differences, which agrees with the mean in sign at every budget and method except UCCD at $40\cm$, where both are indistinguishable from zero. Table~\ref{tab:crc} lists every threshold rule at $20\cm$, and Table~\ref{tab:crcall} the exceedance and Dice of each rule at every budget. Conformal risk control (CRC) at $\alpha=0.05$ and RCPS at $\alpha=0.10$ select the same, very strict, thresholds and keep exceedance at or below 0.019, at a large Dice cost. The label-free CRC, which bounds the predicted volume rather than the false-positive volume, is the most conservative rule for UCCD and cDDPM. Table~\ref{tab:splits} gives the random-split results per method that Table~\ref{tab:crcbeta} summarises. The 0.005 threshold grid limits burden matching to a relative mismatch of at most 5.7\%.
\begin{table}[H]
\caption{Threshold-transfer gap: Dice at the test-matched minus Dice at the validation threshold, mean and median of per-subject differences, with the thresholds and the realised burden at the validation threshold.}
\label{tab:gapfull}
\centering\scriptsize
\setlength{\tabcolsep}{3pt}
\resizebox{\textwidth}{!}{%
\begin{tabular}{@{}lllllllll@{}}
\toprule
Method & b (cm$^3$) & validation $\tau$ & test-matched $\tau$ & Realised FP at validation $\tau$ (cm$^3$) & Dice at validation $\tau$ & Dice at test-matched $\tau$ & Threshold-transfer gap, mean [95 \% CI] & Gap, median of per-subject differences [95 \% CI]\\
\midrule
REFLECT & 5 & 0.930 & 0.902 & 3.3 & 0.241 & 0.314 & +0.073 [0.070, 0.077] & +0.082 [0.079, 0.086]\\
REFLECT & 10 & 0.870 & 0.839 & 7.2 & 0.382 & 0.433 & +0.051 [0.049, 0.053] & +0.054 [0.052, 0.055]\\
REFLECT & 20 & 0.785 & 0.761 & 16.3 & 0.497 & 0.517 & +0.020 [0.019, 0.022] & +0.021 [0.019, 0.022]\\
REFLECT & 40 & 0.685 & 0.667 & 35.1 & 0.554 & 0.558 & +0.003 [0.003, 0.004] & +0.003 [0.002, 0.004]\\
UCCD & 5 & 0.500 & 0.507 & 5.3 & 0.441 & 0.433 & $-$0.008 [$-$0.008, $-$0.007] & $-$0.008 [$-$0.008, $-$0.008]\\
UCCD & 10 & 0.445 & 0.442 & 9.6 & 0.497 & 0.499 & +0.003 [0.002, 0.003] & +0.003 [0.003, 0.003]\\
UCCD & 20 & 0.385 & 0.381 & 18.9 & 0.539 & 0.541 & +0.002 [0.002, 0.002] & +0.003 [0.002, 0.003]\\
UCCD & 40 & 0.340 & 0.321 & 31.8 & 0.551 & 0.550 & $-$0.001 [$-$0.003, 0.001] & +0.001 [$-$0.002, 0.003]\\
cDDPM & 5 & 0.530 & 0.487 & 3.1 & 0.396 & 0.444 & +0.048 [0.045, 0.050] & +0.049 [0.047, 0.051]\\
cDDPM & 10 & 0.470 & 0.427 & 6.1 & 0.462 & 0.502 & +0.040 [0.037, 0.043] & +0.045 [0.042, 0.048]\\
cDDPM & 20 & 0.405 & 0.368 & 12.9 & 0.519 & 0.542 & +0.023 [0.020, 0.027] & +0.028 [0.025, 0.033]\\
cDDPM & 40 & 0.390 & 0.308 & 15.4 & 0.530 & 0.557 & +0.028 [0.019, 0.036] & +0.038 [0.029, 0.045]\\
AnomalyDINO & 5 & 0.810 & 0.797 & 3.9 & 0.064 & 0.076 & +0.012 [0.011, 0.014] & +0.008 [0.006, 0.011]\\
AnomalyDINO & 10 & 0.770 & 0.758 & 8.0 & 0.105 & 0.120 & +0.015 [0.014, 0.017] & +0.014 [0.011, 0.016]\\
AnomalyDINO & 20 & 0.725 & 0.717 & 17.5 & 0.166 & 0.177 & +0.012 [0.010, 0.013] & +0.011 [0.010, 0.013]\\
AnomalyDINO & 40 & 0.680 & 0.672 & 35.5 & 0.230 & 0.240 & +0.011 [0.009, 0.012] & +0.009 [0.008, 0.011]\\
\bottomrule
\end{tabular}}
\end{table}
\begin{table}[H]
\caption{Every threshold rule at $20\cm$: test exceedance $\Pr[\FP_i>20\cm]$, mean realised burden and subject-mean Dice.}
\label{tab:crc}
\centering\scriptsize
\setlength{\tabcolsep}{3pt}
\begin{tabular}{@{}llllll@{}}
\toprule
Method & Threshold rule (20 cm$^3$) & $\tau$ & Test exceedance P(FP\_i $>$ 20 cm$^3$) [95 \% CI] & Mean FP (cm$^3$) & Dice, subject mean\\
\midrule
REFLECT & test-matched & 0.761 & 0.228 [0.179, 0.276] & 19.9 & 0.517\\
REFLECT & validation threshold & 0.785 & 0.157 [0.115, 0.199] & 16.3 & 0.497\\
REFLECT & CRC validation $\alpha$ = 0.05 & 0.989 & 0.013 [0.003, 0.026] & 0.6 & 0.040\\
REFLECT & CRC validation $\alpha$ = 0.1 & 0.831 & 0.074 [0.045, 0.103] & 10.8 & 0.444\\
REFLECT & CRC validation $\alpha$ = 0.2 & 0.790 & 0.154 [0.115, 0.196] & 15.7 & 0.492\\
REFLECT & RCPS validation $\alpha$ = 0.1 & 0.989 & 0.013 [0.003, 0.026] & 0.6 & 0.040\\
REFLECT & label-free CRC validation $\alpha$ = 0.1 & 0.982 & 0.013 [0.003, 0.026] & 0.9 & 0.066\\
UCCD & test-matched & 0.381 & 0.298 [0.250, 0.349] & 19.8 & 0.541\\
UCCD & validation threshold & 0.385 & 0.298 [0.250, 0.349] & 18.9 & 0.539\\
UCCD & CRC validation $\alpha$ = 0.05 & 0.617 & 0.010 [0.000, 0.022] & 1.5 & 0.302\\
UCCD & CRC validation $\alpha$ = 0.1 & 0.476 & 0.096 [0.064, 0.128] & 6.9 & 0.466\\
UCCD & CRC validation $\alpha$ = 0.2 & 0.440 & 0.138 [0.099, 0.176] & 10.2 & 0.501\\
UCCD & RCPS validation $\alpha$ = 0.1 & 0.617 & 0.010 [0.000, 0.022] & 1.5 & 0.302\\
UCCD & label-free CRC validation $\alpha$ = 0.1 & 0.805 & 0.000 [0.000, 0.000] & 0.2 & 0.098\\
cDDPM & test-matched & 0.368 & 0.295 [0.244, 0.346] & 19.8 & 0.542\\
cDDPM & validation threshold & 0.405 & 0.215 [0.170, 0.260] & 12.9 & 0.519\\
cDDPM & CRC validation $\alpha$ = 0.05 & 0.649 & 0.006 [0.000, 0.016] & 0.9 & 0.254\\
cDDPM & CRC validation $\alpha$ = 0.1 & 0.524 & 0.032 [0.013, 0.054] & 3.3 & 0.403\\
cDDPM & CRC validation $\alpha$ = 0.2 & 0.475 & 0.080 [0.051, 0.112] & 5.7 & 0.457\\
cDDPM & RCPS validation $\alpha$ = 0.1 & 0.649 & 0.006 [0.000, 0.016] & 0.9 & 0.254\\
cDDPM & label-free CRC validation $\alpha$ = 0.1 & 0.819 & 0.000 [0.000, 0.000] & 0.1 & 0.084\\
AnomalyDINO & test-matched & 0.717 & 0.317 [0.266, 0.369] & 19.9 & 0.177\\
AnomalyDINO & validation threshold & 0.725 & 0.272 [0.224, 0.321] & 17.5 & 0.166\\
AnomalyDINO & CRC validation $\alpha$ = 0.05 & 0.809 & 0.016 [0.003, 0.032] & 3.9 & 0.064\\
AnomalyDINO & CRC validation $\alpha$ = 0.1 & 0.790 & 0.038 [0.019, 0.061] & 5.6 & 0.083\\
AnomalyDINO & CRC validation $\alpha$ = 0.2 & 0.752 & 0.157 [0.119, 0.199] & 11.0 & 0.128\\
AnomalyDINO & RCPS validation $\alpha$ = 0.1 & 0.809 & 0.016 [0.003, 0.032] & 3.9 & 0.064\\
AnomalyDINO & label-free CRC validation $\alpha$ = 0.1 & 0.810 & 0.016 [0.003, 0.032] & 3.9 & 0.064\\
\bottomrule
\end{tabular}
\end{table}
\begin{table}[H]
\caption{Every threshold rule at every budget; each cell gives test exceedance / subject-mean Dice. RCPS at $\alpha=0.05$ is infeasible with 24 calibration subjects and is omitted.}
\label{tab:crcall}
\centering\scriptsize
\setlength{\tabcolsep}{3pt}
\resizebox{\textwidth}{!}{%
\begin{tabular}{@{}lllllllll@{}}
\toprule
Method & b & test-matched & validation & CRC $\alpha$ 0.05 & CRC $\alpha$ 0.10 & CRC $\alpha$ 0.20 & RCPS $\alpha$ 0.10 & label-free CRC $\alpha$ 0.10\\
\midrule
REFLECT & 5 & 0.135 / 0.314 & 0.067 / 0.241 & 0.013 / 0.010 & 0.051 / 0.227 & 0.183 / 0.352 & 0.013 / 0.010 & 0.013 / 0.014\\
REFLECT & 10 & 0.186 / 0.433 & 0.103 / 0.382 & 0.013 / 0.021 & 0.067 / 0.328 & 0.157 / 0.418 & 0.013 / 0.021 & 0.013 / 0.029\\
REFLECT & 20 & 0.228 / 0.517 & 0.157 / 0.497 & 0.013 / 0.040 & 0.074 / 0.444 & 0.154 / 0.492 & 0.013 / 0.040 & 0.013 / 0.066\\
REFLECT & 40 & 0.256 / 0.558 & 0.202 / 0.554 & 0.010 / 0.085 & 0.096 / 0.535 & 0.109 / 0.540 & 0.010 / 0.085 & 0.013 / 0.157\\
UCCD & 5 & 0.231 / 0.433 & 0.244 / 0.441 & 0.010 / 0.113 & 0.074 / 0.310 & 0.138 / 0.384 & 0.010 / 0.113 & 0.000 / 0.027\\
UCCD & 10 & 0.288 / 0.499 & 0.288 / 0.497 & 0.010 / 0.188 & 0.093 / 0.394 & 0.144 / 0.446 & 0.010 / 0.188 & 0.000 / 0.051\\
UCCD & 20 & 0.298 / 0.541 & 0.298 / 0.539 & 0.010 / 0.302 & 0.096 / 0.466 & 0.138 / 0.501 & 0.010 / 0.302 & 0.000 / 0.098\\
UCCD & 40 & 0.340 / 0.550 & 0.266 / 0.551 & 0.019 / 0.433 & 0.090 / 0.525 & 0.128 / 0.541 & 0.019 / 0.433 & 0.003 / 0.257\\
cDDPM & 5 & 0.247 / 0.444 & 0.163 / 0.396 & 0.016 / 0.121 & 0.042 / 0.259 & 0.109 / 0.355 & 0.016 / 0.121 & 0.000 / 0.027\\
cDDPM & 10 & 0.279 / 0.502 & 0.170 / 0.462 & 0.013 / 0.173 & 0.058 / 0.340 & 0.099 / 0.409 & 0.013 / 0.173 & 0.000 / 0.045\\
cDDPM & 20 & 0.295 / 0.542 & 0.215 / 0.519 & 0.006 / 0.254 & 0.032 / 0.403 & 0.080 / 0.457 & 0.006 / 0.254 & 0.000 / 0.084\\
cDDPM & 40 & 0.333 / 0.557 & 0.109 / 0.530 & 0.010 / 0.374 & 0.022 / 0.452 & 0.090 / 0.518 & 0.010 / 0.374 & 0.000 / 0.226\\
AnomalyDINO & 5 & 0.308 / 0.076 & 0.224 / 0.064 & 0.003 / 0.016 & 0.022 / 0.023 & 0.071 / 0.036 & 0.003 / 0.016 & 0.003 / 0.015\\
AnomalyDINO & 10 & 0.314 / 0.120 & 0.237 / 0.105 & 0.019 / 0.039 & 0.035 / 0.051 & 0.151 / 0.078 & 0.019 / 0.039 & 0.010 / 0.032\\
AnomalyDINO & 20 & 0.317 / 0.177 & 0.272 / 0.166 & 0.016 / 0.064 & 0.038 / 0.083 & 0.157 / 0.128 & 0.016 / 0.064 & 0.016 / 0.064\\
AnomalyDINO & 40 & 0.317 / 0.240 & 0.272 / 0.230 & 0.013 / 0.104 & 0.061 / 0.136 & 0.154 / 0.186 & 0.013 / 0.104 & 0.019 / 0.109\\
\bottomrule
\end{tabular}}
\end{table}
\begin{table}[H]
\caption{CRC ($\alpha=0.10$, $20\cm$) over 1,000 random calibration/evaluation splits of the 312 test subjects, per method and calibration size.}
\label{tab:splits}
\centering\scriptsize
\setlength{\tabcolsep}{3pt}
\begin{tabular}{@{}llllll@{}}
\toprule
Method & Calibration n & Mean exceedance & 2.5th to 97.5th percentile over splits & Splits with exceedance $\le$ $\alpha$ & Infeasible splits\\
\midrule
REFLECT & 24 & 0.078 & [0.007, 0.212] & 69.5 \% & 0\\
REFLECT & 50 & 0.097 & [0.031, 0.206] & 59.3 \% & 0\\
REFLECT & 100 & 0.099 & [0.042, 0.175] & 54.9 \% & 0\\
UCCD & 24 & 0.079 & [0.007, 0.215] & 70.1 \% & 0\\
UCCD & 50 & 0.099 & [0.031, 0.206] & 55.6 \% & 0\\
UCCD & 100 & 0.098 & [0.038, 0.175] & 57.7 \% & 0\\
cDDPM & 24 & 0.082 & [0.007, 0.222] & 71.2 \% & 0\\
cDDPM & 50 & 0.098 & [0.027, 0.191] & 55.6 \% & 0\\
cDDPM & 100 & 0.100 & [0.042, 0.179] & 53.7 \% & 0\\
AnomalyDINO & 24 & 0.081 & [0.007, 0.205] & 70.5 \% & 0\\
AnomalyDINO & 50 & 0.097 & [0.027, 0.202] & 57.0 \% & 0\\
AnomalyDINO & 100 & 0.096 & [0.038, 0.170] & 56.6 \% & 0\\
\bottomrule
\end{tabular}
\end{table}
\subsection{Comparison of methods: further contrasts}\label{app:lead}

Table~\ref{tab:pairs} gives every paired Dice contrast between the residual models, including $10\cm$, and Table~\ref{tab:voxpairs} every paired voxel-metric contrast, including those with AnomalyDINO. In voxel AUROC every residual model exceeds AnomalyDINO and REFLECT exceeds both diffusion models under every aggregation; UCCD and cDDPM differ by less than 0.011. In voxel AUPRC the order of the residual models depends on the aggregation: UCCD is below cDDPM in the subject mean and median ($\pH<0.005$), and REFLECT is above UCCD only when voxels are pooled ($+0.062$, $\pH=0.016$). Table~\ref{tab:holm} summarises how many contrasts of each declared family have Holm-adjusted $p<0.05$: at validation thresholds 20 to 22 of the 24 model-pair Dice contrasts do, at equal realised burden only 12 to 16. The slice-level score also depends on how a slice is summarised (Table~\ref{tab:slicescore}): taking the slice mean instead of the maximum lowers marginal slice AUROC by 0.04 to 0.24 and puts AnomalyDINO below chance (0.425).
\begin{table}[H]
\caption{Paired Dice contrasts between residual models at every budget, Holm-adjusted within each 24-contrast family. Val: validation threshold; matched: equal realised burden (interpolated).}
\label{tab:pairs}
\centering\scriptsize
\setlength{\tabcolsep}{3pt}
\resizebox{\textwidth}{!}{%
\begin{tabular}{@{}llllll@{}}
\toprule
Contrast & Operating point & Budget (cm$^3$) & Subject mean & Subject median & Pooled\\
\midrule
REFLECT $-$ UCCD & validation $\tau$ & 5 & $-$0.200 [$-$0.219, $-$0.180], p\_H $<$ 0.005 & $-$0.205 [$-$0.235, $-$0.180], p\_H $<$ 0.005 & $-$0.137 [$-$0.161, $-$0.113], p\_H $<$ 0.005\\
REFLECT $-$ UCCD & validation $\tau$ & 10 & $-$0.115 [$-$0.135, $-$0.095], p\_H $<$ 0.005 & $-$0.097 [$-$0.136, $-$0.057], p\_H $<$ 0.005 & $-$0.046 [$-$0.069, $-$0.025], p\_H $<$ 0.005\\
REFLECT $-$ UCCD & validation $\tau$ & 20 & $-$0.042 [$-$0.060, $-$0.025], p\_H $<$ 0.005 & $-$0.008 [$-$0.040, 0.015], p\_H 0.74 & +0.020 [0.001, 0.039], p\_H 0.070\\
REFLECT $-$ UCCD & validation $\tau$ & 40 & +0.003 [$-$0.012, 0.018], p\_H 0.72 & +0.024 [0.001, 0.048], p\_H 0.14 & +0.056 [0.038, 0.073], p\_H $<$ 0.005\\
REFLECT $-$ UCCD & matched realised FP & 5 & $-$0.118 [$-$0.209, $-$0.048], p\_H $<$ 0.005 & $-$0.100 [$-$0.225, $-$0.014], p\_H 0.23 & $-$0.049 [$-$0.147, 0.023], p\_H 0.79\\
REFLECT $-$ UCCD & matched realised FP & 10 & $-$0.067 [$-$0.129, $-$0.026], p\_H $<$ 0.005 & $-$0.046 [$-$0.121, 0.017], p\_H 1.00 & +0.001 [$-$0.061, 0.043], p\_H 1.00\\
REFLECT $-$ UCCD & matched realised FP & 20 & $-$0.024 [$-$0.056, $-$0.000], p\_H 0.33 & +0.011 [$-$0.031, 0.040], p\_H 1.00 & +0.036 [0.002, 0.062], p\_H 0.38\\
REFLECT $-$ UCCD & matched realised FP & 40 & +0.008 [$-$0.009, 0.023], p\_H 1.00 & +0.018 [0.001, 0.047], p\_H 0.46 & +0.060 [0.041, 0.077], p\_H $<$ 0.005\\
REFLECT $-$ cDDPM & validation $\tau$ & 5 & $-$0.155 [$-$0.175, $-$0.135], p\_H $<$ 0.005 & $-$0.164 [$-$0.194, $-$0.133], p\_H $<$ 0.005 & $-$0.098 [$-$0.121, $-$0.074], p\_H $<$ 0.005\\
REFLECT $-$ cDDPM & validation $\tau$ & 10 & $-$0.080 [$-$0.102, $-$0.058], p\_H $<$ 0.005 & $-$0.065 [$-$0.106, $-$0.023], p\_H 0.007 & $-$0.019 [$-$0.042, 0.004], p\_H 0.11\\
REFLECT $-$ cDDPM & validation $\tau$ & 20 & $-$0.022 [$-$0.042, $-$0.002], p\_H 0.062 & +0.016 [$-$0.019, 0.040], p\_H 0.74 & +0.032 [0.011, 0.053], p\_H 0.016\\
REFLECT $-$ cDDPM & validation $\tau$ & 40 & +0.025 [0.006, 0.043], p\_H 0.021 & +0.055 [0.029, 0.079], p\_H $<$ 0.005 & +0.069 [0.048, 0.090], p\_H $<$ 0.005\\
REFLECT $-$ cDDPM & matched realised FP & 5 & $-$0.128 [$-$0.223, $-$0.056], p\_H $<$ 0.005 & $-$0.124 [$-$0.247, $-$0.030], p\_H 0.055 & $-$0.067 [$-$0.168, 0.008], p\_H 0.70\\
REFLECT $-$ cDDPM & matched realised FP & 10 & $-$0.069 [$-$0.134, $-$0.027], p\_H $<$ 0.005 & $-$0.047 [$-$0.129, 0.013], p\_H 1.00 & $-$0.010 [$-$0.075, 0.034], p\_H 1.00\\
REFLECT $-$ cDDPM & matched realised FP & 20 & $-$0.025 [$-$0.057, $-$0.002], p\_H 0.30 & +0.002 [$-$0.038, 0.030], p\_H 1.00 & +0.028 [$-$0.007, 0.055], p\_H 0.70\\
REFLECT $-$ cDDPM & matched realised FP & 40 & +0.001 [$-$0.016, 0.016], p\_H 1.00 & +0.003 [$-$0.020, 0.039], p\_H 1.00 & +0.052 [0.032, 0.069], p\_H $<$ 0.005\\
UCCD $-$ cDDPM & validation $\tau$ & 5 & +0.044 [0.034, 0.055], p\_H $<$ 0.005 & +0.041 [0.026, 0.061], p\_H $<$ 0.005 & +0.039 [0.029, 0.050], p\_H $<$ 0.005\\
UCCD $-$ cDDPM & validation $\tau$ & 10 & +0.035 [0.025, 0.045], p\_H $<$ 0.005 & +0.032 [0.014, 0.051], p\_H $<$ 0.005 & +0.028 [0.018, 0.038], p\_H $<$ 0.005\\
UCCD $-$ cDDPM & validation $\tau$ & 20 & +0.020 [0.011, 0.029], p\_H $<$ 0.005 & +0.024 [0.006, 0.040], p\_H 0.032 & +0.012 [0.002, 0.022], p\_H 0.054\\
UCCD $-$ cDDPM & validation $\tau$ & 40 & +0.022 [0.012, 0.032], p\_H $<$ 0.005 & +0.030 [0.012, 0.048], p\_H 0.007 & +0.013 [0.002, 0.024], p\_H 0.070\\
UCCD $-$ cDDPM & matched realised FP & 5 & $-$0.010 [$-$0.031, 0.010], p\_H 1.00 & $-$0.023 [$-$0.047, 0.006], p\_H 1.00 & $-$0.018 [$-$0.039, 0.002], p\_H 0.70\\
UCCD $-$ cDDPM & matched realised FP & 10 & $-$0.002 [$-$0.017, 0.012], p\_H 1.00 & $-$0.001 [$-$0.030, 0.018], p\_H 1.00 & $-$0.011 [$-$0.027, 0.004], p\_H 0.79\\
UCCD $-$ cDDPM & matched realised FP & 20 & $-$0.001 [$-$0.012, 0.009], p\_H 1.00 & $-$0.009 [$-$0.025, 0.009], p\_H 1.00 & $-$0.008 [$-$0.020, 0.003], p\_H 0.79\\
UCCD $-$ cDDPM & matched realised FP & 40 & $-$0.007 [$-$0.015, 0.001], p\_H 0.43 & $-$0.015 [$-$0.037, 0.012], p\_H 1.00 & $-$0.009 [$-$0.017, $-$0.000], p\_H 0.44\\
\bottomrule
\end{tabular}}
\end{table}
\begin{table}[H]
\caption{Paired voxel AUROC and AUPRC contrasts, Holm-adjusted within the voxel-pair families (mean and median, 24 contrasts; pooled, 12).}
\label{tab:voxpairs}
\centering\scriptsize
\setlength{\tabcolsep}{3pt}
\resizebox{\textwidth}{!}{%
\begin{tabular}{@{}lllll@{}}
\toprule
Contrast & Metric & Subject mean & Subject median & Pooled\\
\midrule
REFLECT $-$ UCCD & voxel AUROC & +0.067 [0.060, 0.074], p\_H $<$ 0.005 & +0.072 [0.064, 0.080], p\_H $<$ 0.005 & +0.083 [0.076, 0.088], p\_H $<$ 0.005\\
REFLECT $-$ cDDPM & voxel AUROC & +0.059 [0.051, 0.066], p\_H $<$ 0.005 & +0.062 [0.053, 0.073], p\_H $<$ 0.005 & +0.078 [0.071, 0.084], p\_H $<$ 0.005\\
UCCD $-$ cDDPM & voxel AUROC & $-$0.008 [$-$0.011, $-$0.006], p\_H $<$ 0.005 & $-$0.010 [$-$0.015, $-$0.003], p\_H 0.018 & $-$0.005 [$-$0.008, $-$0.002], p\_H 0.016\\
REFLECT $-$ AnomalyDINO & voxel AUROC & +0.100 [0.093, 0.108], p\_H $<$ 0.005 & +0.104 [0.092, 0.112], p\_H $<$ 0.005 & +0.119 [0.108, 0.130], p\_H $<$ 0.005\\
UCCD $-$ AnomalyDINO & voxel AUROC & +0.033 [0.023, 0.043], p\_H $<$ 0.005 & +0.032 [0.018, 0.042], p\_H $<$ 0.005 & +0.036 [0.024, 0.049], p\_H $<$ 0.005\\
cDDPM $-$ AnomalyDINO & voxel AUROC & +0.041 [0.031, 0.052], p\_H $<$ 0.005 & +0.041 [0.026, 0.052], p\_H $<$ 0.005 & +0.041 [0.029, 0.053], p\_H $<$ 0.005\\
REFLECT $-$ UCCD & voxel AUPRC & +0.003 [$-$0.015, 0.022], p\_H 1.00 & +0.034 [0.001, 0.064], p\_H 0.14 & +0.062 [0.021, 0.101], p\_H 0.016\\
REFLECT $-$ cDDPM & voxel AUPRC & $-$0.020 [$-$0.038, $-$0.001], p\_H 0.14 & +0.003 [$-$0.032, 0.029], p\_H 1.00 & +0.048 [0.007, 0.089], p\_H 0.050\\
UCCD $-$ cDDPM & voxel AUPRC & $-$0.023 [$-$0.031, $-$0.014], p\_H $<$ 0.005 & $-$0.031 [$-$0.052, $-$0.018], p\_H $<$ 0.005 & $-$0.013 [$-$0.026, $-$0.001], p\_H 0.050\\
REFLECT $-$ AnomalyDINO & voxel AUPRC & +0.300 [0.280, 0.321], p\_H $<$ 0.005 & +0.386 [0.349, 0.413], p\_H $<$ 0.005 & +0.355 [0.307, 0.401], p\_H $<$ 0.005\\
UCCD $-$ AnomalyDINO & voxel AUPRC & +0.297 [0.277, 0.317], p\_H $<$ 0.005 & +0.352 [0.321, 0.378], p\_H $<$ 0.005 & +0.294 [0.269, 0.318], p\_H $<$ 0.005\\
cDDPM $-$ AnomalyDINO & voxel AUPRC & +0.319 [0.300, 0.339], p\_H $<$ 0.005 & +0.384 [0.357, 0.410], p\_H $<$ 0.005 & +0.307 [0.282, 0.331], p\_H $<$ 0.005\\
\bottomrule
\end{tabular}}
\end{table}
\begin{table}[H]
\caption{Holm family summary: number of contrasts per declared family with raw and Holm-adjusted $p<0.05$.}
\label{tab:holm}
\centering\scriptsize
\setlength{\tabcolsep}{3pt}
\begin{tabular}{@{}llll@{}}
\toprule
Family & Contrasts & raw p $<$ 0.05 & Holm p $<$ 0.05\\
\midrule
Dice, model vs baseline, equal FP, mean & 32 & 31 & 31\\
Dice, model vs baseline, equal FP, median & 32 & 32 & 32\\
Dice, model vs baseline, equal FP, pooled & 32 & 30 & 29\\
Dice, model vs baseline, test-matched, mean & 32 & 31 & 31\\
Dice, model vs baseline, test-matched, median & 32 & 32 & 32\\
Dice, model vs baseline, test-matched, pooled & 32 & 30 & 29\\
Dice, model vs baseline, validation, mean & 32 & 31 & 31\\
Dice, model vs baseline, validation, median & 32 & 32 & 32\\
Dice, model vs baseline, validation, pooled & 32 & 31 & 31\\
Dice, model pairs, equal FP, mean & 24 & 18 & 16\\
Dice, model pairs, equal FP, median & 24 & 15 & 12\\
Dice, model pairs, equal FP, pooled & 24 & 16 & 14\\
Dice, model pairs, test-matched, mean & 24 & 18 & 16\\
Dice, model pairs, test-matched, median & 24 & 15 & 12\\
Dice, model pairs, test-matched, pooled & 24 & 15 & 14\\
Dice, model pairs, validation, mean & 24 & 23 & 22\\
Dice, model pairs, validation, median & 24 & 22 & 21\\
Dice, model pairs, validation, pooled & 24 & 23 & 20\\
voxel, model vs baseline, mean & 16 & 15 & 15\\
voxel, model vs baseline, median & 16 & 15 & 15\\
voxel, model vs baseline, pooled & 16 & 13 & 13\\
voxel, model pairs, mean and median & 24 & 22 & 20\\
voxel, model pairs, pooled & 12 & 12 & 12\\
\bottomrule
\end{tabular}
\end{table}
\begin{table}[H]
\caption{Marginal slice-level AUROC (subject mean, 95\% interval) under three slice summaries of the normalised map.}
\label{tab:slicescore}
\centering\scriptsize
\begin{tabular}{@{}lccc@{}}
\toprule
Method & slice max & slice 99th pct. & slice mean\\
\midrule
REFLECT & 0.887 [0.873, 0.899] & 0.874 [0.860, 0.888] & 0.646 [0.622, 0.671]\\
UCCD & 0.907 [0.896, 0.918] & 0.900 [0.887, 0.912] & 0.869 [0.852, 0.885]\\
cDDPM & 0.905 [0.894, 0.916] & 0.893 [0.880, 0.905] & 0.843 [0.825, 0.861]\\
AnomalyDINO & 0.598 [0.579, 0.618] & 0.580 [0.559, 0.600] & 0.425 [0.403, 0.448]\\
\bottomrule
\end{tabular}
\end{table}
\subsection{Post-processing}\label{app:pp}

The decomposition of Identity~\ref{prop:m4} at every budget (Table~\ref{tab:m4all}) shows the same pattern as at $20\cm$: the relaxation term is positive with an interval excluding zero for every method and budget, whereas the post-processing step at fixed threshold has an interval excluding zero on the positive side in only one case, UCCD at $40\cm$ ($+0.022$). For AnomalyDINO the step at fixed threshold costs up to 0.075 Dice. Table~\ref{tab:sensarm} gives lesion sensitivity for plain thresholding, hysteresis, and hysteresis with pruning, at three hit criteria. Hysteresis alone lowers above-floor sensitivity by at most 0.015; most of the loss comes with the persistence pruning. The purity rule was evaluated on 52,399 (REFLECT), 43,873 (UCCD), 48,525 (cDDPM) and 92,296 (AnomalyDINO) predicted components.
\begin{table}[H]
\caption{Exact decomposition of the gain from hysteresis with pruning at equal realised burden, at every budget (Identity~\ref{prop:m4}).}
\label{tab:m4all}
\centering\scriptsize
\setlength{\tabcolsep}{3pt}
\resizebox{\textwidth}{!}{%
\begin{tabular}{@{}llllllll@{}}
\toprule
Method & b (cm$^3$) & validation $\tau$ (hysteresis + pruning) & thresholding $\tau$ at matched FP & Matched realised FP (cm$^3$) & Total: hysteresis + pruning $-$ thresholding & Post-processing at fixed $\tau$ & Threshold relaxation\\
\midrule
REFLECT & 5 & 0.930 & 0.941 & 2.77 & +0.0219 [+0.0177, +0.0259] & $-$0.0008 [$-$0.0024, +0.0000] & +0.0227 [+0.0188, +0.0265]\\
REFLECT & 10 & 0.870 & 0.901 & 5.04 & +0.0497 [+0.0419, +0.0569] & $-$0.0012 [$-$0.0036, +0.0000] & +0.0510 [+0.0435, +0.0578]\\
REFLECT & 20 & 0.785 & 0.825 & 11.45 & +0.0335 [+0.0232, +0.0428] & $-$0.0128 [$-$0.0216, $-$0.0050] & +0.0463 [+0.0422, +0.0502]\\
REFLECT & 40 & 0.685 & 0.741 & 23.36 & +0.0294 [+0.0163, +0.0414] & +0.0013 [$-$0.0099, +0.0115] & +0.0281 [+0.0245, +0.0317]\\
UCCD & 5 & 0.500 & 0.593 & 1.98 & +0.0954 [+0.0846, +0.1055] & $-$0.0157 [$-$0.0250, $-$0.0072] & +0.1111 [+0.1063, +0.1156]\\
UCCD & 10 & 0.445 & 0.535 & 3.67 & +0.0939 [+0.0825, +0.1047] & $-$0.0079 [$-$0.0177, +0.0010] & +0.1018 [+0.0970, +0.1065]\\
UCCD & 20 & 0.385 & 0.465 & 7.77 & +0.0846 [+0.0727, +0.0952] & +0.0064 [$-$0.0037, +0.0156] & +0.0782 [+0.0732, +0.0830]\\
UCCD & 40 & 0.340 & 0.411 & 14.13 & +0.0748 [+0.0624, +0.0859] & +0.0221 [+0.0124, +0.0311] & +0.0527 [+0.0471, +0.0580]\\
cDDPM & 5 & 0.530 & 0.656 & 0.80 & +0.1308 [+0.1192, +0.1417] & $-$0.0199 [$-$0.0295, $-$0.0112] & +0.1507 [+0.1444, +0.1571]\\
cDDPM & 10 & 0.470 & 0.603 & 1.38 & +0.1421 [+0.1290, +0.1543] & $-$0.0140 [$-$0.0243, $-$0.0047] & +0.1561 [+0.1490, +0.1632]\\
cDDPM & 20 & 0.405 & 0.516 & 3.58 & +0.1156 [+0.1014, +0.1284] & $-$0.0016 [$-$0.0123, +0.0081] & +0.1173 [+0.1096, +0.1247]\\
cDDPM & 40 & 0.390 & 0.486 & 5.05 & +0.1003 [+0.0863, +0.1126] & +0.0035 [$-$0.0069, +0.0128] & +0.0968 [+0.0896, +0.1038]\\
AnomalyDINO & 5 & 0.810 & 0.860 & 1.45 & +0.0046 [$-$0.0002, +0.0098] & $-$0.0068 [$-$0.0100, $-$0.0041] & +0.0114 [+0.0068, +0.0161]\\
AnomalyDINO & 10 & 0.770 & 0.828 & 2.74 & +0.0046 [$-$0.0025, +0.0119] & $-$0.0216 [$-$0.0282, $-$0.0155] & +0.0263 [+0.0206, +0.0321]\\
AnomalyDINO & 20 & 0.725 & 0.792 & 5.39 & $-$0.0007 [$-$0.0110, +0.0097] & $-$0.0443 [$-$0.0545, $-$0.0346] & +0.0436 [+0.0361, +0.0513]\\
AnomalyDINO & 40 & 0.680 & 0.751 & 11.15 & $-$0.0210 [$-$0.0349, $-$0.0070] & $-$0.0749 [$-$0.0891, $-$0.0611] & +0.0540 [+0.0455, +0.0626]\\
\bottomrule
\end{tabular}}
\end{table}
\begin{table}[H]
\caption{Lesion sensitivity by post-processing at $20\cm$, validation threshold: above-floor sensitivity at three IoU criteria, by lesion size, and hits over all 5,473 components (not a sensitivity).}
\label{tab:sensarm}
\centering\scriptsize
\setlength{\tabcolsep}{3pt}
\resizebox{\textwidth}{!}{%
\begin{tabular}{@{}lllllllll@{}}
\toprule
Method & Post-processing & validation $\tau$ & Above-floor sens., IoU $\ge$ 0.10 (n = 472) & IoU $\ge$ 0.25 & IoU $\ge$ 0.50 & Large (n = 307) & Small (n = 165) & Hits / all 5,473 components (not a sensitivity)\\
\midrule
REFLECT & thresholding & 0.785 & 0.608 & 0.479 & 0.222 & 0.915 & 0.036 & 0.052\\
REFLECT & hysteresis & 0.785 & 0.593 & 0.479 & 0.222 & 0.906 & 0.012 & 0.051\\
REFLECT & hysteresis + pruning & 0.785 & 0.549 & 0.460 & 0.220 & 0.844 & 0.000 & 0.047\\
UCCD & thresholding & 0.385 & 0.674 & 0.614 & 0.186 & 0.997 & 0.073 & 0.058\\
UCCD & hysteresis & 0.385 & 0.674 & 0.614 & 0.186 & 0.997 & 0.073 & 0.058\\
UCCD & hysteresis + pruning & 0.385 & 0.629 & 0.591 & 0.174 & 0.954 & 0.024 & 0.054\\
cDDPM & thresholding & 0.405 & 0.665 & 0.585 & 0.180 & 0.984 & 0.073 & 0.057\\
cDDPM & hysteresis & 0.405 & 0.663 & 0.585 & 0.180 & 0.984 & 0.067 & 0.057\\
cDDPM & hysteresis + pruning & 0.405 & 0.619 & 0.549 & 0.172 & 0.925 & 0.048 & 0.053\\
AnomalyDINO & thresholding & 0.725 & 0.210 & 0.066 & 0.006 & 0.316 & 0.012 & 0.018\\
AnomalyDINO & hysteresis & 0.725 & 0.197 & 0.064 & 0.006 & 0.300 & 0.006 & 0.017\\
AnomalyDINO & hysteresis + pruning & 0.725 & 0.121 & 0.047 & 0.006 & 0.182 & 0.006 & 0.010\\
\bottomrule
\end{tabular}}
\end{table}
\subsection{Slice-level AUROC}\label{app:slice}

Table~\ref{tab:slicefull} adds intervals and the spline estimator with 8 degrees of freedom to Table~\ref{tab:slice}; Table~\ref{tab:caliper} varies the caliper. Matching on area and position jointly keeps only 11.3\% of positive slices and makes the covariate-only controls worse (0.915 to 0.929), as the within-subject gradient predicts. The within-subject permutation null gave 0.497 to 0.501.
\begin{table}[H]
\caption{Slice-level AUROC: marginal, covariate-adjusted and against the borrowed-map null, with 95\% intervals.}
\label{tab:slicefull}
\centering\scriptsize
\setlength{\tabcolsep}{3pt}
\resizebox{\textwidth}{!}{%
\begin{tabular}{@{}llllllll@{}}
\toprule
Scorer & Marginal & Stratified (quintiles) & Matched, 0.2 SD & Matched, 0.05 SD & Regression, df 5 & Regression, df 8 & Observed $-$ borrowed-map AROC (matched 0.2 SD)\\
\midrule
REFLECT & 0.887 [0.873, 0.899] & 0.754 [0.728, 0.780] & 0.734 [0.704, 0.765] & 0.726 [0.689, 0.762] & 0.600 [0.563, 0.636] & 0.551 [0.510, 0.593] & +0.229 [0.201, 0.257]\\
UCCD & 0.907 [0.896, 0.918] & 0.759 [0.737, 0.780] & 0.769 [0.745, 0.792] & 0.743 [0.713, 0.773] & 0.649 [0.616, 0.682] & 0.505 [0.466, 0.545] & +0.266 [0.244, 0.288]\\
cDDPM & 0.905 [0.894, 0.916] & 0.762 [0.740, 0.784] & 0.768 [0.744, 0.793] & 0.744 [0.714, 0.774] & 0.646 [0.612, 0.680] & 0.497 [0.457, 0.536] & +0.271 [0.249, 0.293]\\
AnomalyDINO & 0.598 [0.579, 0.618] & 0.619 [0.590, 0.649] & 0.620 [0.589, 0.652] & 0.624 [0.588, 0.662] & 0.504 [0.466, 0.542] & 0.449 [0.409, 0.489] & +0.124 [0.100, 0.148]\\
control: brain area & 0.884 [0.869, 0.897] & 0.691 [0.676, 0.706] & 0.668 [0.654, 0.682] & 0.560 [0.545, 0.576] & 0.626 [0.592, 0.659] & 0.719 [0.691, 0.746] & n/a\\
control: mid-brain closeness & 0.857 [0.839, 0.874] & 0.576 [0.535, 0.619] & 0.550 [0.505, 0.596] & 0.486 [0.437, 0.539] & 0.530 [0.489, 0.569] & 0.545 [0.508, 0.582] & n/a\\
\bottomrule
\end{tabular}}
\end{table}
\begin{table}[H]
\caption{Matched-pair AROC by caliper, and matched on brain area and relative position jointly; last row: matched fraction of positive slices.}
\label{tab:caliper}
\centering\scriptsize
\setlength{\tabcolsep}{3pt}
\begin{tabular}{@{}lllll@{}}
\toprule
Scorer & Matched on area, 0.2 SD & 0.1 SD & 0.05 SD & Matched on area + position, 0.2 SD\\
\midrule
control: brain area & 0.668 [0.654, 0.682] & 0.610 [0.596, 0.624] & 0.560 [0.545, 0.576] & 0.929 [0.906, 0.950]\\
control: mid-brain closeness & 0.550 [0.505, 0.596] & 0.518 [0.471, 0.568] & 0.486 [0.437, 0.539] & 0.915 [0.889, 0.939]\\
REFLECT & 0.734 [0.704, 0.765] & 0.723 [0.690, 0.756] & 0.726 [0.689, 0.762] & 0.577 [0.544, 0.611]\\
UCCD & 0.769 [0.745, 0.792] & 0.748 [0.721, 0.776] & 0.743 [0.713, 0.773] & 0.631 [0.598, 0.664]\\
cDDPM & 0.768 [0.744, 0.793] & 0.750 [0.722, 0.777] & 0.744 [0.714, 0.774] & 0.639 [0.606, 0.671]\\
AnomalyDINO & 0.620 [0.589, 0.652] & 0.617 [0.584, 0.651] & 0.624 [0.588, 0.662] & 0.501 [0.471, 0.532]\\
matched fraction of positive slices & 0.594 & 0.517 & 0.464 & 0.113\\
\bottomrule
\end{tabular}
\end{table}
\subsection{Lesion definition and detection}\label{app:detect}

Table~\ref{tab:lesiondef} gives pooled and subject-mean sensitivity under four lesion definitions; the hit count barely changes with connectivity (at most one lesion), whereas the denominator changes from 440 to 5,473. Above-floor sensitivity rises with the budget for every method except UCCD, which already detects 0.653 of lesions at $5\cm$ (Table~\ref{tab:sensbudget}). Tables~\ref{tab:jaccard} and~\ref{tab:errmodes} give the overlap of missed-lesion sets and the error modes behind Fig.~\ref{fig:detect}a; AnomalyDINO differs from the residual models in that most of its misses are selected but poorly delineated (189 of 373). The label-informed upper bounds at every budget are in Table~\ref{tab:oracle}. A per-subject threshold with no burden cap (Table~\ref{tab:oracleunc}) bounds what any choice of a single threshold per subject could achieve: 0.580 to 0.605 Dice for the residual models and 0.373 for AnomalyDINO, the latter at a mean false-positive volume of $156\cm$. Table~\ref{tab:s3} gives every detection model.
\begin{table}[H]
\caption{Lesion definition and lesion sensitivity at $20\cm$, validation threshold, IoU $\ge0.10$.}
\label{tab:lesiondef}
\centering\scriptsize
\setlength{\tabcolsep}{3pt}
\begin{tabular}{@{}lllllll@{}}
\toprule
Method & Connectivity & Floor (mm$^3$) & Reference components & Hits & Pooled sensitivity & Subject-mean sensitivity\\
\midrule
AnomalyDINO & 6 & 0 & 5473 & 99 & 0.018 & 0.052\\
AnomalyDINO & 6 & 50 & 472 & 99 & 0.210 & 0.264\\
AnomalyDINO & 26 & 0 & 1685 & 100 & 0.059 & 0.127\\
AnomalyDINO & 26 & 50 & 440 & 100 & 0.227 & 0.275\\
REFLECT & 6 & 0 & 5473 & 287 & 0.052 & 0.173\\
REFLECT & 6 & 50 & 472 & 287 & 0.608 & 0.765\\
REFLECT & 26 & 0 & 1685 & 286 & 0.170 & 0.389\\
REFLECT & 26 & 50 & 440 & 286 & 0.650 & 0.785\\
UCCD & 6 & 0 & 5473 & 318 & 0.058 & 0.189\\
UCCD & 6 & 50 & 472 & 318 & 0.674 & 0.841\\
UCCD & 26 & 0 & 1685 & 318 & 0.189 & 0.430\\
UCCD & 26 & 50 & 440 & 318 & 0.723 & 0.868\\
cDDPM & 6 & 0 & 5473 & 314 & 0.057 & 0.187\\
cDDPM & 6 & 50 & 472 & 314 & 0.665 & 0.831\\
cDDPM & 26 & 0 & 1685 & 314 & 0.186 & 0.424\\
cDDPM & 26 & 50 & 440 & 314 & 0.714 & 0.859\\
\bottomrule
\end{tabular}
\end{table}
\begin{table}[H]
\caption{Above-floor lesion sensitivity (IoU $\ge0.10$, plain thresholding, validation threshold) by budget, with large- and small-lesion sensitivity.}
\label{tab:sensbudget}
\centering\scriptsize
\setlength{\tabcolsep}{3pt}
\resizebox{\textwidth}{!}{%
\begin{tabular}{@{}lllll@{}}
\toprule
Method & 5 cm$^3$ & 10 cm$^3$ & 20 cm$^3$ & 40 cm$^3$\\
\midrule
REFLECT & 0.394 (large 0.603, small 0.006) & 0.523 (large 0.801, small 0.006) & 0.608 (large 0.915, small 0.036) & 0.644 (large 0.967, small 0.042)\\
UCCD & 0.653 (large 0.977, small 0.048) & 0.663 (large 0.987, small 0.061) & 0.674 (large 0.997, small 0.073) & 0.669 (large 0.997, small 0.061)\\
cDDPM & 0.614 (large 0.909, small 0.067) & 0.648 (large 0.964, small 0.061) & 0.665 (large 0.984, small 0.073) & 0.672 (large 0.993, small 0.073)\\
AnomalyDINO & 0.040 (large 0.062, small 0.000) & 0.106 (large 0.163, small 0.000) & 0.210 (large 0.316, small 0.012) & 0.339 (large 0.518, small 0.006)\\
\bottomrule
\end{tabular}}
\end{table}
\begin{table}[H]
\caption{Jaccard index between the sets of above-floor lesions missed at $20\cm$, validation threshold.}
\label{tab:jaccard}
\centering\scriptsize
\setlength{\tabcolsep}{3pt}
\begin{tabular}{@{}lllll@{}}
\toprule
Missed-lesion Jaccard & REFLECT & UCCD & cDDPM & AnomalyDINO\\
\midrule
REFLECT & 1.000 & 0.813 & 0.805 & 0.476\\
UCCD & 0.813 & 1.000 & 0.938 & 0.402\\
cDDPM & 0.805 & 0.938 & 1.000 & 0.416\\
AnomalyDINO & 0.476 & 0.402 & 0.416 & 1.000\\
\bottomrule
\end{tabular}
\end{table}
\begin{table}[H]
\caption{Error modes of the 472 above-floor lesions at $20\cm$, validation threshold.}
\label{tab:errmodes}
\centering\scriptsize
\setlength{\tabcolsep}{3pt}
\begin{tabular}{@{}llllll@{}}
\toprule
Method & Detected (IoU $\ge$ 0.10) & Missed: evidence not selected & Missed: no evidence & Missed: selected, poorly delineated & Total\\
\midrule
REFLECT & 287 & 135 & 15 & 35 & 472\\
UCCD & 318 & 123 & 17 & 14 & 472\\
cDDPM & 314 & 137 & 9 & 12 & 472\\
AnomalyDINO & 99 & 170 & 14 & 189 & 472\\
\bottomrule
\end{tabular}
\end{table}
\begin{table}[H]
\caption{Label-informed upper bounds at 5, 20 and $40\cm$: Dice gain over the validation threshold. Every bound keeps $\FP_i\le b$; they use the reference labels and are ceilings, not methods.}
\label{tab:oracle}
\centering\scriptsize
\setlength{\tabcolsep}{3pt}
\resizebox{\textwidth}{!}{%
\begin{tabular}{@{}llllllll@{}}
\toprule
Method & b (cm$^3$) & Dice at validation $\tau$ & Gain: per-subject $\tau$ upper bound & Gain: per-lesion $\tau$ upper bound & Gain: component selection upper bound & Gain: boundary replacement & Within-lesion coverage at boundary rung\\
\midrule
REFLECT & 5 & 0.241 & +0.198 [0.182, 0.214] & +0.353 [0.331, 0.373] & $-$0.008 [$-$0.016, $-$0.001] & +0.349 [0.309, 0.389] & 0.231\\
REFLECT & 20 & 0.497 & +0.042 [0.034, 0.050] & +0.153 [0.138, 0.166] & $-$0.002 [$-$0.017, 0.013] & +0.400 [0.375, 0.425] & 0.439\\
REFLECT & 40 & 0.554 & +0.011 [0.006, 0.015] & +0.108 [0.096, 0.119] & +0.023 [0.002, 0.043] & +0.395 [0.372, 0.417] & 0.560\\
UCCD & 5 & 0.441 & +0.081 [0.068, 0.094] & +0.203 [0.191, 0.215] & +0.001 [$-$0.009, 0.011] & +0.527 [0.509, 0.544] & 0.314\\
UCCD & 20 & 0.539 & +0.029 [0.022, 0.036] & +0.122 [0.113, 0.132] & +0.015 [$-$0.003, 0.031] & +0.452 [0.436, 0.467] & 0.454\\
UCCD & 40 & 0.551 & +0.028 [0.023, 0.033] & +0.113 [0.102, 0.125] & +0.037 [0.019, 0.055] & +0.433 [0.416, 0.450] & 0.516\\
cDDPM & 5 & 0.396 & +0.151 [0.135, 0.168] & +0.258 [0.243, 0.272] & +0.003 [$-$0.005, 0.010] & +0.510 [0.482, 0.536] & 0.285\\
cDDPM & 20 & 0.519 & +0.073 [0.063, 0.083] & +0.157 [0.147, 0.168] & +0.023 [0.008, 0.037] & +0.462 [0.444, 0.479] & 0.424\\
cDDPM & 40 & 0.530 & +0.071 [0.062, 0.080] & +0.151 [0.141, 0.161] & +0.041 [0.029, 0.053] & +0.458 [0.440, 0.475] & 0.441\\
AnomalyDINO & 5 & 0.064 & +0.040 [0.031, 0.051] & +0.177 [0.160, 0.193] & $-$0.001 [$-$0.005, 0.001] & $-$0.003 [$-$0.023, 0.019] & 0.185\\
AnomalyDINO & 20 & 0.166 & +0.068 [0.054, 0.082] & +0.164 [0.149, 0.179] & $-$0.003 [$-$0.013, 0.006] & +0.142 [0.104, 0.181] & 0.241\\
AnomalyDINO & 40 & 0.230 & +0.072 [0.058, 0.086] & +0.126 [0.114, 0.137] & +0.003 [$-$0.011, 0.015] & +0.276 [0.236, 0.317] & 0.296\\
\bottomrule
\end{tabular}}
\end{table}
\begin{table}[H]
\caption{Per-subject optimal threshold without a burden cap: the best Dice attainable by any single threshold per subject, its mean false-positive volume, and its gain over the validation threshold.}
\label{tab:oracleunc}
\centering\scriptsize
\begin{tabular}{@{}lccccc@{}}
\toprule
Method & best attainable Dice & mean FP (cm$^3$) & gain over validation $\tau$ at 5 & at 20 & at 40\\
\midrule
REFLECT & 0.580 & 51.6 & +0.338 [0.326, 0.351] & +0.083 [0.075, 0.091] & +0.025 [0.021, 0.030]\\
UCCD & 0.585 & 31.0 & +0.145 [0.134, 0.156] & +0.046 [0.041, 0.052] & +0.034 [0.030, 0.039]\\
cDDPM & 0.605 & 26.3 & +0.209 [0.195, 0.223] & +0.086 [0.077, 0.095] & +0.076 [0.068, 0.084]\\
AnomalyDINO & 0.373 & 155.8 & +0.309 [0.293, 0.326] & +0.207 [0.191, 0.223] & +0.143 [0.129, 0.158]\\
\bottomrule
\end{tabular}
\end{table}
\begin{table}[H]
\caption{Determinants of detection: log-odds of a hit per standard deviation of each predictor under four estimators (472 lesions in 312 subjects). The Laplace fit did not converge for REFLECT; its estimates are not interpretable.}
\label{tab:s3}
\centering\scriptsize
\setlength{\tabcolsep}{3pt}
\begin{tabular}{@{}lllll@{}}
\toprule
Method & Model & log volume (per SD) & T2 contrast (per SD) & relative z position (per SD)\\
\midrule
REFLECT & Firth, profile likelihood & +4.84 [3.73, 6.24] & +1.98 [1.22, 2.82] & $-$0.05 [$-$0.43, 0.34]\\
REFLECT & GEE exchangeable, robust SE & +5.09 [3.65, 6.52] & +2.02 [1.33, 2.72] & $-$0.04 [$-$0.44, 0.36]\\
REFLECT & GLMM, variational Bayes & +5.04 [4.28, 5.79] & +1.94 [1.34, 2.54] & $-$0.04 [$-$0.43, 0.36]\\
REFLECT & GLMM, lme4 glmer (Laplace) & +24.52 [6.70, 42.35] & +6.53 [$-$2.49, 15.55] & +0.72 [$-$3.81, 5.25]\\
UCCD & Firth, profile likelihood & +6.03 [4.53, 8.31] & +0.76 [$-$0.65, 2.08] & $-$0.33 [$-$1.01, 0.28]\\
UCCD & GEE exchangeable, robust SE & +6.82 [4.30, 9.33] & +0.74 [$-$0.13, 1.62] & $-$0.35 [$-$0.84, 0.13]\\
UCCD & GLMM, variational Bayes & +6.27 [4.85, 7.68] & +0.69 [$-$0.25, 1.62] & $-$0.32 [$-$0.90, 0.26]\\
UCCD & GLMM, lme4 glmer (Laplace) & +6.82 [4.64, 8.99] & +0.74 [$-$0.67, 2.15] & $-$0.35 [$-$1.02, 0.32]\\
cDDPM & Firth, profile likelihood & +4.97 [3.88, 6.48] & +0.72 [$-$0.28, 1.70] & $-$0.25 [$-$0.79, 0.26]\\
cDDPM & GEE exchangeable, robust SE & +5.38 [3.75, 7.00] & +0.72 [0.16, 1.28] & $-$0.26 [$-$0.68, 0.16]\\
cDDPM & GLMM, variational Bayes & +5.33 [4.20, 6.47] & +0.71 [$-$0.09, 1.50] & $-$0.26 [$-$0.76, 0.24]\\
cDDPM & GLMM, lme4 glmer (Laplace) & +5.38 [3.97, 6.79] & +0.72 [$-$0.29, 1.73] & $-$0.26 [$-$0.79, 0.28]\\
AnomalyDINO & Firth, profile likelihood & +1.57 [1.03, 2.26] & +0.29 [0.03, 0.55] & $-$0.28 [$-$0.58, 0.00]\\
AnomalyDINO & GEE exchangeable, robust SE & +1.65 [1.11, 2.18] & +0.29 [0.04, 0.54] & $-$0.29 [$-$0.60, 0.02]\\
AnomalyDINO & GLMM, variational Bayes & +1.72 [1.39, 2.06] & +0.34 [0.09, 0.58] & $-$0.31 [$-$0.61, $-$0.01]\\
AnomalyDINO & GLMM, lme4 glmer (Laplace) & +1.68 [1.00, 2.36] & +0.30 [0.02, 0.58] & $-$0.30 [$-$0.60, 0.01]\\
\bottomrule
\end{tabular}
\end{table}
\subsection{Model-free baselines}\label{app:base}

Tables~\ref{tab:t2all}, \ref{tab:flairall} and~\ref{tab:outrank} give the baseline contrasts at every budget and aggregation and the share of universes in which a baseline outranks a model. T2 intensity almost never outranks a residual model (at most 1.1\% of universes) but outranks AnomalyDINO in 26 to 27\%; FLAIR, which the models never see, outranks the residual models in 25 to 38\% and AnomalyDINO in 95 to 96\%.
\begin{table}[H]
\caption{Model minus z-scored T2 intensity, Dice at matched realised burden, subject mean (unadjusted).}
\label{tab:t2all}
\centering\scriptsize
\setlength{\tabcolsep}{3pt}
\begin{tabular}{@{}lllll@{}}
\toprule
Model $-$ T2 intensity, Dice & 5 cm$^3$ & 10 cm$^3$ & 20 cm$^3$ & 40 cm$^3$\\
\midrule
REFLECT & +0.283 [0.263, 0.302] & +0.383 [0.361, 0.405] & +0.437 [0.415, 0.460] & +0.431 [0.408, 0.453]\\
UCCD & +0.401 [0.386, 0.417] & +0.449 [0.431, 0.467] & +0.461 [0.441, 0.481] & +0.423 [0.401, 0.446]\\
cDDPM & +0.412 [0.395, 0.429] & +0.452 [0.433, 0.470] & +0.463 [0.441, 0.484] & +0.430 [0.406, 0.454]\\
AnomalyDINO & +0.044 [0.031, 0.058] & +0.070 [0.053, 0.088] & +0.098 [0.076, 0.119] & +0.113 [0.088, 0.138]\\
\bottomrule
\end{tabular}
\end{table}
\begin{table}[H]
\caption{FLAIR and T2 intensity minus each model, Holm-adjusted within the baseline families.}
\label{tab:flairall}
\centering\scriptsize
\setlength{\tabcolsep}{3pt}
\resizebox{\textwidth}{!}{%
\begin{tabular}{@{}llllll@{}}
\toprule
Contrast & Voxel AUROC, subject mean & Voxel AUROC, pooled & Voxel AUPRC, subject mean & Voxel AUPRC, pooled & Dice 20 cm$^3$ validation $\tau$, subject mean\\
\midrule
FLAIR $-$ REFLECT & +0.018 [0.009, 0.027], p\_H $<$ 0.005 & $-$0.071 [$-$0.085, $-$0.057], p\_H $<$ 0.005 & +0.105 [0.085, 0.126], p\_H $<$ 0.005 & $-$0.127 [$-$0.190, $-$0.059], p\_H $<$ 0.005 & $-$0.173 [$-$0.205, $-$0.142], p\_H 0.006\\
FLAIR $-$ UCCD & +0.085 [0.076, 0.094], p\_H $<$ 0.005 & +0.012 [$-$0.005, 0.027], p\_H 0.43 & +0.108 [0.087, 0.129], p\_H $<$ 0.005 & $-$0.065 [$-$0.110, $-$0.019], p\_H 0.026 & $-$0.216 [$-$0.247, $-$0.184], p\_H 0.006\\
FLAIR $-$ cDDPM & +0.077 [0.067, 0.086], p\_H $<$ 0.005 & +0.007 [$-$0.009, 0.022], p\_H 0.43 & +0.086 [0.065, 0.106], p\_H $<$ 0.005 & $-$0.078 [$-$0.123, $-$0.033], p\_H 0.007 & $-$0.195 [$-$0.228, $-$0.163], p\_H 0.006\\
FLAIR $-$ AnomalyDINO & +0.118 [0.106, 0.130], p\_H $<$ 0.005 & +0.048 [0.029, 0.066], p\_H $<$ 0.005 & +0.405 [0.381, 0.430], p\_H $<$ 0.005 & +0.229 [0.184, 0.274], p\_H $<$ 0.005 & +0.158 [0.123, 0.192], p\_H 0.006\\
T2 $-$ REFLECT & $-$0.096 [$-$0.103, $-$0.088], p\_H $<$ 0.005 & $-$0.130 [$-$0.143, $-$0.119], p\_H $<$ 0.005 & $-$0.353 [$-$0.373, $-$0.333], p\_H $<$ 0.005 & $-$0.418 [$-$0.464, $-$0.367], p\_H $<$ 0.005 & $-$0.411 [$-$0.434, $-$0.389], p\_H 0.006\\
T2 $-$ UCCD & $-$0.029 [$-$0.039, $-$0.018], p\_H $<$ 0.005 & $-$0.048 [$-$0.062, $-$0.034], p\_H $<$ 0.005 & $-$0.350 [$-$0.371, $-$0.328], p\_H $<$ 0.005 & $-$0.356 [$-$0.382, $-$0.328], p\_H $<$ 0.005 & $-$0.454 [$-$0.474, $-$0.433], p\_H 0.006\\
T2 $-$ cDDPM & $-$0.037 [$-$0.047, $-$0.027], p\_H $<$ 0.005 & $-$0.052 [$-$0.067, $-$0.039], p\_H $<$ 0.005 & $-$0.372 [$-$0.395, $-$0.350], p\_H $<$ 0.005 & $-$0.369 [$-$0.395, $-$0.342], p\_H $<$ 0.005 & $-$0.434 [$-$0.455, $-$0.412], p\_H 0.006\\
T2 $-$ AnomalyDINO & +0.005 [$-$0.006, 0.016], p\_H 0.42 & $-$0.012 [$-$0.027, 0.004], p\_H 0.43 & $-$0.053 [$-$0.073, $-$0.033], p\_H $<$ 0.005 & $-$0.062 [$-$0.088, $-$0.035], p\_H $<$ 0.005 & $-$0.080 [$-$0.102, $-$0.059], p\_H 0.006\\
\bottomrule
\end{tabular}}
\end{table}
\begin{table}[H]
\caption{Share of universes in which a model-free baseline outranks a model.}
\label{tab:outrank}
\centering\scriptsize
\setlength{\tabcolsep}{3pt}
\begin{tabular}{@{}lllll@{}}
\toprule
Universe set & Baseline & Model & Share of universes baseline $>$ model & Tied\\
\midrule
defensible & T2 intensity & REFLECT & 0.1 \% & 1.7 \%\\
defensible & T2 intensity & UCCD & 0.0 \% & 1.7 \%\\
defensible & T2 intensity & cDDPM & 0.2 \% & 1.8 \%\\
defensible & T2 intensity & AnomalyDINO & 27.0 \% & 7.6 \%\\
defensible & FLAIR intensity & REFLECT & 25.1 \% & 2.2 \%\\
defensible & FLAIR intensity & UCCD & 32.5 \% & 2.2 \%\\
defensible & FLAIR intensity & cDDPM & 30.6 \% & 2.2 \%\\
defensible & FLAIR intensity & AnomalyDINO & 95.8 \% & 4.1 \%\\
reference C, all & T2 intensity & REFLECT & 0.0 \% & 1.6 \%\\
reference C, all & T2 intensity & UCCD & 1.0 \% & 1.6 \%\\
reference C, all & T2 intensity & cDDPM & 1.1 \% & 1.5 \%\\
reference C, all & T2 intensity & AnomalyDINO & 26.3 \% & 6.5 \%\\
reference C, all & FLAIR intensity & REFLECT & 28.2 \% & 2.2 \%\\
reference C, all & FLAIR intensity & UCCD & 37.7 \% & 2.2 \%\\
reference C, all & FLAIR intensity & cDDPM & 34.9 \% & 2.2 \%\\
reference C, all & FLAIR intensity & AnomalyDINO & 95.0 \% & 3.4 \%\\
\bottomrule
\end{tabular}
\end{table}
\subsection{Latent aggregation for REFLECT}\label{app:c1}

Table~\ref{tab:c1full} lists every score rule against the published map in both checkpoints, Table~\ref{tab:c1deploy} each rule at its own validation threshold, and Table~\ref{tab:c1gate} the retention criterion for MAD. Every latent aggregation improves on the published map in both checkpoints and under both matchings; latent-only scoring is worse in every setting; the image stream alone is better than the published fusion in all four settings at $20\cm$. The image-only rule is infeasible at $5\cm$ on the validation set, and removing the clip has inconsistent effects.
\begin{table}[H]
\caption{Score rule minus the published REFLECT map, Dice at $20\cm$, matched to the published map's realised burden over the cohort or per subject; REFLECT trained on IXI-HH+Guys and on MOOD+IXI.}
\label{tab:c1full}
\centering\scriptsize
\setlength{\tabcolsep}{3pt}
\resizebox{\textwidth}{!}{%
\begin{tabular}{@{}lllll@{}}
\toprule
Score rule (REFLECT) & IXI-HH+Guys: cohort-matched burden & IXI-HH+Guys: subject-matched burden & MOOD+IXI: cohort-matched & MOOD+IXI: subject-matched\\
\midrule
MAD (L2) & +0.052 [0.045, 0.058] & +0.053 [0.046, 0.060] & +0.038 [0.033, 0.043] & +0.034 [0.028, 0.040]\\
robust max (L4) & +0.030 [0.024, 0.035] & +0.022 [0.016, 0.027] & +0.031 [0.027, 0.036] & +0.022 [0.019, 0.026]\\
Euclidean norm (L3) & +0.029 [0.023, 0.034] & +0.023 [0.018, 0.028] & +0.029 [0.024, 0.033] & +0.022 [0.018, 0.025]\\
whitened mean (L5) & +0.030 [0.024, 0.035] & +0.021 [0.015, 0.026] & +0.027 [0.022, 0.031] & +0.015 [0.012, 0.019]\\
mean (L1) & +0.027 [0.022, 0.033] & +0.018 [0.013, 0.023] & +0.028 [0.024, 0.032] & +0.018 [0.015, 0.022]\\
image only & +0.016 [0.013, 0.020] & +0.014 [0.010, 0.018] & +0.014 [0.011, 0.016] & +0.010 [0.006, 0.014]\\
smoothing 1 mm & +0.021 [0.014, 0.027] & +0.006 [$-$0.001, 0.013] & +0.020 [0.016, 0.024] & +0.007 [0.002, 0.011]\\
no clipping & $-$0.005 [$-$0.016, 0.006] & $-$0.015 [$-$0.025, $-$0.005] & $-$0.014 [$-$0.025, $-$0.004] & +0.009 [0.004, 0.015]\\
latent only & $-$0.063 [$-$0.076, $-$0.049] & $-$0.042 [$-$0.054, $-$0.032] & $-$0.073 [$-$0.087, $-$0.059] & $-$0.038 [$-$0.046, $-$0.029]\\
\bottomrule
\end{tabular}}
\end{table}
\begin{table}[H]
\caption{Score rules at their own validation thresholds (0.01 grid), REFLECT trained on IXI-HH+Guys.}
\label{tab:c1deploy}
\centering\scriptsize
\setlength{\tabcolsep}{3pt}
\resizebox{\textwidth}{!}{%
\begin{tabular}{@{}lllllll@{}}
\toprule
Score rule & b (cm$^3$) & own validation $\tau$ & Test Dice, mean [95 \% CI] & Test Dice, median [95 \% CI] & Test FP mean (cm$^3$) & Above-floor sensitivity\\
\midrule
published map & 5 & 0.93 & 0.241 [0.223, 0.260] & 0.235 [0.207, 0.256] & 3.3 [1.5, 6.0] & 0.394\\
published map & 20 & 0.79 & 0.492 [0.468, 0.516] & 0.542 [0.510, 0.569] & 15.7 [11.6, 21.2] & 0.606\\
published map & 40 & 0.69 & 0.553 [0.530, 0.576] & 0.604 [0.577, 0.628] & 33.9 [28.3, 41.3] & 0.644\\
MAD (L2) & 5 & 0.70 & 0.364 [0.344, 0.384] & 0.385 [0.353, 0.411] & 4.5 [3.0, 6.6] & 0.521\\
MAD (L2) & 20 & 0.57 & 0.544 [0.521, 0.566] & 0.597 [0.570, 0.624] & 15.3 [11.5, 20.2] & 0.631\\
MAD (L2) & 40 & 0.47 & 0.599 [0.577, 0.621] & 0.653 [0.626, 0.680] & 33.3 [27.8, 40.3] & 0.655\\
robust max (L4) & 5 & 0.70 & 0.405 [0.383, 0.427] & 0.438 [0.406, 0.467] & 6.9 [5.0, 9.5] & 0.557\\
robust max (L4) & 20 & 0.59 & 0.539 [0.516, 0.562] & 0.599 [0.565, 0.621] & 18.7 [14.7, 23.9] & 0.623\\
robust max (L4) & 40 & 0.51 & 0.580 [0.556, 0.603] & 0.633 [0.612, 0.660] & 34.5 [29.1, 41.5] & 0.648\\
mean (L1) & 5 & 0.70 & 0.418 [0.396, 0.440] & 0.453 [0.424, 0.482] & 7.2 [5.5, 9.5] & 0.564\\
mean (L1) & 20 & 0.60 & 0.537 [0.513, 0.560] & 0.598 [0.562, 0.627] & 18.4 [14.8, 23.1] & 0.623\\
mean (L1) & 40 & 0.52 & 0.577 [0.553, 0.600] & 0.631 [0.608, 0.652] & 34.5 [29.4, 40.9] & 0.650\\
image only & 5 & none & infeasible on validation &  &  & \\
image only & 20 & 0.83 & 0.508 [0.483, 0.533] & 0.559 [0.528, 0.587] & 16.0 [12.0, 21.3] & 0.600\\
image only & 40 & 0.70 & 0.562 [0.539, 0.585] & 0.611 [0.587, 0.634] & 33.5 [27.9, 40.7] & 0.646\\
latent only & 5 & 0.55 & 0.342 [0.322, 0.361] & 0.357 [0.325, 0.379] & 7.7 [6.5, 9.0] & 0.530\\
latent only & 20 & 0.43 & 0.480 [0.457, 0.501] & 0.516 [0.483, 0.558] & 25.1 [22.0, 28.5] & 0.617\\
latent only & 40 & 0.37 & 0.519 [0.496, 0.541] & 0.566 [0.539, 0.587] & 45.7 [40.5, 51.3] & 0.631\\
\bottomrule
\end{tabular}}
\end{table}
\begin{table}[H]
\caption{Retention criterion for MAD (REFLECT, IXI-HH+Guys): MAD is retained only if its interval does not overlap that of the best other aggregation. Growth: 1 with hysteresis, 0 without.}
\label{tab:c1gate}
\centering\scriptsize
\setlength{\tabcolsep}{3pt}
\begin{tabular}{@{}lllllll@{}}
\toprule
Matching & Growth & b (cm$^3$) & MAD $-$ published & Best other family member & Best other $-$ published & Verdict\\
\midrule
cohort & 0 & 5 & +0.068 [0.056, 0.079] & whitened mean (L5) & +0.073 [0.062, 0.085] & MAD demoted\\
cohort & 0 & 10 & +0.048 [0.039, 0.058] & whitened mean (L5) & +0.040 [0.031, 0.049] & MAD demoted\\
cohort & 0 & 20 & +0.052 [0.045, 0.058] & robust max (L4) & +0.030 [0.024, 0.035] & MAD retained\\
cohort & 0 & 40 & +0.046 [0.042, 0.051] & robust max (L4) & +0.027 [0.023, 0.030] & MAD retained\\
cohort & 1 & 5 & +0.057 [0.043, 0.070] & Euclidean norm (L3) & +0.059 [0.045, 0.072] & MAD demoted\\
cohort & 1 & 10 & +0.034 [0.023, 0.046] & whitened mean (L5) & +0.010 [$-$0.000, 0.021] & MAD retained\\
cohort & 1 & 20 & +0.051 [0.044, 0.057] & whitened mean (L5) & +0.028 [0.022, 0.034] & MAD retained\\
cohort & 1 & 40 & +0.047 [0.042, 0.052] & robust max (L4) & +0.028 [0.024, 0.032] & MAD retained\\
subject & 0 & 5 & +0.026 [0.010, 0.042] & Euclidean norm (L3) & +0.005 [$-$0.010, 0.019] & MAD demoted\\
subject & 0 & 10 & +0.042 [0.028, 0.055] & Euclidean norm (L3) & +0.006 [$-$0.007, 0.017] & MAD retained\\
subject & 0 & 20 & +0.053 [0.046, 0.060] & Euclidean norm (L3) & +0.023 [0.018, 0.028] & MAD retained\\
subject & 0 & 40 & +0.045 [0.041, 0.049] & robust max (L4) & +0.024 [0.021, 0.027] & MAD retained\\
subject & 1 & 5 & +0.000 [$-$0.016, 0.017] & Euclidean norm (L3) & +0.008 [$-$0.006, 0.022] & MAD demoted\\
subject & 1 & 10 & $-$0.007 [$-$0.023, 0.010] & published map & +0.000 [0.000, 0.000] & MAD demoted\\
subject & 1 & 20 & +0.035 [0.026, 0.045] & Euclidean norm (L3) & +0.008 [0.000, 0.016] & MAD retained\\
subject & 1 & 40 & +0.044 [0.039, 0.048] & robust max (L4) & +0.023 [0.020, 0.026] & MAD retained\\
\bottomrule
\end{tabular}
\end{table}
On maps produced with the uncorrected mapping, each small lesion was compared with an equal-area region of healthy-appearing tissue around it (Table~\ref{tab:j2}). Local AUPRC was weakly above chance for REFLECT trained on MOOD+IXI and cDDPM trained on IXI-HH+Guys, and indistinguishable from chance for UCCD trained on IXI-HH when small lesions were defined in pixels; cDDPM's local AUROC interval included 0.5. Because these maps predate the corrected geometry, the analysis is descriptive only.
\begin{table}[H]
\caption{Local contrast of small lesions against an equal-area surrounding region (maps produced with the uncorrected mapping; chance level 0.5). Small lesions defined in pixels or in physical volume.}
\label{tab:j2}
\centering\scriptsize
\setlength{\tabcolsep}{3pt}
\begin{tabular}{@{}llllll@{}}
\toprule
Model (training set) & Small-lesion definition & Lesions & Subjects & Local AUPRC [95 \% CI] (chance 0.5) & Local AUROC [95 \% CI]\\
\midrule
REFLECT (MOOD+IXI) & pixel & 122 & 71 & 0.531 [0.511, 0.554] & 0.537 [0.516, 0.558]\\
REFLECT (MOOD+IXI) & physical & 133 & 77 & 0.557 [0.533, 0.584] & 0.561 [0.538, 0.587]\\
cDDPM (IXI-HH+Guys) & pixel & 100 & 62 & 0.535 [0.504, 0.567] & 0.484 [0.432, 0.536]\\
cDDPM (IXI-HH+Guys) & physical & 127 & 81 & 0.584 [0.552, 0.617] & 0.538 [0.491, 0.584]\\
UCCD (IXI-HH) & pixel & 120 & 74 & 0.509 [0.484, 0.536] & 0.495 [0.462, 0.528]\\
UCCD (IXI-HH) & physical & 136 & 85 & 0.544 [0.514, 0.575] & 0.527 [0.493, 0.562]\\
\bottomrule
\end{tabular}
\end{table}
\subsection{Computational cost}\label{app:compute}

Table~\ref{tab:compute} gives the recorded wall-clock and GPU time on Tesla V100 GPUs, including two methods that were also trained but not carried forward (IterMask3D~\cite{liang2025itermask3d} and Dinomaly2~\cite{guo2025dinomaly2}). Training times for UCCD and cDDPM could not be verified from the surviving records and are omitted, and the REFLECT training run listed predates the checkpoint used here, so it indicates the order of magnitude only. Evaluation itself needs no GPU: every MIRTO analysis runs on stored maps. One inference over 24 validation volumes (UCCD) drew 89.9\,W on average, against the 300\,W rating of the GPU.
\begin{table}[H]
\caption{Computational cost (Tesla V100-SXM2-32GB). Dinomaly2 times are approximate and were not taken from job accounting.}
\label{tab:compute}
\centering\scriptsize
\setlength{\tabcolsep}{4pt}
\begin{tabular}{@{}lllll@{}}
\toprule
Method & Phase & Wall clock & GPU-hours & s per volume\\
\midrule
REFLECT & train (stage 1) & 6d 17:25:53 & 484.29 & \\
REFLECT & train (stage 2) & 4d 07:31:24 & 207.05 & \\
REFLECT & inference & 03:07:18 & 3.12 & 32.5\\
AnomalyDINO & memory-native build & 00:01:07 & 0.02 & \\
AnomalyDINO & inference & 00:15:23 & 0.26 & 2.7\\
IterMask3D & train & 09:15:03 & 18.50 & \\
IterMask3D & inference & 02:43:04 & 2.72 & 28.3\\
Dinomaly2 & train (per configuration) & 4h50m to 16h39m, scaling with training-set size, single V100 & 4.83 to 16.65 & \\
Dinomaly2 & calibration (per dataset) & ~00:15:00 & 0.25 & \\
Dinomaly2 & inference (per dataset) & ~00:15:00 & 0.25 & 2.6\\
\bottomrule
\end{tabular}
\end{table}

\section{Training data}\label{app:config}

\begin{table}[h]
\caption{Dice in the reference universe ($20\cm$, validation threshold, subject mean) when all four methods share one training set, and the REFLECT $-$ UCCD contrast. T2 and FLAIR do not depend on training data (0.085 and 0.324).}
\label{tab:config}
\centering\scriptsize
\begin{tabular}{@{}l l cccc l@{}}
\toprule
Training set & Sources & REFLECT & UCCD & cDDPM & AnomalyDINO & REFLECT $-$ UCCD\\
\midrule
IXI-HH & IXI Hammersmith, T2 & 0.425 & 0.500 & 0.467 & 0.148 & \\
IXI-Guys & IXI Guy's, T2 & 0.407 & 0.576 & 0.507 & 0.158 & \\
IXI-HH+Guys & IXI, both sites, T2 & 0.497 & 0.539 & 0.519 & 0.166 & $-0.042$ [$-0.060$, $-0.025$]\\
MOOD+IXI-HH & MOOD, IXI Hammersmith, T1 + T2 & 0.521 & 0.496 & 0.417 & 0.152 & \\
MOOD+IXI & MOOD, all IXI, T1 + T2 & 0.552 & 0.452 & 0.216 & 0.161 & $+0.100$ [0.079, 0.121]\\
\bottomrule
\end{tabular}
\end{table}

Table~\ref{tab:config} summarises the comparison with a common training set. Within IXI-HH+Guys the methods kept their own training protocols: REFLECT trained on all 504 IXI T2 volumes of the pool (stage 1, 200 epochs; re-flow, 50 epochs); UCCD on 324 training and 81 validation volumes (encoder and diffusion model taken at their best validation loss); cDDPM on 361 training and 92 validation volumes (best validation loss at epoch 1,426 of 1,600, 2,854 optimisation steps); AnomalyDINO built its memory bank from 50 reference volumes with slice stride 2. The optimisation budget of a single method varies strongly between training sets (cDDPM: 2,854 steps on IXI-HH+Guys, 79,040 on MOOD+IXI), so effects of the training set are effects of the training data together with the protocol that came with it. One model was trained per method and training set, so variance between training runs is not estimated; it is the most plausible competing explanation for the smaller differences between methods. The ranking-instability and evaluation-dominance hypotheses were supported with every common training set ($\phi_{\max}$ from 0.295 to 0.485).

\paragraph{Further results on training data.} Table~\ref{tab:cfgorder} gives, for every comparison set, the share of defensible universes in which the four-method order differs from the order obtained with IXI-HH+Guys, by metric; REFLECT keeps the highest voxel AUROC in every set, but the order of the other three changes. With the training set as an eleventh axis (Table~\ref{tab:cfgvar}) the method $\times$ training-set interaction carries 0.115 of Dice variance and 0.142 of voxel AUPRC variance, against 0.028 and 0.056 for the main effect of the training set, so the training data mainly change \emph{which} method benefits. Table~\ref{tab:matchedarm} gives the contrasts with a common training set for all three metrics. Table~\ref{tab:allcfg} lists all twenty method--training-set combinations with the six selection criteria and the test regret of each; IXI-HH+Guys was the best training set on test for cDDPM and AnomalyDINO, and cost REFLECT 0.055 and UCCD 0.037 Dice. Table~\ref{tab:selstab} gives the stability of each selection rule. Selecting per method with the threshold refitted inside leave-one-subject-out folds of the validation set chooses REFLECT on MOOD+IXI, UCCD and cDDPM on IXI-Guys and AnomalyDINO on IXI-HH+Guys, and reproduces the test-selected order (UCCD 0.576 $>$ REFLECT 0.552 $>$ cDDPM 0.507 $>$ AnomalyDINO 0.166); that joint choice, however, occurs in only 22.0\% of validation resamples, and its four-method order differs from the IXI-HH+Guys order in 81.4\% of defensible universes. Table~\ref{tab:Hsets} gives the verdicts of H1 to H9 in every set of training data: only test-tuning optimism and conformal control change verdict between sets.

In the test-selected set (REFLECT on MOOD+IXI, UCCD on IXI-Guys, cDDPM on IXI-HH+Guys, AnomalyDINO on MOOD+IXI-HH), which the main text replaces, the reference-universe order was UCCD 0.576 $>$ REFLECT 0.552 $>$ cDDPM 0.519 $>$ AnomalyDINO 0.152; REFLECT $-$ UCCD Dice at $20\cm$ was $-0.025\ci{-0.041}{-0.008}$ ($\pH=0.008$) at validation thresholds and $-0.009\ci{-0.031}{0.010}$ at equal realised burden; the threshold-transfer gap at $20\cm$ was $+0.023\ci{0.022}{0.025}$ for REFLECT and $+0.008\ci{0.006}{0.010}$ for UCCD; and the four-method order agreed with the reference universe in 15.6\% of universes (mean Kendall $\tau_b$ 0.714). A provisional set (REFLECT on MOOD+IXI, UCCD on IXI-HH, cDDPM on IXI-HH+Guys, AnomalyDINO on MOOD+IXI) and the validation-selected set (chosen by median validation slice Dice) appear in Tables~\ref{tab:cfgorder} and~\ref{tab:Hsets}.
\begin{table}[H]
\caption{Share of defensible universes in which the four-method order differs from the order with IXI-HH+Guys, per comparison set and metric.}
\label{tab:cfgorder}
\centering\scriptsize
\setlength{\tabcolsep}{3pt}
\resizebox{\textwidth}{!}{%
\begin{tabular}{@{}lllllllll@{}}
\toprule
Comparison set vs IXI-HH+Guys & Configurations & All metrics & Dice & Voxel AUROC & Voxel AUPRC & Lesion sens. & Slice AUROC marg. & Slice AUROC adj.\\
\midrule
test-selected set & REFLECT (MOOD+IXI), UCCD (IXI-Guys), cDDPM (IXI-HH+Guys), AnomalyDINO (MOOD+IXI-HH) & 69.3 \% & 70.8 \% & 100.0 \% & 100.0 \% & 28.3 \% & 16.7 \% & 100.0 \%\\
provisional set & REFLECT (MOOD+IXI), UCCD (IXI-HH), cDDPM (IXI-HH+Guys), AnomalyDINO (MOOD+IXI) & 50.4 \% & 81.5 \% & 0.0 \% & 50.0 \% & 54.0 \% & 83.3 \% & 33.3 \%\\
validation-selected set & REFLECT (IXI-HH+Guys), UCCD (IXI-Guys), cDDPM (IXI-HH+Guys), AnomalyDINO (MOOD+IXI-HH) & 62.0 \% & 50.5 \% & 100.0 \% & 100.0 \% & 21.5 \% & 0.0 \% & 100.0 \%\\
validation-tuned set & REFLECT (MOOD+IXI), UCCD (IXI-Guys), cDDPM (IXI-Guys), AnomalyDINO (IXI-HH+Guys) & 81.4 \% & 71.8 \% & 100.0 \% & 100.0 \% & 33.2 \% & 83.3 \% & 100.0 \%\\
all IXI-HH & all IXI-HH & 79.0 \% & 71.8 \% & 100.0 \% & 100.0 \% & 52.3 \% & 100.0 \% & 50.0 \%\\
all IXI-Guys & all IXI-Guys & 71.9 \% & 75.0 \% & 100.0 \% & 100.0 \% & 39.7 \% & 16.7 \% & 100.0 \%\\
all MOOD+IXI-HH & all MOOD+IXI-HH & 86.3 \% & 69.4 \% & 100.0 \% & 100.0 \% & 48.4 \% & 100.0 \% & 100.0 \%\\
all MOOD+IXI & all MOOD+IXI & 91.5 \% & 77.3 \% & 100.0 \% & 100.0 \% & 71.6 \% & 100.0 \% & 100.0 \%\\
\bottomrule
\end{tabular}}
\end{table}
\begin{table}[H]
\caption{Variance shares with the training set as an eleventh axis (five training sets; defensible universes, 1,000 resamples).}
\label{tab:cfgvar}
\centering\scriptsize
\setlength{\tabcolsep}{3pt}
\begin{tabular}{@{}llll@{}}
\toprule
Metric subset & Method share & Configuration main effect & Method $\times$ configuration\\
\midrule
all metrics & 0.129 [0.116, 0.141] & 0.005 [0.004, 0.006] & 0.018 [0.016, 0.020]\\
Dice & 0.617 [0.588, 0.637] & 0.028 [0.022, 0.033] & 0.115 [0.104, 0.127]\\
voxel AUROC & 0.849 [0.825, 0.868] & 0.043 [0.033, 0.053] & 0.068 [0.057, 0.080]\\
voxel AUPRC & 0.764 [0.734, 0.786] & 0.056 [0.045, 0.067] & 0.142 [0.125, 0.161]\\
lesion sensitivity & 0.118 [0.108, 0.128] & 0.004 [0.003, 0.005] & 0.020 [0.018, 0.022]\\
slice AUROC, marginal & 0.919 [0.901, 0.931] & 0.006 [0.004, 0.008] & 0.038 [0.031, 0.048]\\
slice AROC, adjusted & 0.607 [0.507, 0.682] & 0.036 [0.021, 0.049] & 0.097 [0.066, 0.131]\\
\bottomrule
\end{tabular}
\end{table}
\begin{table}[H]
\caption{Paired contrasts with a common training set (all methods on IXI-HH+Guys, all on MOOD+IXI), subject mean at $20\cm$, validation threshold, Holm-adjusted. On MOOD+IXI cDDPM was trained on 72\% T1 volumes.}
\label{tab:matchedarm}
\centering\scriptsize
\setlength{\tabcolsep}{3pt}
\begin{tabular}{@{}lllll@{}}
\toprule
Common training set & Metric (subject mean) & Contrast & Estimate [95 \% CI] & p\_Holm\\
\midrule
all IXI-HH+Guys & Dice & REFLECT $-$ UCCD & $-$0.042 [$-$0.060, $-$0.025] & $<$ 0.005\\
all IXI-HH+Guys & Dice & REFLECT $-$ cDDPM & $-$0.022 [$-$0.042, $-$0.002] & 0.062\\
all IXI-HH+Guys & Dice & UCCD $-$ cDDPM & +0.020 [0.011, 0.029] & $<$ 0.005\\
all IXI-HH+Guys & voxel AUROC & REFLECT $-$ UCCD & +0.067 [0.060, 0.074] & $<$ 0.005\\
all IXI-HH+Guys & voxel AUROC & REFLECT $-$ cDDPM & +0.059 [0.051, 0.066] & $<$ 0.005\\
all IXI-HH+Guys & voxel AUROC & UCCD $-$ cDDPM & $-$0.008 [$-$0.011, $-$0.006] & $<$ 0.005\\
all IXI-HH+Guys & voxel AUPRC & REFLECT $-$ UCCD & +0.003 [$-$0.015, 0.022] & 1.00\\
all IXI-HH+Guys & voxel AUPRC & REFLECT $-$ cDDPM & $-$0.020 [$-$0.038, $-$0.001] & 0.14\\
all IXI-HH+Guys & voxel AUPRC & UCCD $-$ cDDPM & $-$0.023 [$-$0.031, $-$0.014] & $<$ 0.005\\
all MOOD+IXI & Dice & REFLECT $-$ UCCD & +0.100 [0.079, 0.121] & $<$ 0.005\\
all MOOD+IXI & Dice & REFLECT $-$ cDDPM & +0.336 [0.312, 0.360] & $<$ 0.005\\
all MOOD+IXI & Dice & UCCD $-$ cDDPM & +0.236 [0.219, 0.253] & $<$ 0.005\\
all MOOD+IXI & voxel AUROC & REFLECT $-$ UCCD & +0.091 [0.083, 0.100] & $<$ 0.005\\
all MOOD+IXI & voxel AUROC & REFLECT $-$ cDDPM & +0.133 [0.124, 0.142] & $<$ 0.005\\
all MOOD+IXI & voxel AUROC & UCCD $-$ cDDPM & +0.042 [0.037, 0.046] & $<$ 0.005\\
all MOOD+IXI & voxel AUPRC & REFLECT $-$ UCCD & +0.136 [0.114, 0.158] & $<$ 0.005\\
all MOOD+IXI & voxel AUPRC & REFLECT $-$ cDDPM & +0.366 [0.345, 0.387] & $<$ 0.005\\
all MOOD+IXI & voxel AUPRC & UCCD $-$ cDDPM & +0.230 [0.214, 0.246] & $<$ 0.005\\
\bottomrule
\end{tabular}
\end{table}
\begin{table}[H]
\caption{All twenty method--training-set combinations: validation criteria of the six selection rules, test Dice and AUPRC, and test regret relative to the method's best training set (Dice at $20\cm$, validation threshold, subject mean). ``test'' denotes the median-slice-Dice rule applied to the test cohort.}
\label{tab:allcfg}
\centering\scriptsize
\setlength{\tabcolsep}{3pt}
\resizebox{\textwidth}{!}{%
\begin{tabular}{@{}lllllllllllll@{}}
\toprule
Method & Config & validation median slice Dice [95 \% CI] & validation pooled Dice & validation median volume Dice & validation canonical Dice 20 cm$^3$ & validation canonical AUPRC & test canonical Dice 20 cm$^3$ (validation $\tau$) & test voxel AUPRC & test regret vs best config & chosen by & primary & flag\\
\midrule
AnomalyDINO & IXI-HH & 0.198 [0.159, 0.232] & 0.319 & 0.309 & 0.176 & 0.293 & 0.148 & 0.312 & 0.018 & AUPRC &  & \\
AnomalyDINO & IXI-Guys & 0.190 [0.155, 0.228] & 0.317 & 0.305 & 0.175 & 0.277 & 0.158 & 0.310 & 0.007 & none &  & \\
AnomalyDINO & IXI-HH+Guys & 0.227 [0.195, 0.251] & 0.320 & 0.297 & 0.189 & 0.286 & 0.166 & 0.311 & 0.000 & pooled Dice, Dice at 20 cm$^3$ & yes & \\
AnomalyDINO & MOOD+IXI-HH & 0.247 [0.218, 0.273] & 0.309 & 0.291 & 0.164 & 0.261 & 0.152 & 0.297 & 0.013 & slice Dice, test &  & \\
AnomalyDINO & MOOD+IXI & 0.169 [0.131, 0.208] & 0.309 & 0.316 & 0.180 & 0.273 & 0.161 & 0.289 & 0.005 & volume Dice &  & \\
REFLECT & IXI-HH & 0.538 [0.516, 0.564] & 0.524 & 0.479 & 0.404 & 0.509 & 0.425 & 0.569 & 0.127 & none &  & \\
REFLECT & IXI-Guys & 0.526 [0.499, 0.548] & 0.518 & 0.496 & 0.377 & 0.509 & 0.407 & 0.585 & 0.145 & none &  & \\
REFLECT & IXI-HH+Guys & 0.561 [0.537, 0.579] & 0.574 & 0.543 & 0.474 & 0.567 & 0.497 & 0.611 & 0.055 & slice Dice & yes & \\
REFLECT & MOOD+IXI-HH & 0.542 [0.522, 0.567] & 0.572 & 0.532 & 0.487 & 0.601 & 0.521 & 0.645 & 0.031 & none &  & \\
REFLECT & MOOD+IXI & 0.560 [0.536, 0.588] & 0.599 & 0.585 & 0.498 & 0.605 & 0.552 & 0.663 & 0.000 & pooled Dice, volume Dice, Dice at 20 cm$^3$, AUPRC, test &  & \\
UCCD & IXI-HH & 0.526 [0.503, 0.549] & 0.538 & 0.525 & 0.488 & 0.550 & 0.500 & 0.569 & 0.076 & none &  & \\
UCCD & IXI-Guys & 0.618 [0.595, 0.639] & 0.605 & 0.554 & 0.531 & 0.621 & 0.576 & 0.655 & 0.000 & slice Dice, pooled Dice, volume Dice, Dice at 20 cm$^3$, AUPRC, test &  & \\
UCCD & IXI-HH+Guys & 0.569 [0.543, 0.592] & 0.563 & 0.548 & 0.505 & 0.573 & 0.539 & 0.608 & 0.037 & none & yes & \\
UCCD & MOOD+IXI-HH & 0.507 [0.480, 0.530] & 0.534 & 0.495 & 0.477 & 0.535 & 0.496 & 0.565 & 0.080 & none &  & \\
UCCD & MOOD+IXI & 0.481 [0.463, 0.505] & 0.494 & 0.450 & 0.460 & 0.517 & 0.452 & 0.526 & 0.125 & none &  & \\
cDDPM & IXI-HH & 0.427 [0.368, 0.498] & 0.528 & 0.522 & 0.458 & 0.518 & 0.467 & 0.550 & 0.052 & none &  & \\
cDDPM & IXI-Guys & 0.457 [0.363, 0.524] & 0.569 & 0.567 & 0.496 & 0.564 & 0.507 & 0.587 & 0.012 & pooled Dice, volume Dice, Dice at 20 cm$^3$ &  & \\
cDDPM & IXI-HH+Guys & 0.482 [0.414, 0.538] & 0.562 & 0.539 & 0.471 & 0.586 & 0.519 & 0.630 & 0.000 & slice Dice, AUPRC, test & yes & \\
cDDPM & MOOD+IXI-HH & 0.365 [0.319, 0.420] & 0.484 & 0.491 & 0.409 & 0.461 & 0.417 & 0.483 & 0.102 & none &  & \\
cDDPM & MOOD+IXI & 0.345 [0.311, 0.381] & 0.351 & 0.345 & 0.243 & 0.289 & 0.216 & 0.297 & 0.303 & none &  & multi-contrast training (72 \% T1); training completed\\
\bottomrule
\end{tabular}}
\end{table}
\begin{table}[H]
\caption{Training-set selection on the 24 validation subjects: point choice, its frequency over $10^4$ bootstrap resamples and its test regret.}
\label{tab:selstab}
\centering\scriptsize
\setlength{\tabcolsep}{3pt}
\resizebox{\textwidth}{!}{%
\begin{tabular}{@{}llllllll@{}}
\toprule
Selection rule & Method & Point choice & Bootstrap frequency of point choice & Test regret of point choice & Mean test regret over resamples & 2.5th to 97.5th pct & Label\\
\midrule
validation median slice Dice & REFLECT & IXI-HH+Guys & 36.8 \% & 0.055 & 0.031 & [0.000, 0.055] & unstable\\
validation median slice Dice & UCCD & IXI-Guys & 97.8 \% & 0.000 & 0.001 & [0.000, 0.000] & stable\\
validation median slice Dice & cDDPM & IXI-HH+Guys & 57.1 \% & 0.000 & 0.010 & [0.000, 0.052] & unstable\\
validation median slice Dice & AnomalyDINO & MOOD+IXI-HH & 27.3 \% & 0.013 & 0.008 & [0.000, 0.018] & unstable\\
validation pooled Dice & REFLECT & MOOD+IXI & 86.4 \% & 0.000 & 0.006 & [0.000, 0.055] & stable\\
validation pooled Dice & UCCD & IXI-Guys & 99.3 \% & 0.000 & 0.000 & [0.000, 0.000] & stable\\
validation pooled Dice & cDDPM & IXI-Guys & 62.8 \% & 0.012 & 0.008 & [0.000, 0.012] & unstable\\
validation pooled Dice & AnomalyDINO & IXI-HH+Guys & 36.4 \% & 0.000 & 0.010 & [0.000, 0.018] & unstable\\
validation median volume Dice & REFLECT & MOOD+IXI & 85.2 \% & 0.000 & 0.006 & [0.000, 0.055] & stable\\
validation median volume Dice & UCCD & IXI-Guys & 76.4 \% & 0.000 & 0.011 & [0.000, 0.080] & unstable\\
validation median volume Dice & cDDPM & IXI-Guys & 68.1 \% & 0.012 & 0.010 & [0.000, 0.052] & unstable\\
validation median volume Dice & AnomalyDINO & MOOD+IXI & 16.2 \% & 0.005 & 0.009 & [0.000, 0.018] & unstable\\
validation canonical Dice 20 cm$^3$ & REFLECT & MOOD+IXI & 67.7 \% & 0.000 & 0.011 & [0.000, 0.055] & unstable\\
validation canonical Dice 20 cm$^3$ & UCCD & IXI-Guys & 91.4 \% & 0.000 & 0.004 & [0.000, 0.037] & stable\\
validation canonical Dice 20 cm$^3$ & cDDPM & IXI-Guys & 70.5 \% & 0.012 & 0.017 & [0.000, 0.052] & unstable\\
validation canonical Dice 20 cm$^3$ & AnomalyDINO & IXI-HH+Guys & 51.8 \% & 0.000 & 0.004 & [0.000, 0.018] & unstable\\
validation canonical AUPRC & REFLECT & MOOD+IXI & 53.4 \% & 0.000 & 0.015 & [0.000, 0.031] & unstable\\
validation canonical AUPRC & UCCD & IXI-Guys & 99.6 \% & 0.000 & 0.000 & [0.000, 0.000] & stable\\
validation canonical AUPRC & cDDPM & IXI-HH+Guys & 82.6 \% & 0.000 & 0.002 & [0.000, 0.012] & stable\\
validation canonical AUPRC & AnomalyDINO & IXI-HH & 90.0 \% & 0.018 & 0.016 & [0.000, 0.018] & stable\\
validation Dice at 20 cm$^3$, nested $\tau$ & REFLECT & MOOD+IXI & 66.6 \% & 0.000 & 0.012 & [0.000, 0.055] & unstable\\
validation Dice at 20 cm$^3$, nested $\tau$ & UCCD & IXI-Guys & 86.2 \% & 0.000 & 0.007 & [0.000, 0.076] & stable\\
validation Dice at 20 cm$^3$, nested $\tau$ & cDDPM & IXI-Guys & 70.2 \% & 0.012 & 0.018 & [0.000, 0.052] & unstable\\
validation Dice at 20 cm$^3$, nested $\tau$ & AnomalyDINO & IXI-HH+Guys & 55.8 \% & 0.000 & 0.004 & [0.000, 0.018] & unstable\\
\bottomrule
\end{tabular}}
\end{table}
\begin{table}[H]
\caption{Verdicts of H1 to H9 in every set of training data.}
\label{tab:Hsets}
\centering\scriptsize
\setlength{\tabcolsep}{3pt}
\resizebox{\textwidth}{!}{%
\begin{tabular}{@{}lllll@{}}
\toprule
Hypothesis & provisional set & test-selected set & validation-selected set & primary\\
\midrule
ranking instability & supported (max. flip share 0.302 [0.278, 0.374]) & supported (0.459 [0.357, 0.497]) & supported (0.472 [0.393, 0.498]) & supported (0.472 [0.414, 0.499])\\
evaluation dominance & supported (0.607 vs 0.173) & supported (0.598 vs 0.184) & supported (0.613 vs 0.169) & supported (0.614 vs 0.162)\\
intensity baseline & not & not & not & not\\
contrast dependence & not & not & not & partly tested; not\\
pruning trade-off & not & not & not & not (point estimates)\\
test-tuning optimism & REFLECT only & REFLECT (MOOD+IXI) only (lower bound 0.022) & none (lower bound 0.019) & not (lower bound 0.0191 REFLECT, 0.0199 cDDPM)\\
diagnostic power & supported & supported & supported & supported (power 1.000)\\
conformal control & 0/128 over & 1/128 over & 1/128 over & supported (0/128)\\
anatomy adjustment & not & not & not & not (0.668 [0.654, 0.682])\\
\bottomrule
\end{tabular}}
\end{table}

\section{Secondary hypotheses and deviations}\label{app:prereg}

Table~\ref{tab:H} gives the nine primary hypotheses. Table~\ref{tab:P} gives the eight secondary hypotheses, each a prediction about a specific effect that follows from the identities of Section~\ref{sec:method} or from the primary analyses, tested with all methods trained on IXI-HH+Guys. Deviations from the stated analyses are listed after the table.

\begin{table}[h]
\caption{Secondary hypotheses and their outcome (all methods trained on IXI-HH+Guys).}
\label{tab:P}
\centering\scriptsize
\begin{tabular}{@{}>{\raggedright\arraybackslash}p{0.5\textwidth}>{\raggedright\arraybackslash}p{0.3\textwidth}l@{}}
\toprule
Secondary hypothesis & Estimate & Verdict\\
\midrule
UCCD, $20\cm$: the threshold-relaxation term is positive, its interval excludes 0, and it exceeds the fixed-threshold term & $+0.078$ [0.073, 0.083] vs $+0.006$ & held\\
AnomalyDINO, $20\cm$: the total gain from hysteresis with pruning is $\le0$ & $-0.001$ [$-0.011$, 0.010] & held\\
Above-floor sensitivity falls with hysteresis and pruning (UCCD, AnomalyDINO) & $-0.045$; $-0.089$ & held\\
UCCD small-lesion sensitivity $<0.10$ & 0.073 & held\\
Jaccard index of the missed-lesion sets of UCCD with REFLECT and with cDDPM $\ge0.70$ & 0.813; 0.938 & held\\
UCCD transfer gap at $5\cm$ $>0$, interval excludes 0 & $-0.008$ [$-0.008$, $-0.007$] & failed\\
UCCD voxel AUROC falls under antero-posterior and left-right flips; AnomalyDINO slice-level AUROC does not fall under an antero-posterior flip & $-0.198$, $-0.260$; $+0.181$ & held\\
Observed minus borrowed-map AROC $>0$ (UCCD, AnomalyDINO) & $+0.266$; $+0.124$ [0.100, 0.148] & held\\
\bottomrule
\end{tabular}
\end{table}

\paragraph{Deviations and disclosures.} (1) The stated description of the training data names all three IXI sites; our IXI T2 data contain no subjects from the third site (IOP), so IXI-HH+Guys comprises the Hammersmith and Guy's sites, identically for all methods. (2) The canonical arrays are a through-plane-cropped copy (98--149 slices) of the $240\times240\times155$ originals; no tumour voxel is affected. (3) The contrast-dependence hypothesis could be tested only in part, because the available labels are binary. (4) The pruning trade-off was judged from point estimates, as no interval was computed for it. (5) The stratified and matched slice-level estimators named in H9 fail their covariate-only check; the regression estimator, added as an exploratory alternative, fails as well. (6) The family of six latent-aggregation rules specified for this experiment could not be computed from the stored latents; the family reported here is exploratory. (7) A pre-specified through-plane offset check was positive for AnomalyDINO, whose voxel AUROC peaks three slices from the canonical position ($+0.008$); its export was verified to be identical to the canonical data and its numbers are reported without a shift, a decision made with knowledge of the check's result. (8) One subject fails the overlap floor of the registration gate for UCCD and cDDPM and is kept; excluding it is immaterial (Section~\ref{sec:res-R}). (9) No untouched BraTS 2020 subjects remain for a hold-out set, so confirmation requires a new cohort, to be evaluated under a pre-registered protocol that fixes the training set and every threshold rule.

\paragraph{Software.} Analyses used Python 3.11 with NumPy, pandas, SciPy and statsmodels, and R with lme4 for the mixed models and logistf for Firth regression. The evaluator implements every gate, diagnostic, metric and identity of Section~\ref{sec:method} and is checked against 31 synthetic test cases, each with a known answer. Computational cost is reported in Appendix~\ref{app:compute}.

\end{document}